%% file: main.tex
\documentclass[10pt]{article} 
\usepackage[preprint]{tmlr}

\input{math_commands.tex}

\usepackage{hyperref}
\usepackage{url}
\usepackage{booktabs}
\usepackage{siunitx}
\usepackage{mathtools}
\usepackage{algorithm}
\usepackage{algpseudocode}
\usepackage{graphicx}
\usepackage{caption}
\usepackage[subrefformat=parens]{subcaption}
\usepackage{amsmath, amssymb}
\usepackage{enumitem}
\usepackage{comment}
\usepackage{pifont}
\usepackage{makecell}

\title{Self-Augmented Diffusion Guidance for Physics-Informed Generation}

\author{\name Akira Osaka \email akr-osaka@g.ecc.u-tokyo.ac.jp \\
      \addr School of Engineering\\
      The University of Tokyo
      \AND
      \name Naoya Takeishi \email ntake@g.ecc.u-tokyo.ac.jp \\
      \addr School of Engineering\\
      The University of Tokyo
      \AND
      \name Takehisa Yairi \email yairi@g.ecc.u-tokyo.ac.jp\\
      \addr School of Engineering\\
      The University of Tokyo}

\def\month{MM}  
\def\year{YYYY} 
\def\openreview{\url{https://openreview.net/forum?id=XXXX}} 

\begin{document}

\maketitle

\begin{abstract}
Diffusion models can be used to generate spatiotemporal signals of physical phenomena, such as time-series images of fluid dynamics. However, a major limitation of standard diffusion models is that they do not incorporate constraints derived from the underlying physical laws. Consequently, generated samples may appear visually plausible while deviating substantially from the true dynamics. In this study, we propose a simple yet effective physics-informed approach based on diffusion guidance with self-generated data augmentation. The proposed method learns the data distribution conditioned on the degree of deviation from the physically correct dynamics and generates samples by explicitly setting the deviation condition to be zero.
The method decouples the evaluation of the governing equations from the diffusion model training and sampling processes, avoiding the need to solve the governing equations at every iteration of the denoising process. This design makes the method applicable to problems requiring computationally expensive numerical simulations and enables faster sample generation. Experimental results demonstrate that the proposed model not only significantly reduces the deviations compared with standard diffusion models but also achieves further reductions when combined with existing physics-constrained diffusion methods.
\end{abstract}

\section{Introduction}

Diffusion models provide a powerful framework for learning probability distributions and generating realistic samples. Their applications extend beyond general image generation to physical phenomena in fields such as fluid dynamics \citep{shu2023}, oceanography \citep{xu2025}, and atmospheric science \citep{wang2025}. However, when applied to physical problems, standard diffusion models, including score-based models \citep{song2019, song2021} and denoising models \citep{sohl-dickstein2015, ho2020, song2021b}, face a critical limitation that they do not explicitly account for the underlying physical laws. Consequently, they may generate samples that appear visually plausible but are physically inconsistent, reducing their validity and practical utility.

Several previous studies have addressed this issue by developing diffusion models that enforce adherence to governing physical laws. The Physics-Informed Diffusion Model (PIDM) \citep{bastek2025} reduces discrepancies between generated samples and the underlying physical constraints by incorporating the residuals of the governing equations into the training loss. PIDM imposes these constraints only during training, leaving the sampling process unchanged from that of standard diffusion models. Among the methods that incorporate physical constraints during sampling, CoCoGen \citep{jacobsen2025} modifies the sampling procedure so that denoising proceeds in a direction that reduces the constraint violation. Similarly, Physics-Guided Diffusion (PG Diffusion) \citep{shu2023} integrates residual information into the model architecture using classifier-free guidance. Regarding PG Diffusion, \citet{bastek2025} point out that although the method incorporates gradient information into the model, it does not explicitly enforce residual minimization, resulting in insufficient physical consistency. This observation highlights the need for methods that explicitly impose a zero-residual condition during sampling.

Another critical challenge is computational cost. Existing methods \citep{bastek2025, jacobsen2025, shu2023} require the evaluation and differentiation of the constraints during either training or sampling. In some problems, such as the generation of time-series data in fluid dynamics, snapshots are generated at spatial and temporal resolutions coarser than those required for stable and accurate numerical differentiation. Such nature of data makes it inapplicable to evaluate the residuals of the partial differential equations (PDEs) as a mean to impose physics constraints expressed as PDEs. In such cases, instead, the degrees of constraint satisfaction should be evaluated by measuring the deviations between generated samples and physically consistent time-series samples obtained by numerically integrating the governing equations with smaller internal steps. Evaluating such integration-based constraints and their derivatives are typically more computationally expensive than diffusion model training or sampling, especially in high-dimensional settings, and can result in prohibitively long runtimes. To broaden the applicability of physics-informed generative models, it is important to develop a method that separates constraint evaluation from diffusion training and sampling while still incorporating physical information into the model.

In this study, we propose a simple yet effective physics-aware diffusion guidance method based on the classifier-free guidance framework. Our method is based on a self-augmentation strategy in which a base diffusion model is used to generate a kind of negative samples, while the original dataset is treated as positive samples. The diffusion model is then guided to generate samples that are closer to the positive samples and further from the negative.
Furthermore, for problems requiring computationally expensive numerical simulations, our approach decouples constraint evaluation from diffusion model training and sampling. It eliminates the need for calculating gradients of the constraint functions during either training or sampling.

Unlike PG Diffusion \citep{shu2023}, which also employs classifier-free guidance, our method explicitly enforces physical consistency by sampling from a diffusion model conditioned on zero constraint violation. The Extrapolation-Correction-Interpolation (ECI) sampling method \citep{cheng2025}, based on flow matching \citep{lipman2023}, also achieves gradient-free generation by taking advantage of its deterministic flow structure. However, it assumes the availability of an appropriate correction algorithm that projects the samples to satisfy the constraints. Our approach further removes this requirement and is applicable as long as the constraints can be evaluated somehow. Table \ref{tb_method_requirements} compares the requirements of previous approaches with those of our method. Our method does not rely on any of the assumptions required by the existing methods.

\begin{table*}[t]
    \centering
    \caption{Comparison of assumptions required by previous physics-informed generative methods and ours.}
    \label{tb_method_requirements}
    \setlength{\tabcolsep}{5pt}
    \small
    \begin{tabular}{lcccc}
        \toprule
        Method & \makecell{Constraint Evaluation\\during Training} & \makecell{Constraint Evaluation\\during Sampling} & \makecell{Gradient \\of Constraint} & \makecell{Projection /\\Correction}\\
        \midrule
        PG Diffusion \citep{shu2023} & \checkmark & \checkmark & \checkmark & --\\
        PIDM \citep{bastek2025} & \checkmark & -- & \checkmark & -- \\
        CoCoGen \citep{jacobsen2025} & -- & \checkmark & \checkmark & --\\
        ECI \citep{cheng2025} & -- & -- & -- & \checkmark \\
        Proposed & -- & -- & -- & --\\
        \bottomrule
    \end{tabular}
    \vspace{0.3mm}
    \begin{minipage}{0.97\linewidth}
        \footnotesize
        Note: ECI is based on flow matching, whereas PG Diffusion, PIDM, CoCoGen, and ours are based on diffusion models.
    \end{minipage}
\end{table*}

We evaluate our method on fluid dynamics tasks and demonstrate that it significantly reduces deviations from the underlying physical laws. Our approach can also be readily combined with existing physics-informed diffusion methods in a plug-and-play manner. We show that combining it with previous approaches such as PIDM and CoCoGen yields further improvements in physical consistency. 

\section{Related work}

As diffusion models and flow matching have become increasingly popular for generative modeling, constrained generation has attracted growing attention in recent years. One of the early studies in this area is \citet{fishman2023}, which investigated diffusion models defined on constrained domains. Subsequent studies have explored several directions, such as constraint-aware diffusion training \citep{khalafi2024}, gradient-guided sampling for constraint satisfaction \citep{huang2024b}, and projection onto constrained regions during sampling \citep{christopher2024}. In flow matching, \citet{li2026b} utilized its differential-equation structure to formulate constrained generation as an optimal control problem.

Generative modeling under physical constraints is a major application area of constrained generation. A common approach in physics-informed diffusion models and flow matching is to incorporate physical knowledge into the sampling process. \citet{yuan2023} proposed PhysDiff, which directly modifies intermediate samples during denoising via imitation learning to generate physically plausible human motions. In the context of flow matching, \citet{cheng2025}, \citet{utkarsh2025}, and \citet{christopher2026} developed methods that extrapolate intermediate states to the terminal time, project or refine the resulting terminal-state estimates to satisfy the constraints, and then propagate the corrections back to the current sampling steps. These methods exploit the deterministic nature of flow matching and apply the constraint operations to estimated terminal states rather than directly to intermediate states. For diffusion models, \citet{li2026} also introduced projections toward feasible regions during denoising. Similarly, \citet{blanke2026} integrated projection steps into Langevin sampling. Diffusion posterior sampling \citep{chung2023, yao2025}, which is designed to solve inverse problems, has also been applied to physics-constrained generation. \citet{huang2024} and \citet{gallon2026} proposed methods that correct intermediate samples along the gradient of the residual. \citet{peng2026} injected the physics gradient into the noise-prediction steps. Similarly, CoCoGen \citep{jacobsen2025} modifies the sampling process so that the denoising steps move in a direction that reduces the residual:
\begin{equation}
    x_{i-1}
    =
    \operatorname{Solver}(x_i)
    -
    \gamma \nabla_{x_i}\lVert r(x_i)\rVert_2^2,
\end{equation}
where $x_i$ is an intermediate state, $r(x)$ is the residual of the governing equation, and $\gamma$ is a hyperparameter.

Another major approach is to introduce constraints during training to reduce deviations from physically consistent dynamics. In the context of flow matching, \citet{tauberschmidt2026} incorporated the Adjoint Matching framework \citep{domingo-enrich2025} to fine-tune the flow for lower deviations using an additional control term. For diffusion models, PIDM \citep{bastek2025} and related training-based approaches \citep{wang2025, zeng2026, bai2026, ismayilzada2026} incorporate physics-based constraints such as physics residuals into the training objective. For example, when training a model to estimate the added noise $\epsilon$, a simplified form of the PIDM objective can be written as
\begin{equation}
    \lVert \epsilon_\theta (x_t, t) - \epsilon \rVert^2 + \lambda\lVert r(\mathbb{E}[x_0 \mid x_t])\rVert^2, \label{eq_pidm}
\end{equation}
where $\epsilon_\theta (x_t, t)$ denotes the function that estimates the noise, $r(x)$ denotes the residual, and $\lambda$ is a hyperparameter \citep{bastek2025}. Related extensions of PIDM have also been explored in settings such as knowledge distillation \citep{zhang2026} and generation without full knowledge of the underlying dynamics \citep{xu2025, tan2026}.

Classifier-free guidance \citep{ho2022}, which provides a method for learning a conditional distribution $p(x\mid c)$, has also been applied to physics-informed generation. Although it was originally introduced for general image generation, PG Diffusion \citep{shu2023} uses classifier-free guidance to integrate physical information into the model architecture when reconstructing high-resolution images from low-fidelity inputs. By using the gradient of the physical residual as conditioning information within the classifier-free guidance framework, they incorporate residual information into the model and enables physics-aware training and sampling. However, as noted by \citet{bastek2025}, PG Diffusion does not explicitly guide the sampling process toward a zero-residual condition.

Although these approaches enable physics-aware generation, they require constraint evaluation during either training or sampling. In some problem settings, such as generating snapshot time-series of 3D fluid flow, the physical simulations required for constraint evaluation are computationally expensive and can substantially increase training and sampling times. Motivated by this limitation, our approach separates physical simulations required for constraint evaluation from both diffusion training and sampling, avoiding a substantial increase in computational cost even in such cases.

\section{Background}

\subsection{Diffusion models}

Diffusion models \citep{ho2020} are generative models that learn the reverse-time dynamics of a diffusion process, allowing data to be generated from Gaussian noise. Let $t \in [0, T]$ denote a time step and $x_t\sim p_t(x)$ be an intermediate noisy sample. The forward diffusion process is defined as
\begin{equation}
    q(x_t \mid x_{t - 1}) = \mathcal{N}(x_t; \sqrt{1 - \beta_t}x_{t - 1}, \beta_t I),
\end{equation}
where $\beta_t$ denotes the noise variance added at time $t$. Given Gaussian noise $\epsilon\sim\mathcal{N}(0, I)$, a noisy sample at time $t$ can be obtained as follows:
\begin{equation}
    x_t(x_0, \epsilon) = \sqrt{\bar{\alpha}_t}x_0 + \sqrt{1 - \bar{\alpha}_t}\epsilon,~~\bar{\alpha}_t = \prod_{s=1}^t (1 - \beta_s).
\end{equation}

The corresponding reverse process, parameterized by a set of learnable parameters $\theta$, can be written as
\begin{equation}
    p_\theta (x_{t - 1}\mid x_t) = \mathcal{N} (x_{t - 1}; \mu_\theta (x_t, t), \sigma_t^2 I),
\end{equation}
where $\sigma_t^2$ is a fixed variance. The diffusion model is trained by optimizing $\theta$ so that the learned reverse process approximates the data distribution.

According to \citet{ho2020}, the parameter set $\theta$ can be trained by predicting the added noise $\epsilon$ from the noisy sample $x_t$:
\begin{equation}
    \min_\theta \mathbb{E}_{t, x_0, \epsilon} \Vert\epsilon - \epsilon_\theta(\sqrt{\bar{\alpha}_t}x_0 + \sqrt{1 - \bar{\alpha}_t}\epsilon, t)\Vert^2, \label{eq_noise_est}
\end{equation}
where the function $\epsilon_\theta (x, t)$ predicts the noise added to $x$.

After training, samples can be generated by starting from Gaussian noise at $t = T$ and iteratively applying the following reverse process until $t = 0$:
\begin{equation}
    x_{t - 1} = \frac{1}{\sqrt{1 - \beta_t}}\left(x_t - \frac{\beta_t}{\sqrt{1 - \bar{\alpha}_t}}\epsilon_\theta(x_t, t)\right) + \sigma_t z,
\end{equation}
where $z\sim \mathcal{N}(0, I)$.

\subsection{Diffusion guidance}

\citet{dhariwal2021} proposed classifier guidance, which uses the gradient of a classifier to guide a diffusion model toward samples satisfying a condition $c$. The conditional noise-estimation model takes the condition $c$ as an additional input. The model parameters are optimized as
\begin{equation}
    \min_\theta \quad
    \left\lVert
        \epsilon_\theta(x_t, t, c) - \epsilon
    \right\rVert^2.\label{eq_conditional_training}
\end{equation}

During sampling, the estimated noise $\epsilon_\theta$ is modified using a classifier $p_\phi(c\mid x_t)$ trained on noisy samples \citep{ho2022}:
\begin{equation}
    \tilde{\epsilon}_\theta(x_t, t, c)
    =
    \epsilon_\theta(x_t, t, c)
    -
    w\sigma_t
    \nabla_{x_t}\log p_\phi(c\mid x_t),
    \label{eq_clg}
\end{equation}
where $w$ is a guidance scale, and $\sigma_t$ is the standard deviation of the noise at time $t$. As discussed in \citet{ho2022}, Eq. \ref{eq_clg} effectively modifies the probability distribution as
\begin{equation}
    \tilde{p}_\theta(x\mid c)
    \propto
    p_\theta(x\mid c)
    p_\phi(c\mid x)^w,
\end{equation}
which increases the likelihood of samples that the classifier associates with condition $c$ and guides the sampling process accordingly.

Although classifier guidance improves sample quality, it requires training an additional classifier and complicates the overall procedure. To address this issue, \citet{ho2022} proposed classifier-free guidance, which achieves a similar effect without an explicit classifier. In this approach, the same model is trained with and without condition $c$, and sampling is guided by modifying the noise estimate $\epsilon_\theta$ as
\begin{equation}
    \tilde{\epsilon}_\theta(x_t, t, c)
    =
    (1+w)\epsilon_\theta(x_t, t, c)
    -
    w\epsilon_\theta(x_t, t),
    \label{eq_cfg}
\end{equation}
which can be interpreted as using an implicit classifier:
\begin{equation}
    p^\text{i}(c\mid x)
    \propto
    \frac{p(x\mid c)}{p(x)}.
\end{equation}

\section{Method}

We propose a physics-informed diffusion framework with data augmentation. The main idea of the proposed approach is to train the guidance model using original data, which serve as physically consistent positive samples, and augmented data, which serve as negative samples with some deviations from the underlying physics. The sampling process is then guided toward the positive samples. 

\begin{figure}[t]
    \centering
    \includegraphics[width=\linewidth]{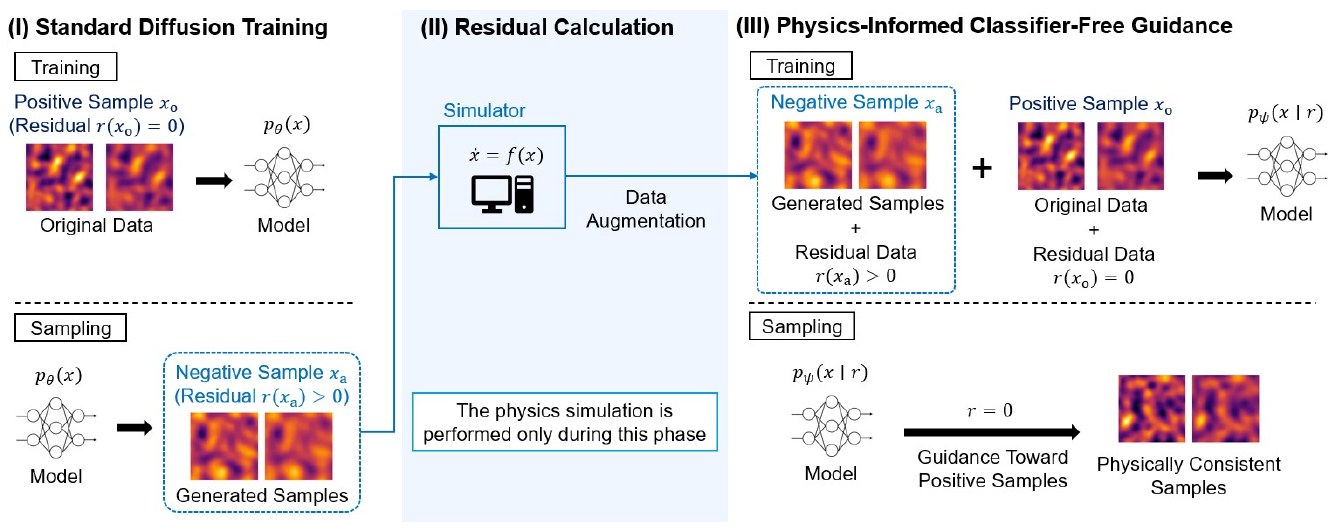}
    \caption{Overview of the proposed guidance approach.}
    \label{methods}
\end{figure}

The augmented data consist of visually plausible but physically inconsistent negative samples generated by an ordinary diffusion model before the guidance model is trained. As this initial diffusion model generates the augmented data that are subsequently used to train the guidance model, we refer to this process as self-augmentation. An overview of the proposed framework is shown in Figure \ref{methods}.

Let $\mathcal{D}_\text{o} = \{x_\text{o}\}$ denote the original dataset, where each sample satisfies the governing equation $E(x) = 0$. The objective is to generate samples that are as close as possible to satisfying $E(x) = 0$. We evaluate deviations from the correct dynamics either by calculating the PDE residuals $\lVert E(x) \rVert$ or by measuring the deviations between generated time-series samples and numerically integrated reference samples. For simplicity, we use the term residual to refer to both types of deviation throughout this study and define a residual function $r(x)$ to quantify them.

\subsection{Standard diffusion training for data augmentation and constraint evaluation}

First, we train a diffusion model without guidance on the original dataset $\mathcal{D}_\text{o}$ to obtain an approximate distribution $p_\theta (x)$. Any diffusion algorithm, such as DDPM \citep{ho2020}, can be used for this purpose. After training, we sample from the model to construct an augmentation dataset:
\begin{equation}
    \mathcal{D}_\text{a} = \{x_\text{a} \sim p_\theta (x)\}.
\end{equation}

Subsequently, we evaluate residuals with $r(x)$. The original samples $x_\text{o}$ are assumed to satisfy the governing equation, such that $r(x_\text{o}) = 0$. In contrast, the generated samples $x_\text{a}$ generally exhibit nonzero residuals because the model is trained without physical constraints, resulting in $r(x_\text{a}) > 0$. Thus, we treat $x_\text{o}$ as positive samples satisfying the physical law, and $x_\text{a}$ as negative samples violating it.

We then construct an augmented dataset by combining the original and generated samples with their corresponding residuals:
\begin{equation}
    \mathcal{D}' = \{(x, r)\} = \{(x_\mathrm{o}, r(x_\mathrm{o}))\} \cup \{(x_\mathrm{a}, r(x_\mathrm{a}))\}.
\end{equation}
    
\subsection{Physics-informed classifier-free guidance}

Finally, we train a classifier-free guidance model \citep{ho2022} conditioned on the residual $r$ using the dataset $\mathcal{D}'$, thereby obtaining an approximate distribution $p_\psi(x\mid r)$. Based on the general conditional training objective in Eq. \ref{eq_conditional_training}, we use the physical residual $r$ as the condition $c$ and define the training objective as follows:
\begin{equation}
    \min_\psi \quad \lVert \epsilon_\psi (x_t, t, r) - \epsilon \rVert^2. \label{eq_training}
\end{equation}
As in standard classifier-free guidance, the condition $r$ is randomly replaced with a null condition during training, allowing the same model to learn both conditional and unconditional noise estimates.

After training, we generate samples from $p_\psi(x\mid 0)$ by setting the residual condition to $r=0$, which is expected to guide the model toward samples with lower physical residuals. Following Eq. \ref{eq_cfg}, the noise estimate is modified as
\begin{equation}
    \tilde{\epsilon}_\psi (x_t, t, 0) = (1 + w)\epsilon_\psi (x_t, t, 0) - w\epsilon_\psi (x_t, t). \label{eq_sampling}
\end{equation}

In the proposed method, any physical simulation required for residual evaluation is performed independently of diffusion training and sampling. During the classifier-free guidance step, the model uses only the precomputed residuals. Algorithm \ref{alg1} presents the overall procedure of the proposed method.

\begin{algorithm}[htbp]
    \caption{Physics-informed classifier-free guidance}
    \label{alg1}
    \begin{algorithmic}[1]
        \Require $r(x)$: function to evaluate residuals, $\mathcal{D}_{\mathrm{o}}$: original dataset, $\mathcal{L}_{\mathrm{DM}}$: diffusion loss, $\mathcal{L}_{\mathrm{CFG}}$: classifier-free guidance loss
        
        \State \textbf{Diffusion training and sampling}
        \State $\theta \leftarrow \arg\min_{\theta}\mathcal{L}_{\mathrm{DM}}(\theta;\mathcal{D}_{\mathrm{o}})$
        \State $\mathcal{D}_{\mathrm{a}}=\{x_\mathrm{a}\},\quad x_\mathrm{a}\sim p_{\theta}(x)$
        \State \textbf{Residual calculation and data augmentation}
        \State $\mathcal{D}' \leftarrow \{(x_\text{o},0)\mid x_\text{o}\in\mathcal{D}_{\mathrm{o}}\}\cup\{(x_\text{a}, r(x_\text{a}))\mid x_\text{a}\in\mathcal{D}_{\mathrm{a}}\}$
        
        \State \textbf{Guidance model training and sampling}
        \State $\psi \leftarrow \arg\min_{\psi}\mathcal{L}_{\mathrm{CFG}}(\psi;\mathcal{D}')$
        \State $x^{*} \sim p_{\psi}(\cdot\mid0)$
    \end{algorithmic}
\end{algorithm}

\subsection{Extensibility}
While the proposed algorithm can improve physical consistency on its own, it can also be readily extended to achieve further performance improvements.

First, the proposed algorithm can be applied iteratively by using samples generated by the guidance model as additional negative samples. Specifically, these newly generated samples $x_\text{a}' \in \mathcal{D_\text{a}'}$ are added to the existing set of negative samples, and the guidance model is retrained on the expanded dataset $\mathcal{D}'' = \mathcal{D}' \cup \{(x_\mathrm{a}', r(x_\mathrm{a}'))\}$. Samples are then generated again by conditioning on zero residual. Repeating this procedure allows the model to progressively generate samples with lower residuals.

Second, the proposed method can be readily combined with existing approaches that incorporate physical constraints during training (e.g., PIDM \citep{bastek2025}) or during sampling (e.g., CoCoGen \citep{jacobsen2025}). The proposed method differs from ordinary diffusion models in four main aspects, (i) the generation of negative samples, (ii) the inclusion of a simulation step, (iii) the introduction of the conditional variable $r$ into the noise estimation model $\epsilon_\psi$ in Eq. \ref{eq_training}, (iv) the guidance mechanism in Eq. \ref{eq_sampling}. These modifications do not fundamentally alter the structure of the diffusion model, which allows the proposed method to be applied to a wide range of existing diffusion-based approaches in plug-and-play manner.

\section{Experiments}

We demonstrate physics-guided generation using the proposed method on Darcy flow and time-series fluid data.

\subsection{Darcy flow}\label{sec_exp_darcy}

\subsubsection{Problem setting}

In this section, we compare the proposed method with existing methods on the Darcy flow problem, which is commonly used as a benchmark in previous studies.

The Darcy flow equations describe fluid flow through porous media. Let $K(x)$ denote the permeability, $p(x)$ the pressure field, $u(x)$ the velocity field, and $f(x)$ a source function representing where fluid enters and exits the domain. As introduced in the related studies \citep{bastek2025, jacobsen2025}, the governing equations are given by:
\begin{equation}
    \begin{aligned}
        u(x) &= -K(x)\nabla p(x), \quad x\in \Omega,\\
        \nabla\cdot u(x) &= f(x), \quad x\in\Omega,\\
        u(x)\cdot\hat{n}(x) &= 0, \quad x\in\partial\Omega,\\
        \int_\Omega p(x)dx &= 0,
    \end{aligned}
    \label{eq_darcy}
\end{equation}
where $\hat{n}(x)$ denotes the outward normal vector on the boundary.

In this experiment, the purpose of the diffusion model is to generate samples of the pressure field $p(x)$ and the permeability field $K(x)$ while maintaining consistency with the static relations defined in Eq.~\ref{eq_darcy}. The residual is defined as PDE residual of the governing equation.

We use the Darcy flow dataset released by \citet{bastek2025}. The dataset consists of 10000 training samples and 1000 validation samples, each with a resolution of $64 \times 64$. The number of negative samples generated in the first step of our method is 10000, matching the size of the training set.

We compare the proposed method with a standard DDPM without physical constraints and evaluate the reduction in residuals achieved by the proposed method. We also evaluate the extensibility of our approach by applying a second round of guidance. We further combine the proposed method with other physics-informed diffusion approaches, including PIDM \citep{bastek2025}, which incorporates physical constraints during training, and CoCoGen \citep{jacobsen2025}, which incorporates them during sampling. In addition, we compare our method with PG Diffusion \citep{shu2023}, another approach using classifier-free guidance. In each setting, we repeated the experiments 8 times using different random seeds. More detailed settings are provided in Appendix \ref{ap_darcy_set}.

\subsubsection{Results}\label{subsec_darcy_res}

\begin{figure}[t]
    \centering

    \begin{subfigure}{\linewidth}
        \centering
        \includegraphics[width=0.48\linewidth]{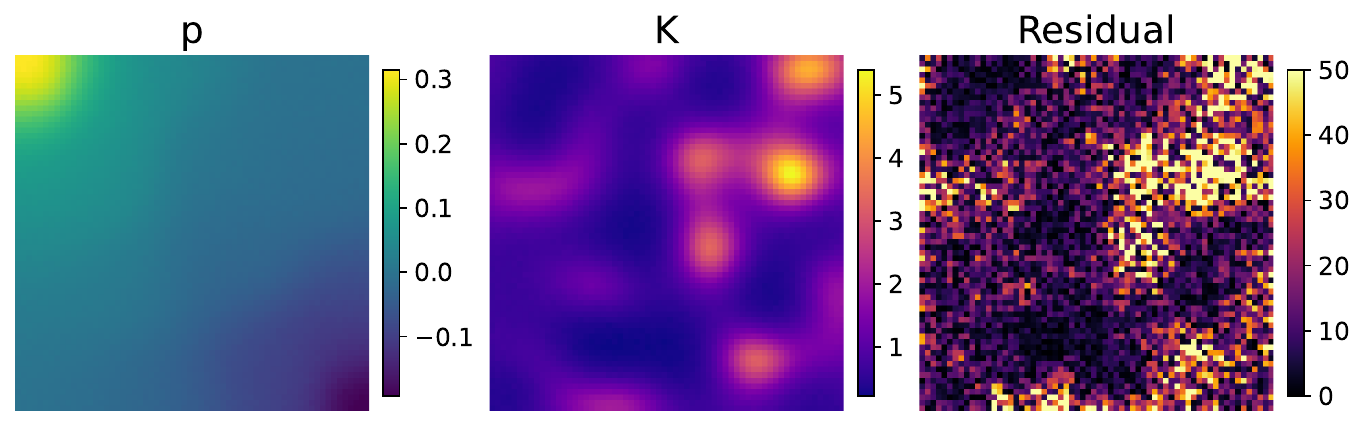}
        \hfill
        \includegraphics[width=0.48\linewidth]{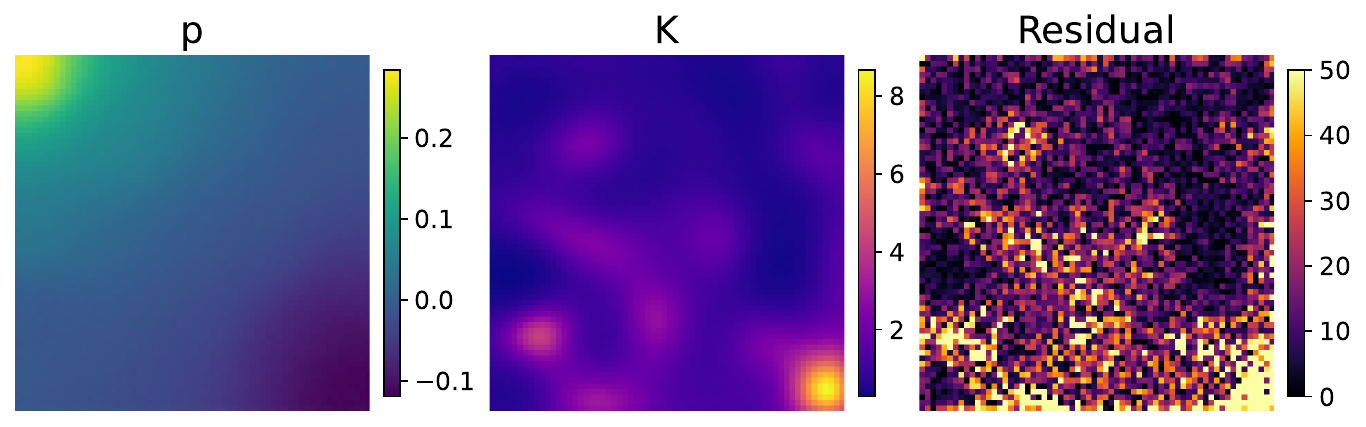}
        \caption{Standard diffusion model.}
    \end{subfigure}

    \vspace{0.6em}

    \begin{subfigure}{\linewidth}
        \centering
        \includegraphics[width=0.48\linewidth]{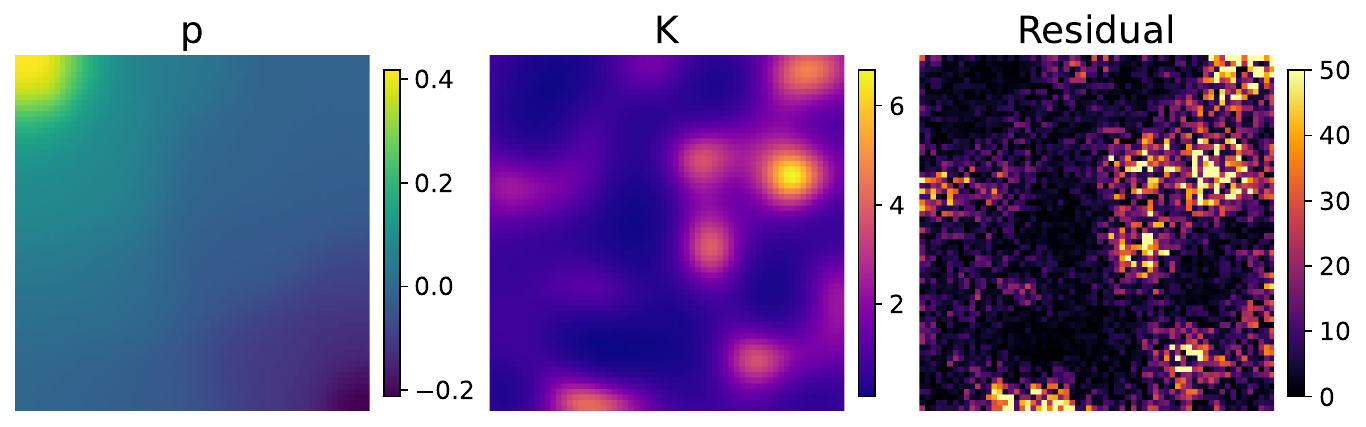}
        \hfill
        \includegraphics[width=0.48\linewidth]{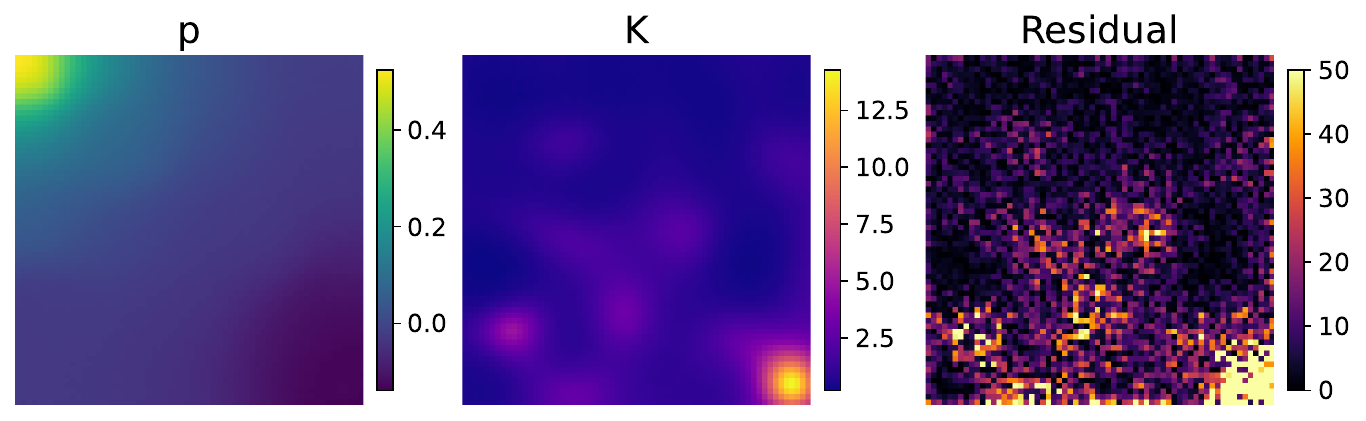}
        \caption{Proposed guidance model.}
    \end{subfigure}

    \caption{Comparison of samples and residuals generated by the standard diffusion model and the proposed guidance model in the Darcy flow experiment. The residual represents the PDE residual of the Darcy flow equation.}
    \label{comparison_darcy}
\end{figure}

Figure \ref{comparison_darcy} compares samples generated by the standard diffusion model and the proposed guidance model. The comparison visually demonstrates that our classifier-free guidance approach reduces the residuals.

\begin{figure}[t]
    \centering
    \includegraphics[width=\linewidth]{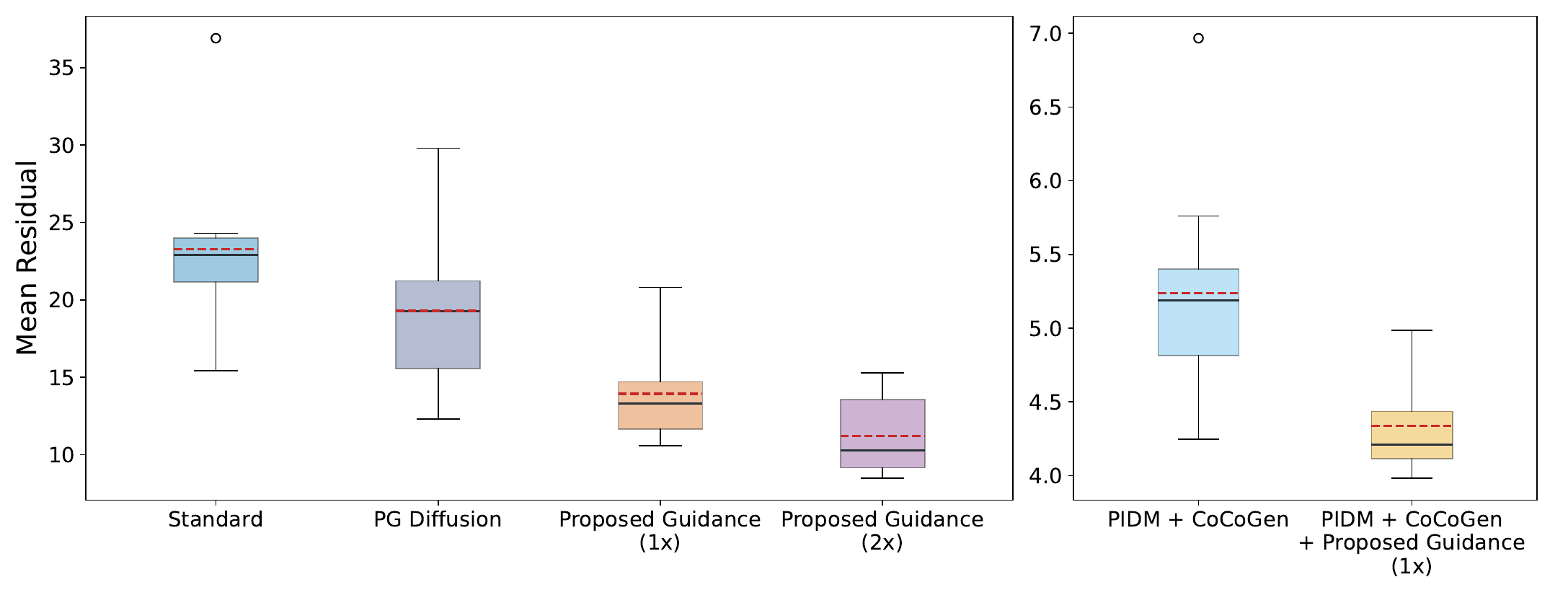}
    \caption{Distribution of the mean residuals across independent runs under different conditions in the Darcy flow experiment. The red dashed lines show the mean values. The notation 1x and 2x indicate that the proposed method is applied once and twice, respectively.}
    \label{comparison_mean_darcy}
\end{figure}

\begin{figure}[t]
    \centering

    \begin{subfigure}{\linewidth}
        \centering
        \includegraphics[width=0.48\linewidth]{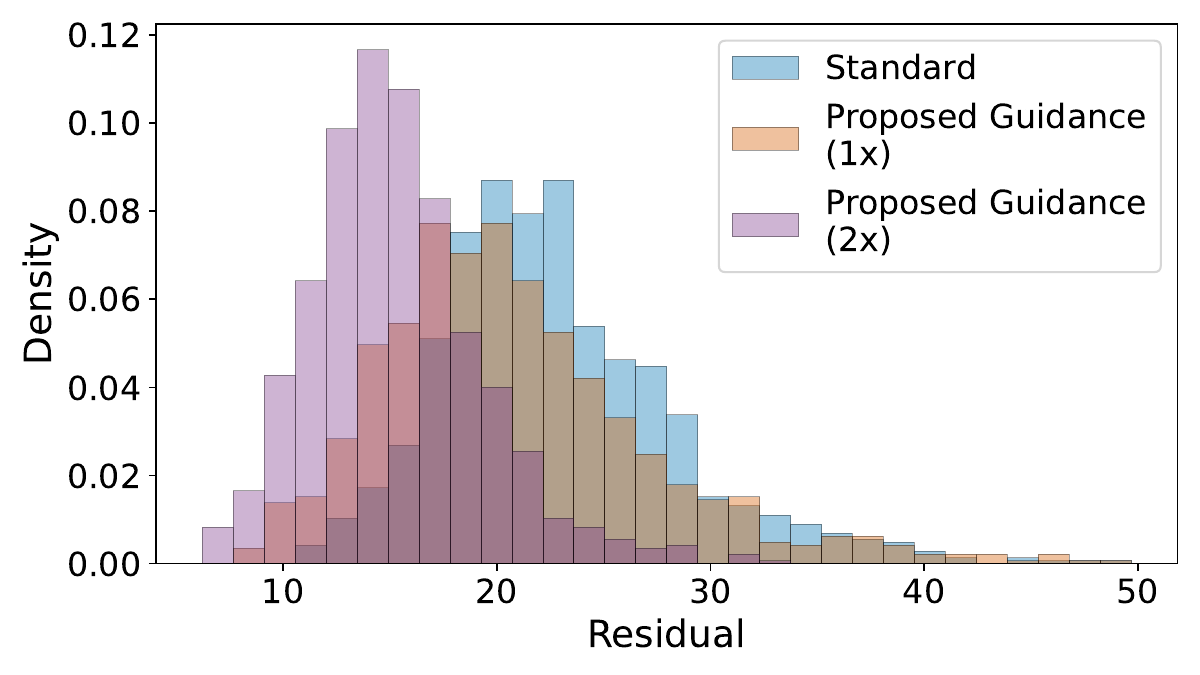}
        \hfill
        \includegraphics[width=0.48\linewidth]{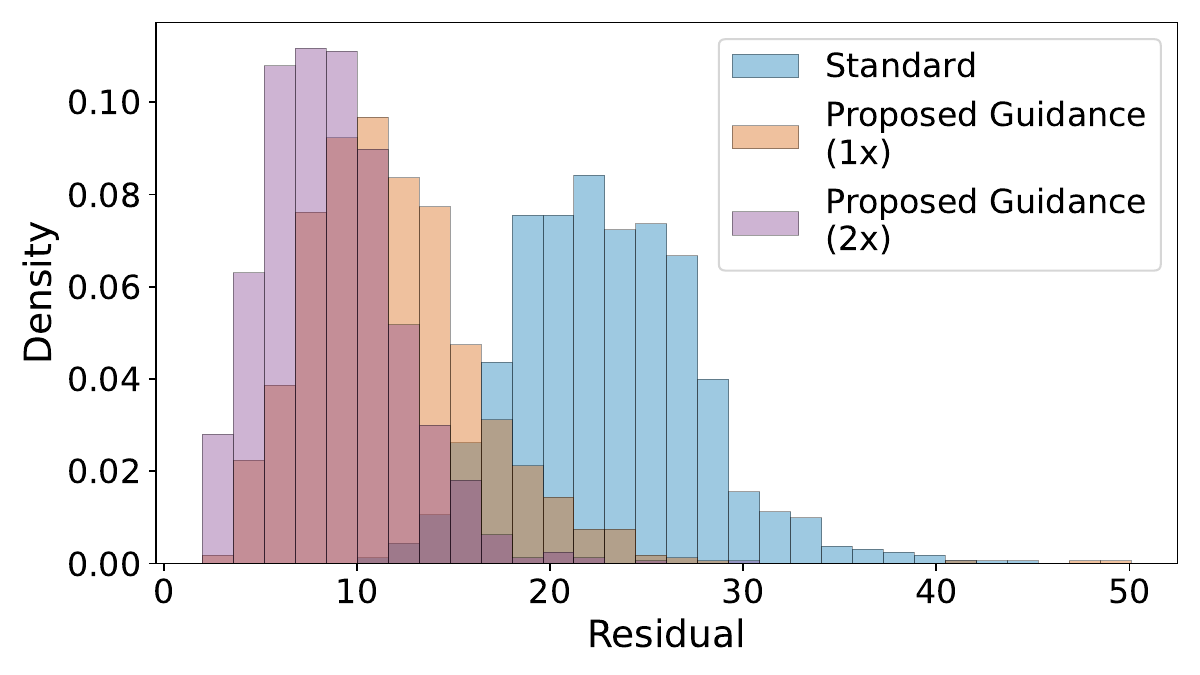}
    \end{subfigure}
    
    \caption{Residual distributions of 1000 generated samples in 2 different trials in the Darcy flow experiment.}
    \label{sample_dist_darcy}
\end{figure}

Figure \ref{comparison_mean_darcy} shows the distribution of the mean residuals over 8 trials with different random seeds. In each trial, 1000 samples were generated to calculate the mean. The results quantitatively demonstrate that the proposed method reduces the residual relative to the baseline and that repeated application of the proposed method further reduces it. A similar trend is observed in Figure \ref{sample_dist_darcy}, which shows the residual distributions within individual trials. The distributions shift toward zero after applying the proposed method, with an additional shift observed after the second application.

Figure \ref{comparison_mean_darcy} also demonstrates that the proposed method achieves lower residuals than PG Diffusion. We hypothesize that this improvement is attributable to the sampling strategy of the proposed method, which uses zero residual as an explicit target condition. Furthermore, combining the proposed method with both PIDM and CoCoGen achieves the best overall performance, outperforming the PIDM + CoCoGen baseline. This result highlights a key advantage of our approach, which is compatible with existing physics-informed diffusion methods for further improvements in physical consistency. Additional experimental results are provided in Appendix \ref{ap_darcy_ex}.

\subsection{Time-series generation of 2D fluid dynamics}\label{sec_exp_fluid}

\subsubsection{Problem setting}

The second experiment considers the generation of time-series snapshots of 2D fluid dynamics.

Let $u$ denote the velocity field, $p$ the pressure, $f$ the external forcing, $Re$ the Reynolds number, and $\rho$ the density. We consider 2D incompressible flows as introduced in \citep{kochkov2021}, described by the following Navier--Stokes equations:
\begin{equation}
    \frac{\partial u}{\partial t} = -\nabla\cdot (u\otimes u) + \frac{1}{Re}\nabla^2 u - \frac{1}{\rho}\nabla p + f, \label{eq_ns1}
\end{equation}
\begin{equation}
    \nabla \cdot u = 0, \label{eq_ns2}
\end{equation}
where $\otimes$ denotes the tensor product. The vorticity associated with the velocity field in Eqs. \ref{eq_ns1} and \ref{eq_ns2} is defined as
\begin{equation}
    \omega = \frac{\partial u_y}{\partial x} - \frac{\partial u_x}{\partial y}.
\end{equation}

In this experiment, we consider the generation of 2D vorticity fields in the absence of external forcing (i.e., $f = 0$, corresponding to decaying flow). We prepared 10000 training samples and 1000 validation samples, each consisting of 4 snapshots showing the evolution over 3 seconds at a resolution of $64 \times 64$. The number of negative samples generated in the first step of our method is 10000, matching the size of the training set.

To calculate the residuals, we use the generated snapshot at $t=0$ as the initial condition and numerically integrate Eq. \ref{eq_ns1} and \ref{eq_ns2} to obtain a physically consistent reference trajectory. The residuals are defined as the deviations between the generated samples and the corresponding reference snapshots. For numerical stability, the integration timestep $\Delta t$ is set to 0.05, which is smaller than the snapshot interval of 1.0. As the snapshots are generated at a coarser temporal resolution than that adopted for stable numerical integration, directly calculated PDE residuals at the snapshot interval are not suitable for evaluating physical consistency.

As in the Darcy flow experiment, we compare the proposed method with a standard DDPM, PG Diffusion \citep{shu2023}, PIDM \citep{bastek2025}, and CoCoGen \citep{jacobsen2025} baselines. In each setting, we repeated the experiments 8 times using different random seeds. More detailed settings are provided in Appendix \ref{ap_fluid_set}.

\subsubsection{Results}\label{subsec_fluid_res}

\begin{figure}[t]
    \centering

    \begin{subfigure}{\linewidth}
        \centering
        \includegraphics[width=0.48\linewidth]{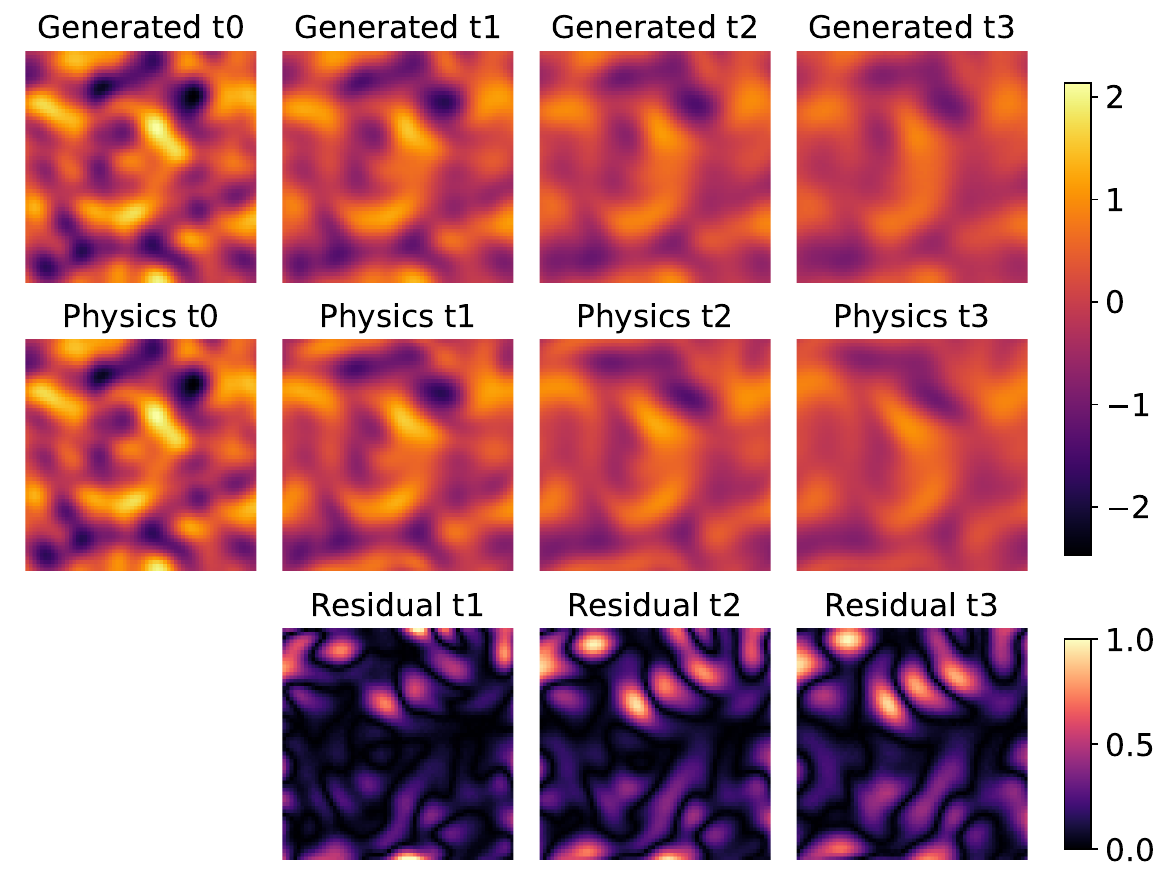}
        \hfill
        \includegraphics[width=0.48\linewidth]{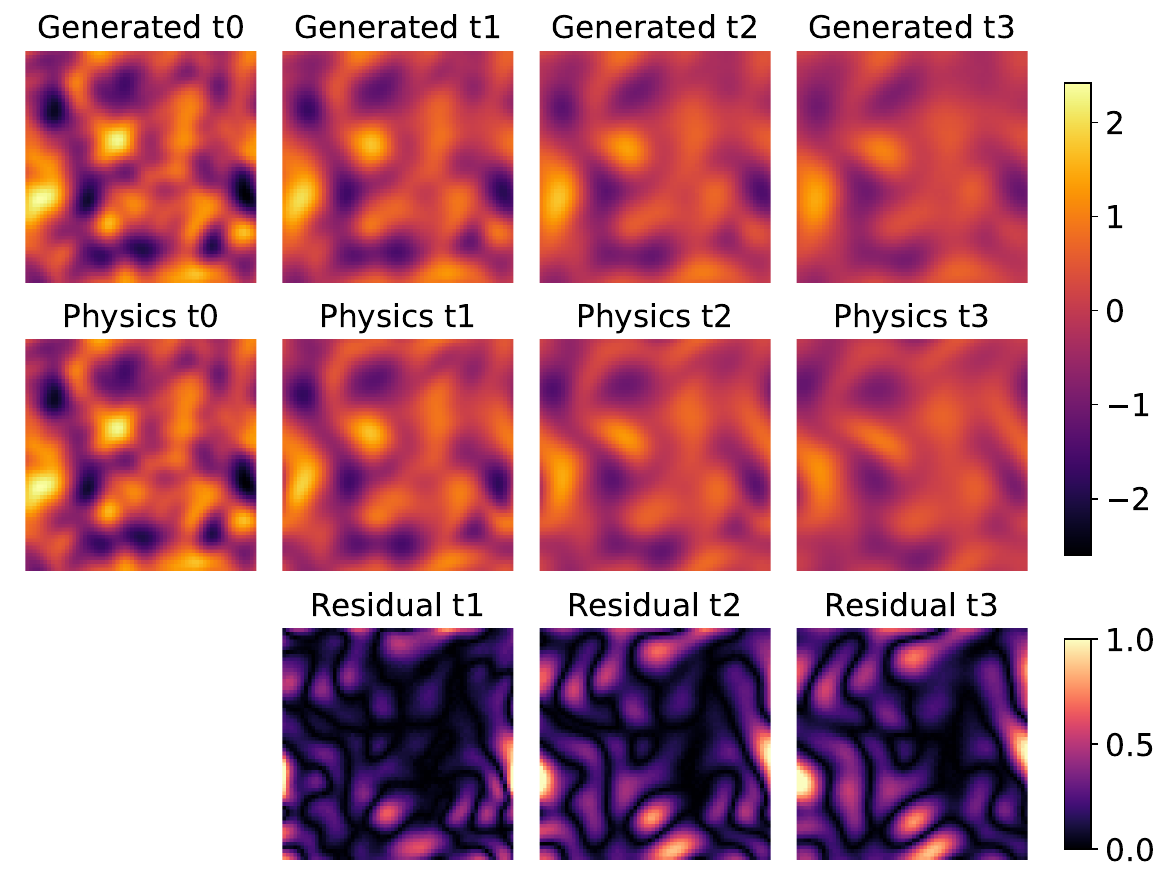}
        \caption{Standard diffusion model.}
    \end{subfigure}

    \vspace{0.6em}

    \begin{subfigure}{\linewidth}
        \centering
        \includegraphics[width=0.48\linewidth]{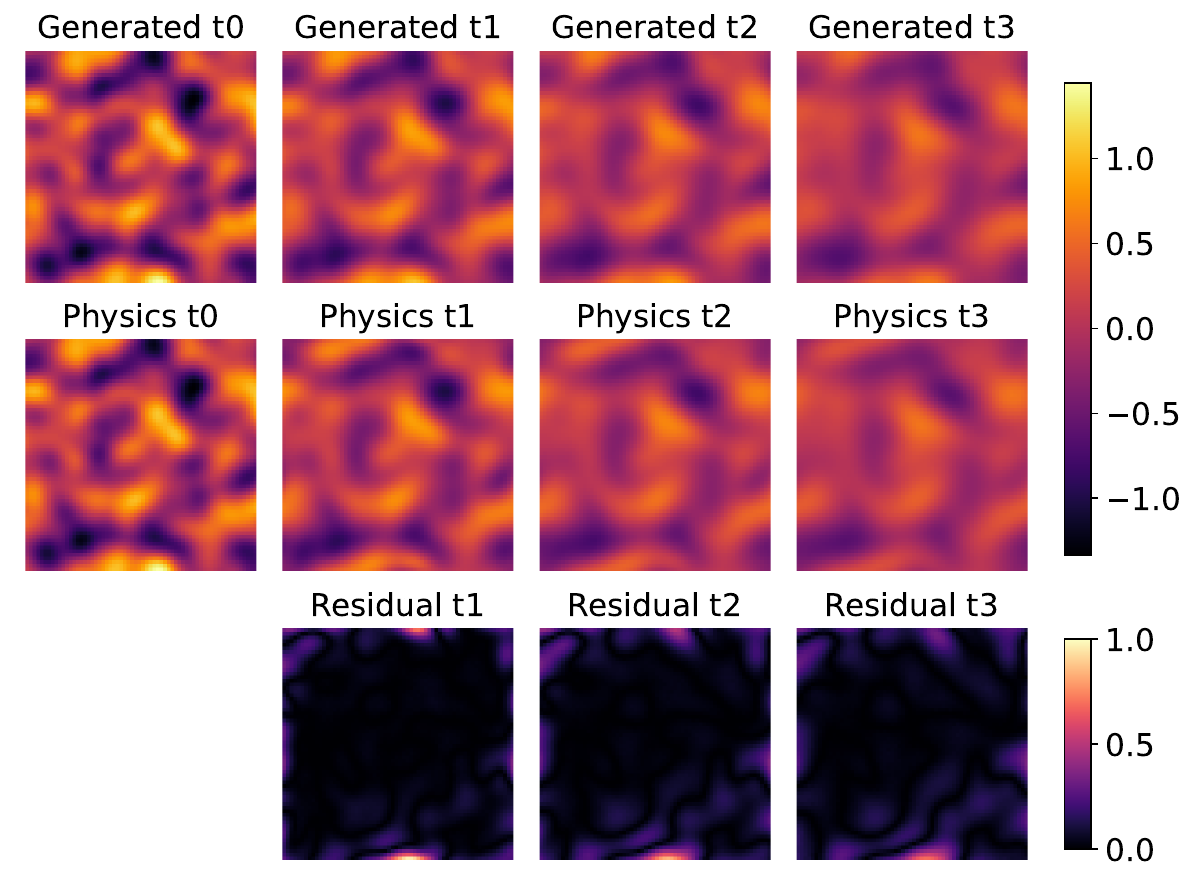}
        \hfill
        \includegraphics[width=0.48\linewidth]{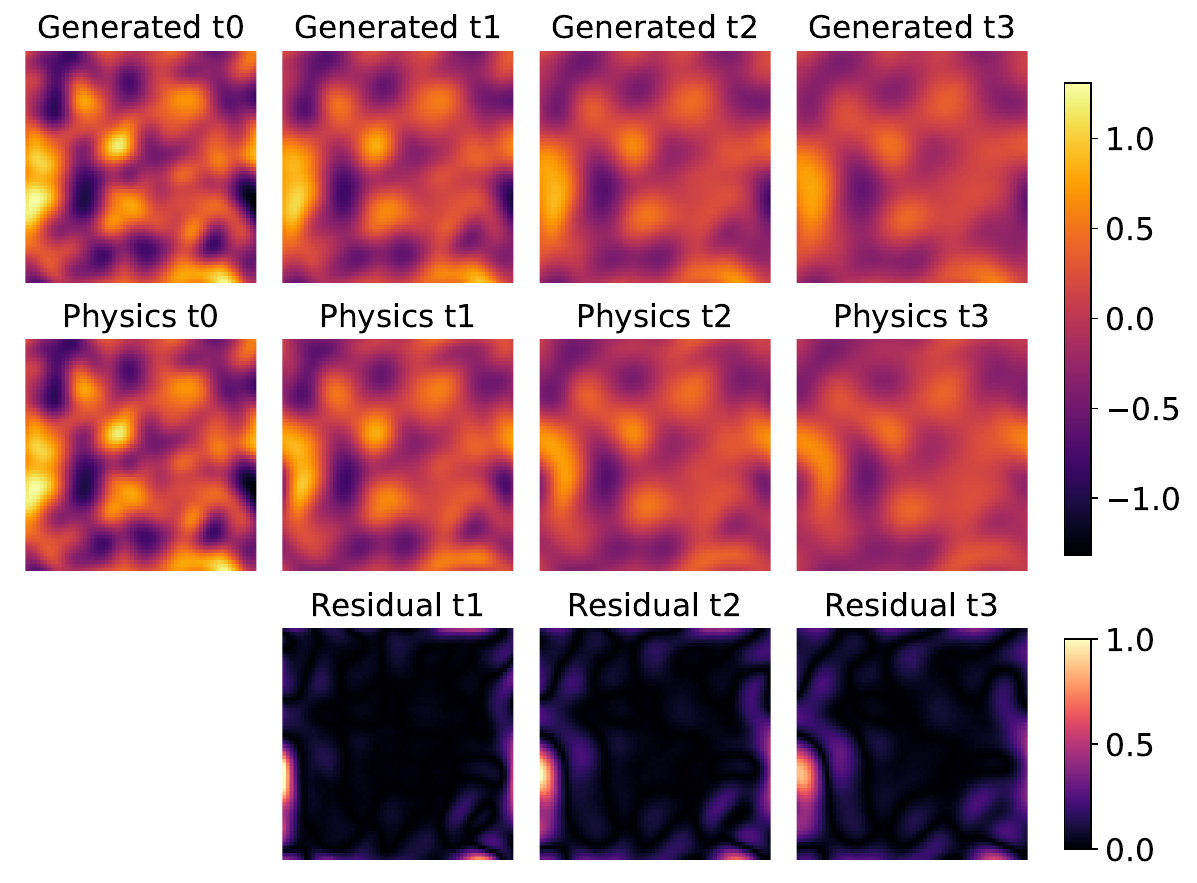}
        \caption{Proposed guidance model.}
    \end{subfigure}

    \caption{Comparison of samples and residuals generated by the standard diffusion model and the proposed guidance model in the time-series fluid experiment. The residual represents the deviation between generated samples and the corresponding snapshots obtained by numerically integrating the Navier--Stokes equations from the same initial states.}
    \label{comparison_fluid}
\end{figure}

Figure \ref{comparison_fluid} compares samples generated by the standard diffusion model and the proposed guidance model. The comparison visually demonstrates that our classifier-free guidance approach reduces the residuals.

\begin{figure}[t]
    \centering
    \includegraphics[width=\linewidth]{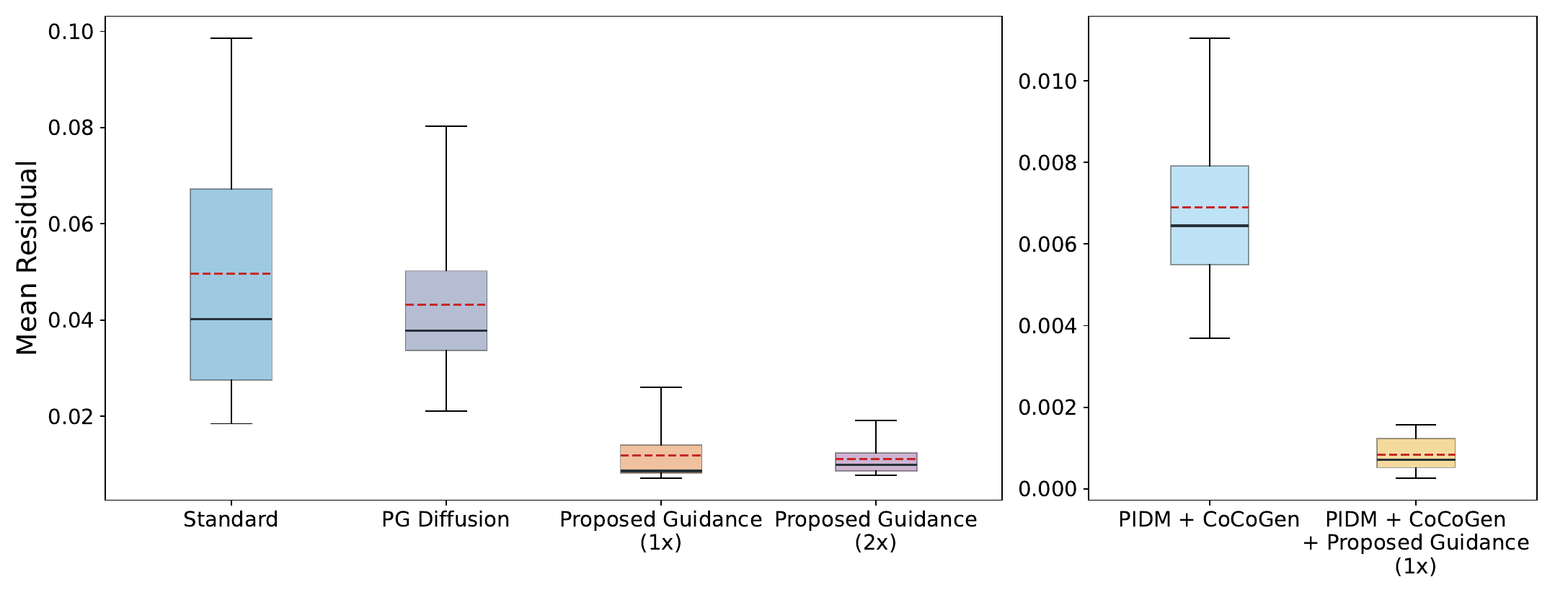}
    \caption{Distribution of the mean residuals across independent runs under different conditions in the time-series fluid experiment. The red dashed lines show the mean values. The notation 1x and 2x indicate that the proposed method is applied once and twice, respectively.}
    \label{comparison_mean_fluid}
\end{figure}

\begin{figure}[t]
    \centering

    \begin{subfigure}{\linewidth}
        \centering
        \includegraphics[width=0.48\linewidth]{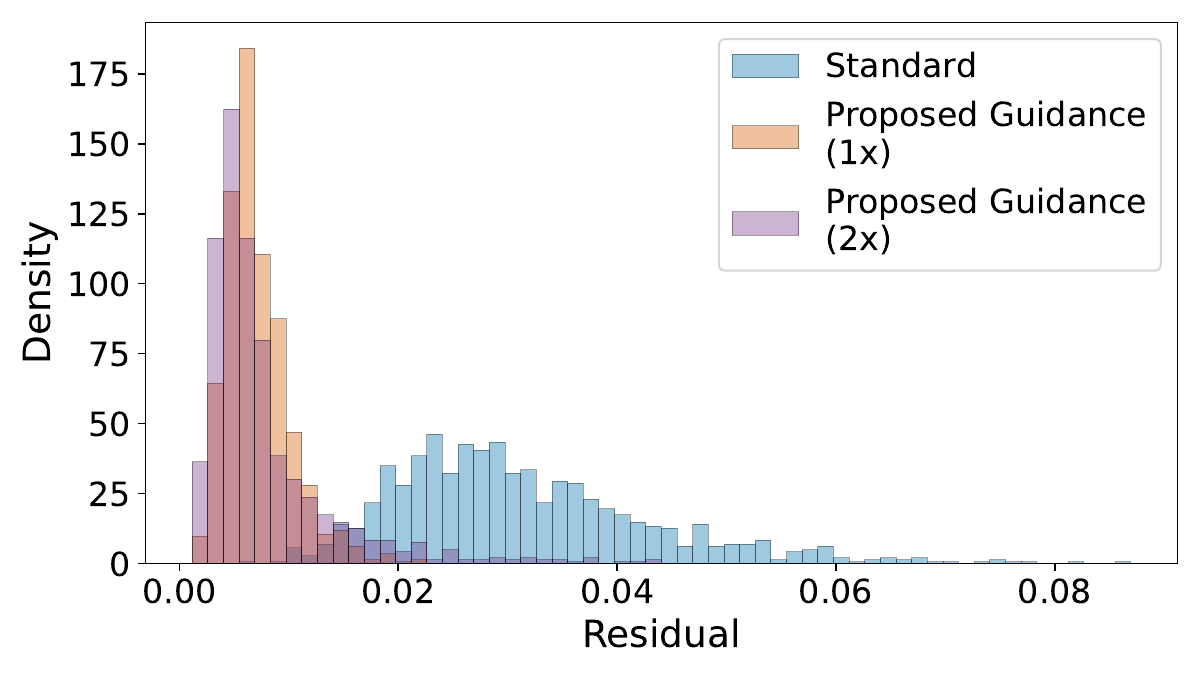}
        \hfill
        \includegraphics[width=0.48\linewidth]{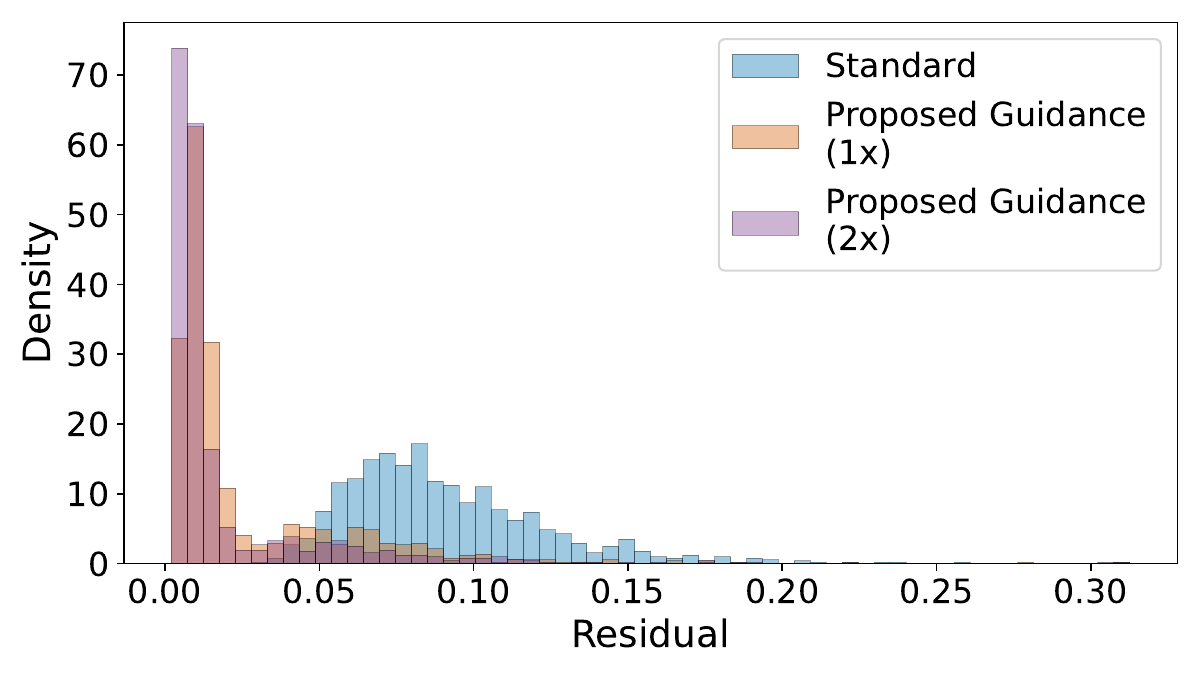}
    \end{subfigure}

    \caption{Residual distributions of 1000 generated samples in 2 different trials in the time-series fluid experiment.}
    \label{2dist_fluid}
\end{figure}

Figure \ref{comparison_mean_fluid} shows the distribution of the mean residuals over 8 trials with different random seeds. In each trial, 1000 samples were generated to calculate the mean. The trends are similar to those observed in the Darcy flow experiment, and the proposed guidance models outperform the standard diffusion and PG Diffusion baselines. Although the difference is not clearly visible in the figure, the average mean residual decreases from $1.19\times 10^{-2}$ for Proposed Guidance applied once (1x) to $1.11\times 10^{-2}$ for Proposed Guidance applied twice (2x). In addition, the largest mean residual among the trials is lower for Proposed Guidance (2x), demonstrating the effectiveness of repeated applications of the proposed method. This trend is also reflected in Figure \ref{2dist_fluid}, which shows the residual distribution within individual trials. Furthermore, combining the proposed model with PIDM and CoCoGen achieves the best performance and substantially improves from the PIDM + CoCoGen baseline. Additional experimental results are provided in Appendix \ref{ap_fluid_ex}.

\subsubsection{Comparison of computational cost}
We compare the computational cost among different sampling algorithms. We measure the time to generate 1000 samples using PG Diffusion \citep{shu2023}, CoCoGen \citep{jacobsen2025}, and our proposed diffusion guidance approach. PG Diffusion and CoCoGen repeatedly require gradient calculations during the sampling process, while the proposed approach does not. The experimental configurations used for these measurements correspond to PG Diffusion, PIDM + CoCoGen, and Proposed Guidance (1x) in Section \ref{subsec_fluid_res}.

Table \ref{tb_sampling_time} reports the average sampling time over 8 trials for each method. The results show that the proposed approach substantially reduces the sampling time compared with gradient-based PG Diffusion and CoCoGen. This reduction demonstrates the computational advantage of our simulation-free sampling strategy, which avoids repeated evaluation of physical residuals and their gradients during sampling.

We expect this advantage to become more effective for computationally demanding problems such as 3D fluid dynamics. In such problems, a single numerical simulation can require hours or days, making simulation at every sampling iteration computationally prohibitive. Our approach instead uses a model trained in advance on precomputed residuals and incorporates information about the underlying physics without performing simulations during generation, in a manner analogous to amortized inference \citep{cranmer2020, mangion2025}.

\begin{table}[t]
    \centering
    \caption{
    Comparison of the wall-clock time required to generate 1000 samples in the time-series fluid experiment. Results are reported as the mean $\pm$ standard deviation across trials.}
    \label{tb_sampling_time}
    \begin{tabular}{lc}
        \toprule
        Sampling method
        & Mean Sampling Time $\pm$ Std. [s] \\
        \midrule
        PG Diffusion & $265.9 \pm 7.4$ \\
        CoCoGen & $135.6 \pm 4.4$ \\
        Classifier-free guidance-based (proposed) & $\mathbf{66.8 \pm 0.8}$ \\
        \bottomrule
    \end{tabular}
\end{table}

\subsubsection{Data augmentation with noise}

In our approach, we first train a standard diffusion model to obtain negative samples. However, negative samples can also be constructed by adding noise to the original data and then evaluating the residuals of the noisy samples, without using diffusion training or sampling. This provides a simpler variant of our approach. Here, we conduct experiments to examine whether the first-stage diffusion training can be replaced by noise-based data augmentation.

We use the same time-series 2D fluid dataset and consider two experimental settings, with the standard diffusion model and PIDM + CoCoGen serving as the respective baselines. For each setting, we generate negative samples by adding Gaussian noise to the original data. We adjust the noise variance so that the mean residual of the resulting negative samples matches that of the corresponding baseline. For the standard diffusion setting, the mean residuals are $4.97\times 10^{-2}$ for the baseline and $4.93\times 10^{-2}$ for the noise-augmented negative samples. For the PIDM + CoCoGen setting, the corresponding values are $6.90\times 10^{-3}$ and $6.83\times 10^{-3}$, respectively.

We train the classifier-free guidance model using the noise-augmented dataset. The model parameters and training hyperparameters are kept the same as those used in the diffusion-based augmentation experiments, and we conduct 8 trials with different random seeds. Table \ref{tb_noise_aug} compares the mean residuals calculated over 1000 generated samples. Although noise augmentation also reduces the residuals, the diffusion-based augmentation adopted in our method results in lower mean residuals. These results indicate that augmentation with random noise provides a potential alternative, particularly when training the first-stage diffusion model is prohibitively expensive. Nevertheless, when computationally feasible, first-stage diffusion training is preferable because employing visually plausible negative samples tends to provide better physical consistency.

\begin{table}[t]
    \centering
    \caption{
    Comparison of mean residuals between models augmented with Gaussian noise and with diffusion-generated samples. Results across trials are reported as the mean $\pm$ standard deviation and the median with the interquartile range shown in parentheses.}
    \label{tb_noise_aug}
    \begin{tabular}{lcc}
        \toprule
        Experimental condition
        & Mean Residual $\pm$ Std.
        & Median Residual (IQR) \\
        \midrule
        Standard Diffusion (baseline) & $4.97\times10^{-2} \pm 2.90\times10^{-2}$ & $4.02\times10^{-2}\ (3.98\times10^{-2})$ \\
        Noise Augmentation (guidance) & $1.28\times10^{-2} \pm 7.92\times10^{-4}$ & $1.31\times10^{-2}\ (8.95\times10^{-4})$ \\
        Diffusion-Based Augmentation (guidance) & $\mathbf{1.19\times10^{-2} \pm 5.96\times10^{-3}}$ & $\mathbf{8.64\times10^{-3}\ (5.84\times10^{-3})}$ \\
        \midrule
        PIDM + CoCoGen (baseline) & $6.90\times10^{-3} \pm 2.23\times10^{-3}$ & $6.45\times10^{-3}\ (2.42\times10^{-3})$ \\
        Noise Augmentation (guidance) & $1.07\times10^{-3} \pm 9.89\times10^{-5}$ & $1.08\times10^{-3}\ (1.64\times10^{-4})$ \\
        Diffusion-Based Augmentation (guidance) & $\mathbf{8.43\times10^{-4} \pm 4.64\times10^{-4}}$ & $\mathbf{7.11\times10^{-4}\ (7.15\times10^{-4})}$ \\
        \bottomrule
    \end{tabular}
\end{table}

\section{Conclusion}

In this study, we proposed a method to enhance the physical consistency of diffusion models using classifier-free guidance with self-generated data augmentation. The proposed model learns a distribution conditioned on residuals and guides the generation process toward samples with lower deviations from physical laws by setting the residual condition to zero. In addition, our approach decouples physics simulation from diffusion training and sampling, avoiding repeated and time-consuming constraint evaluation as well as gradient calculations of the constraints.

Experimental results demonstrate that the proposed method reduces residuals compared with standard diffusion models such as DDPM, and further improves physical consistency when combined with existing physics-constrained training and sampling approaches. Moreover, our empirical results show that multiple application of the proposed method yields additional improvements in physical consistency without requiring evaluation of the governing equations during diffusion training or sampling. This property broadens the applicability of physics-aware diffusion models to computationally demanding problems.

The proposed framework is applicable to a wide range of constrained generation tasks, provided that consistency with a given set of rules can be evaluated. Such rules are not limited to the physical laws considered in this study. For example, the method could be applied to generating control inputs for robotic systems under motion constraints or generating robot trajectories that comply with operational rules. Exploring such extensions remains an important direction for future work.

\subsubsection*{Acknowledgments}
This work was supported by JST PRESTO JPMJPR24T6, JSPS KAKENHI JP25H01454, JP26K02968, and JST SPRING JPMJSP2108.

\bibliography{references_updated}
\bibliographystyle{tmlr}

\appendix
\section{Details of the Darcy flow experiment}\label{ap_darcy_set}
\subsection{Dataset}

\begin{figure}[t]
    \centering
    \begin{subfigure}{0.46\linewidth}
        \centering
        \includegraphics[width=\linewidth]{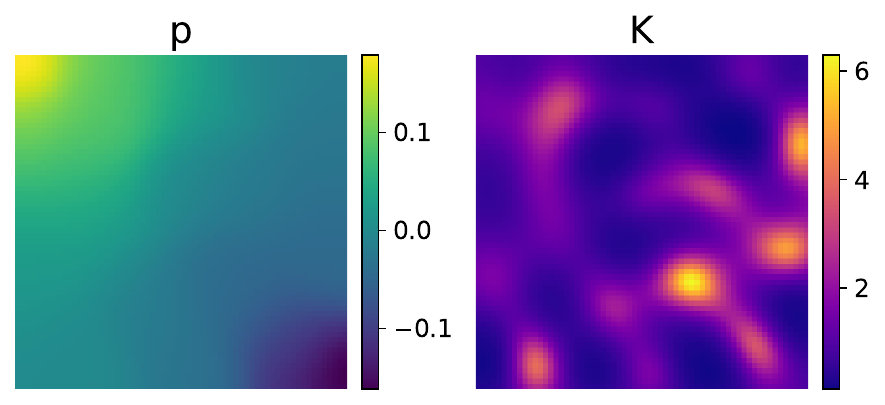}
    \end{subfigure}
    \hfill
    \begin{subfigure}{0.46\linewidth}
        \centering
        \includegraphics[width=\linewidth]{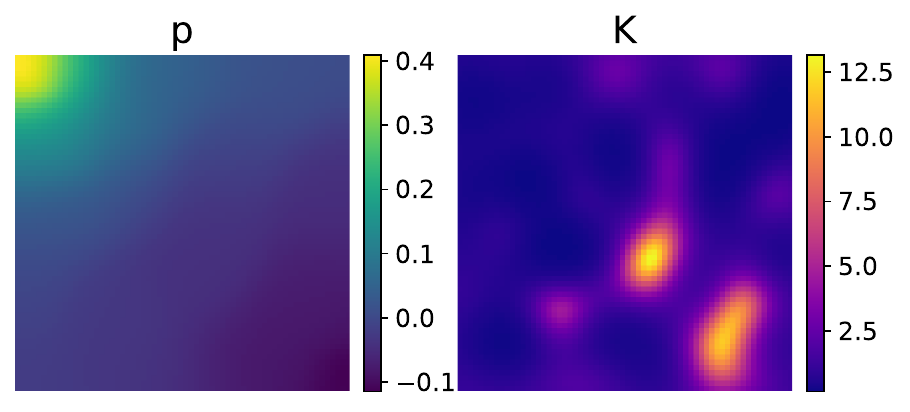}
    \end{subfigure}

    \vspace{0.2em}

    \begin{subfigure}{0.46\linewidth}
        \centering
        \includegraphics[width=\linewidth]{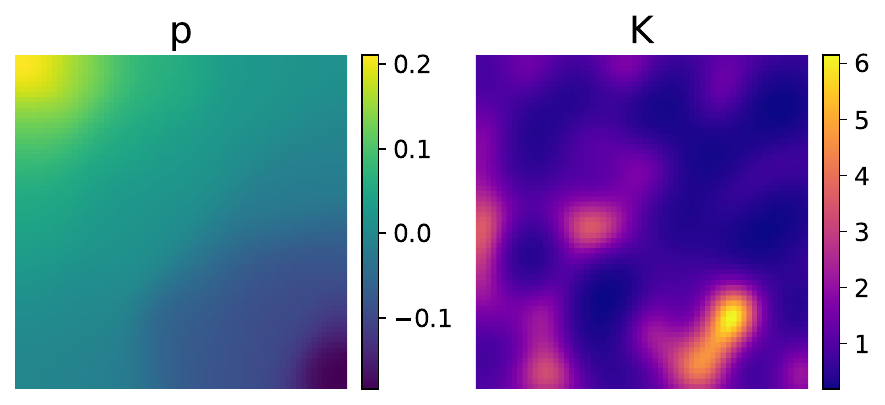}
    \end{subfigure}
    \hfill
    \begin{subfigure}{0.46\linewidth}
        \centering
        \includegraphics[width=\linewidth]{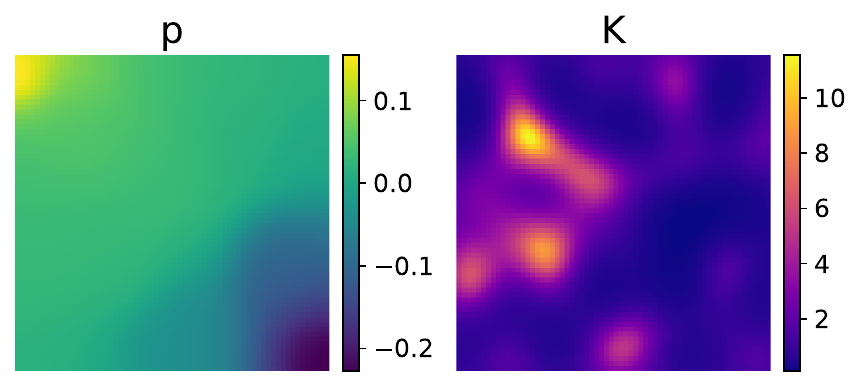}
    \end{subfigure}

    \caption{Examples from the Darcy flow dataset.}
    \label{darcy_data_samples}
\end{figure}

The Darcy flow dataset released by \citet{bastek2025} consists of 10000 training samples and 1000 validation samples. Each sample contains the pressure field $p$ and permeability field $K$, both at a resolution of $64\times 64$. We concatenate $p$ and $K$ along the channel dimension so that each sample is represented as $x \in \mathbb{R}^{2\times 64\times 64}$. Figure \ref{darcy_data_samples} shows examples from the dataset.

\subsection{Implementation details}

The proposed method was implemented based on the code provided for Denoising Diffusion Implicit Models (DDIM) \citep{song2021b}. All experiments were conducted on an NVIDIA GH200 GPU for both training and sampling.

The pressure and permeability channels were independently normalized to $[-1,1]$ using the minimum and maximum values computed from the corresponding training set. Before residual evaluation, generated samples were transformed back to their original physical scales.

Following \citet{song2021b}, we use a U-Net architecture \citep{ronneberger2015} to estimate the added noise. The model parameters are optimized using Adam \citep{kingma2017}. The residual $r$ is input to the model as a condition vector $c=[r,1] \in \mathbb{R}^2$, where the second component distinguishes a valid condition from the null condition used for classifier-free guidance. The condition vector is embedded with an encoder neural network, and added to residual blocks of the U-Net. During classifier-free guidance training, the condition is replaced with the null condition $c_{\varnothing}=[-1,0]$ with probability $0.1$. During sampling, the zero-residual condition is represented by $c=[0,1]$.

Table \ref{tb_unet_darcy} summarizes the model and experimental settings. The same model configuration is used for both the first-stage diffusion model and the subsequent guidance model.

\begin{table}[t]
    \centering
    \small
    \caption{Experimental settings for the Darcy flow experiments.}
    \label{tb_unet_darcy}
    \begin{tabular}{ll}
        \toprule
        Parameter & Setting \\
        \midrule
        \multicolumn{2}{l}{\textbf{Model}} \\
        Architecture & U-Net \\
        Base channels & 64 \\
        Channel multipliers & $(1,2,4,8)$ \\
        Residual blocks & 2 per resolution \\
        Attention & $16\times16$ resolution \\
        Dropout & 0.1 \\
        EMA decay & 0.9999 \\

        \midrule
        \multicolumn{2}{l}{\textbf{Diffusion}} \\
        Number of diffusion steps & 1000 \\
        Variance schedule & Linear ($1\times 10^{-4}\rightarrow2\times10^{-2}$) \\

        \midrule
        \multicolumn{2}{l}{\textbf{Training}} \\
        Optimizer & Adam \\
        Learning rate & $1.0\times10^{-4}$ \\
        Batch size & 64 \\
        Epochs & 500 \\
        Weight decay & 0.0 \\
        Gradient clipping & 1.0 \\
        Classifier-free dropout & 0.1 \\

        \midrule
        \multicolumn{2}{l}{\textbf{Sampling}} \\
        Sampling steps & 100 \\
        DDIM stochasticity parameter $\eta$ & 1.0 \\
        Guidance scale $w$ & 2.0 \\
        \bottomrule
    \end{tabular}
\end{table}

\section{Additional results of the Darcy flow experiment}\label{ap_darcy_ex}
\subsection{Residual distributions of individual trials}

\begin{figure}[t]
    \centering

    \begin{subfigure}{\linewidth}
        \centering
        \includegraphics[width=0.48\linewidth]{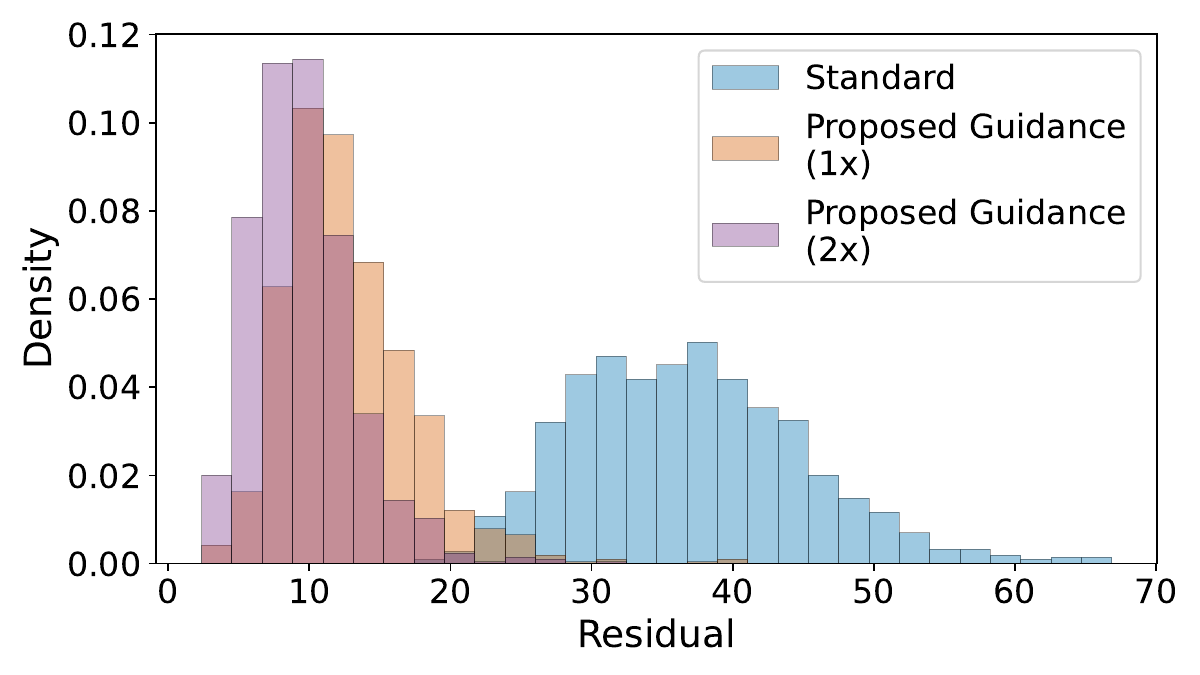}
        \hfill
        \includegraphics[width=0.48\linewidth]{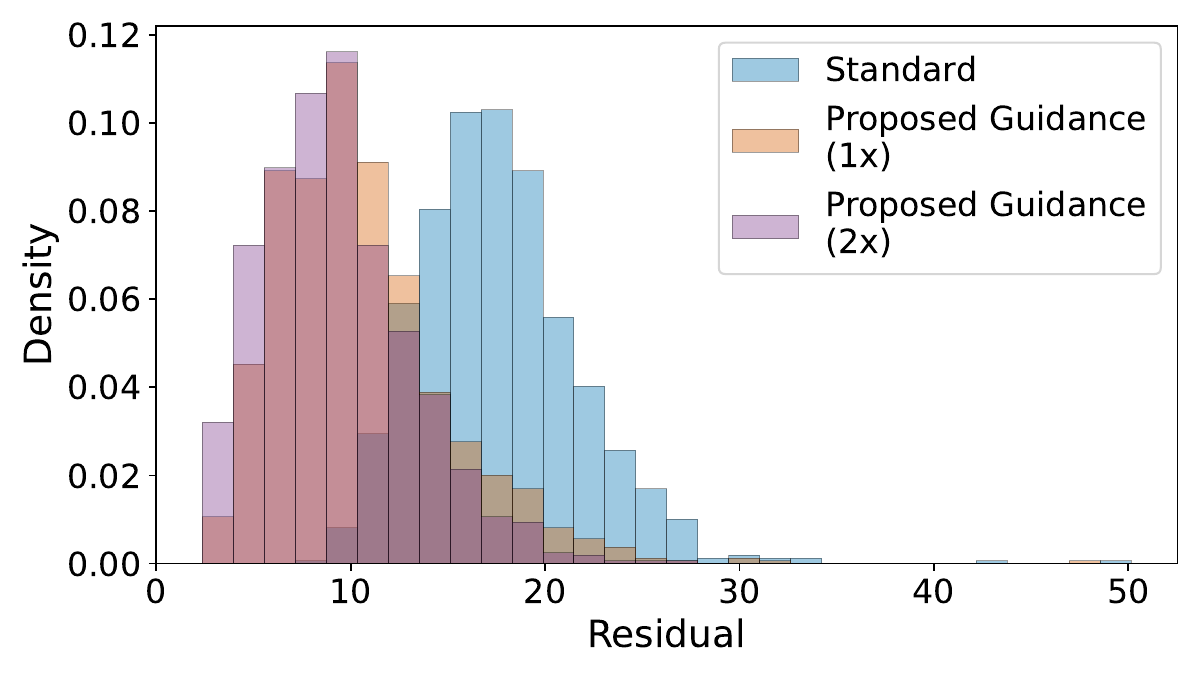}
    \end{subfigure}

    \vspace{0.5em}

    \begin{subfigure}{\linewidth}
        \centering
        \includegraphics[width=0.48\linewidth]{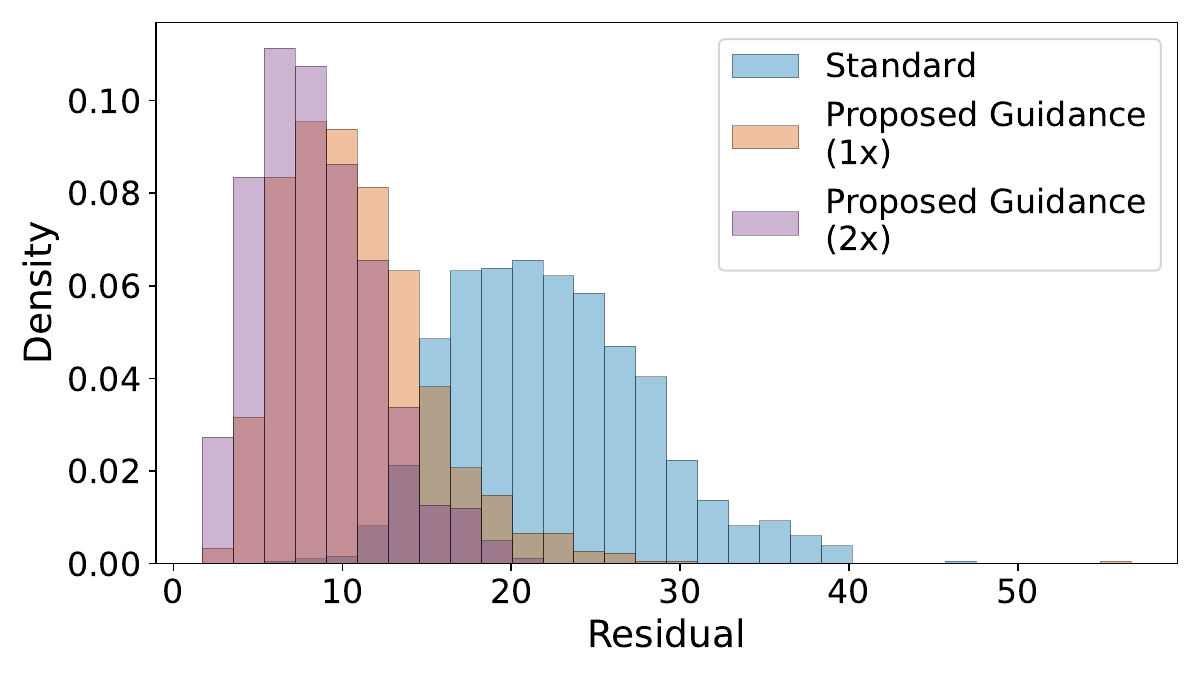}
        \hfill
        \includegraphics[width=0.48\linewidth]{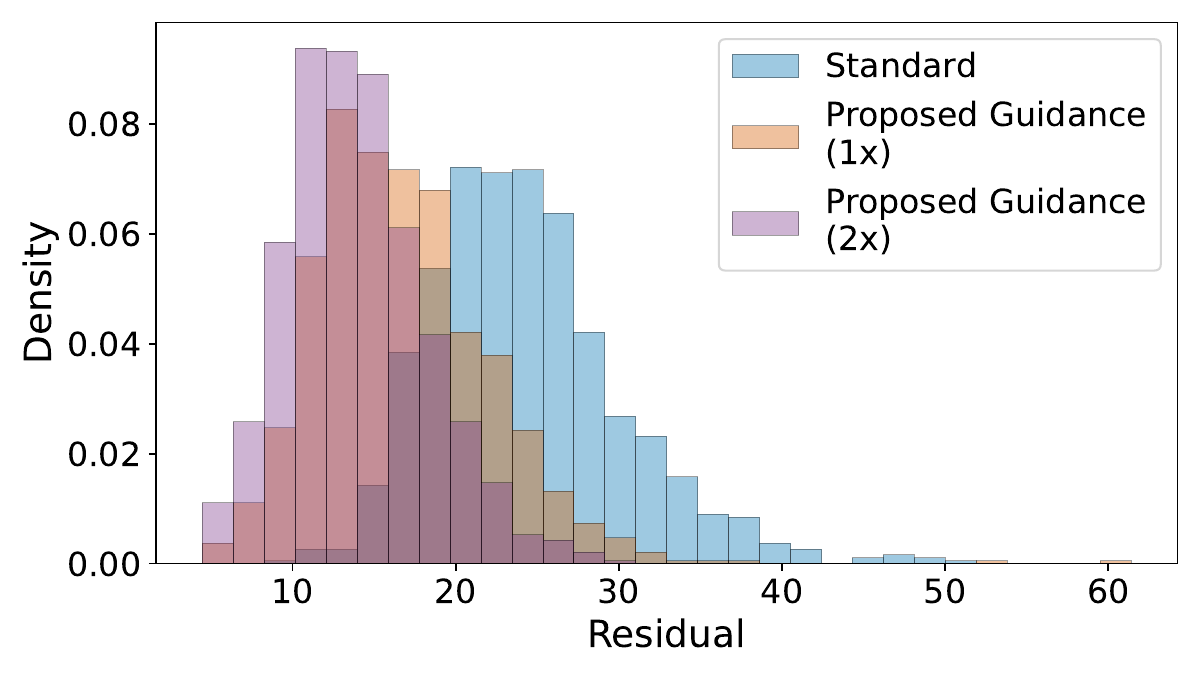}
    \end{subfigure}

    \vspace{0.5em}

    \begin{subfigure}{\linewidth}
        \centering
        \includegraphics[width=0.48\linewidth]{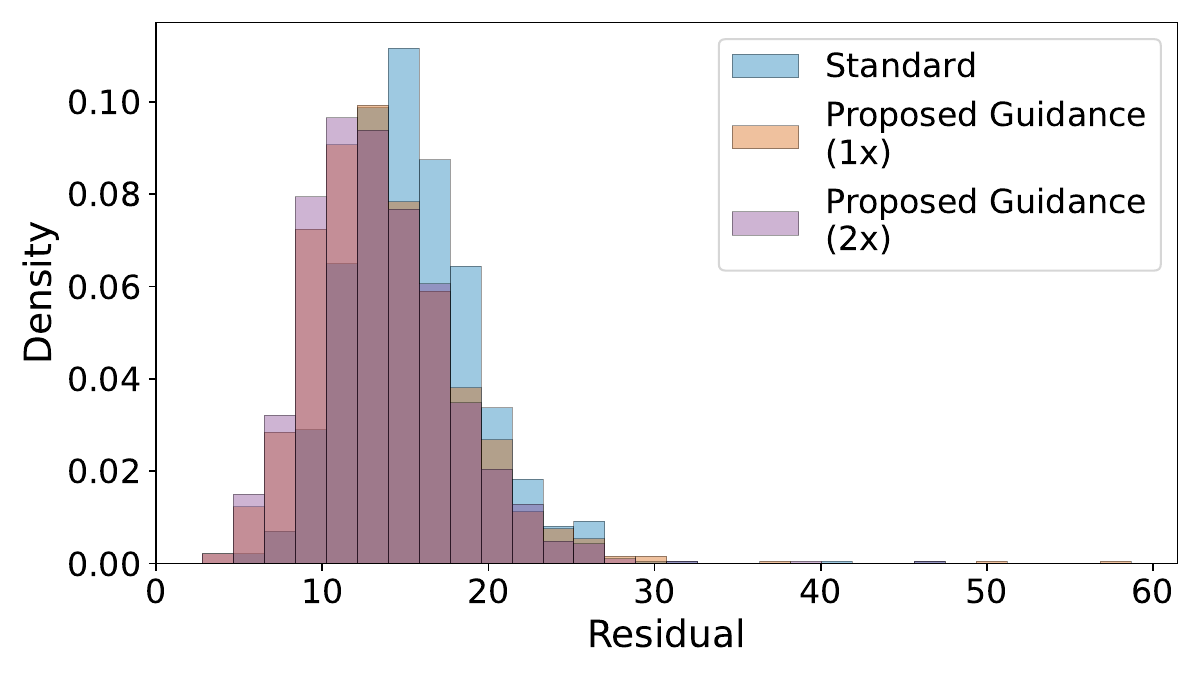}
        \hfill
        \includegraphics[width=0.48\linewidth]{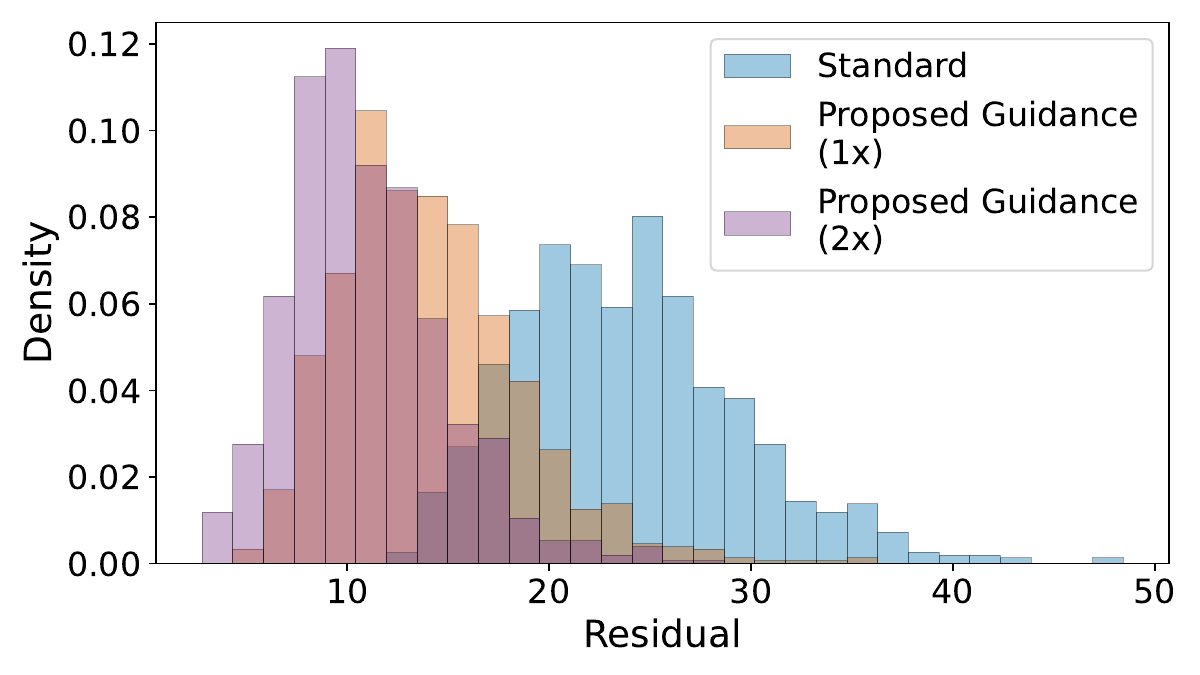}
    \end{subfigure}
    
    \caption{Residual distributions over 1000 generated samples for 6 independent trials in the Darcy flow experiment. Each histogram corresponds to a different random seed.}
    \label{ap_sample_dist_darcy}
\end{figure}

Figure \ref{sample_dist_darcy} in Section \ref{sec_exp_darcy} shows the residual distributions for 2 of the 8 trials, while Figure \ref{ap_sample_dist_darcy} presents those for the remaining 6 trials. Each histogram is computed from 1000 samples generated in a single independent trial with a different random seed. In all trials, the distributions shift toward lower residual values when the proposed guidance is applied. The distributions shift further toward zero when the proposed method is applied twice by adding the samples generated by the first guidance model to the existing training data and retraining the guidance model. These results are consistent with the findings reported in Section \ref{sec_exp_darcy}.

\subsection{Generated samples}

\begin{figure}[t]
    \centering

    \begin{subfigure}{\linewidth}
        \centering
        \includegraphics[width=0.48\linewidth]{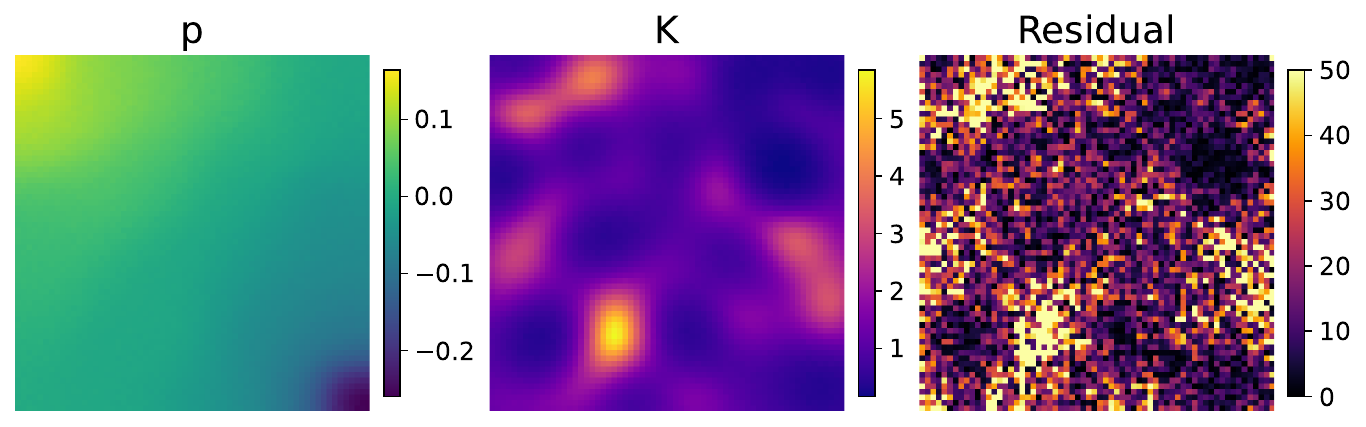}
        \hfill
        \includegraphics[width=0.48\linewidth]{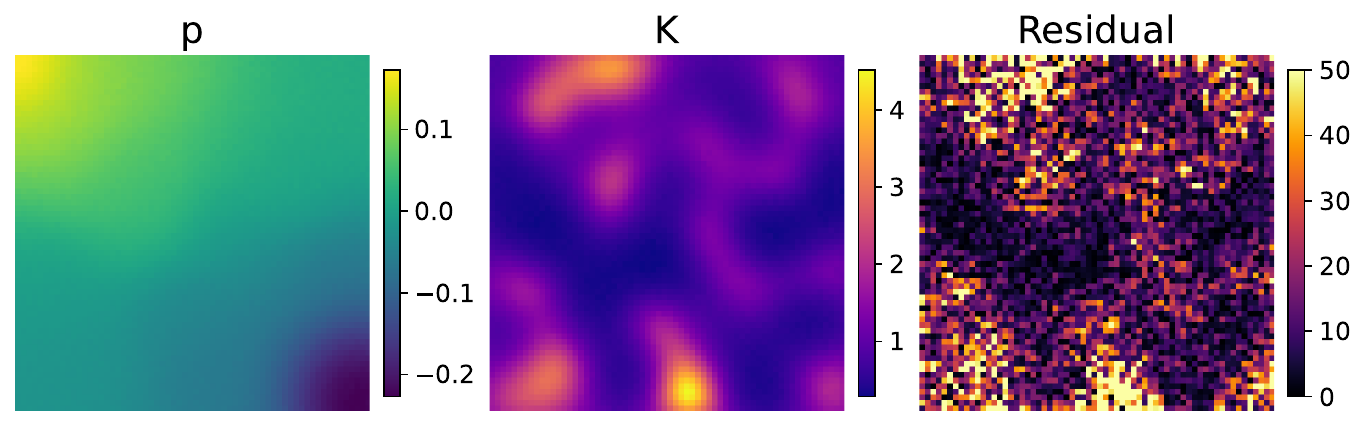}
        \caption{Standard diffusion model.}
    \end{subfigure}

    \vspace{0.6em}

    \begin{subfigure}{\linewidth}
        \centering
        \includegraphics[width=0.48\linewidth]{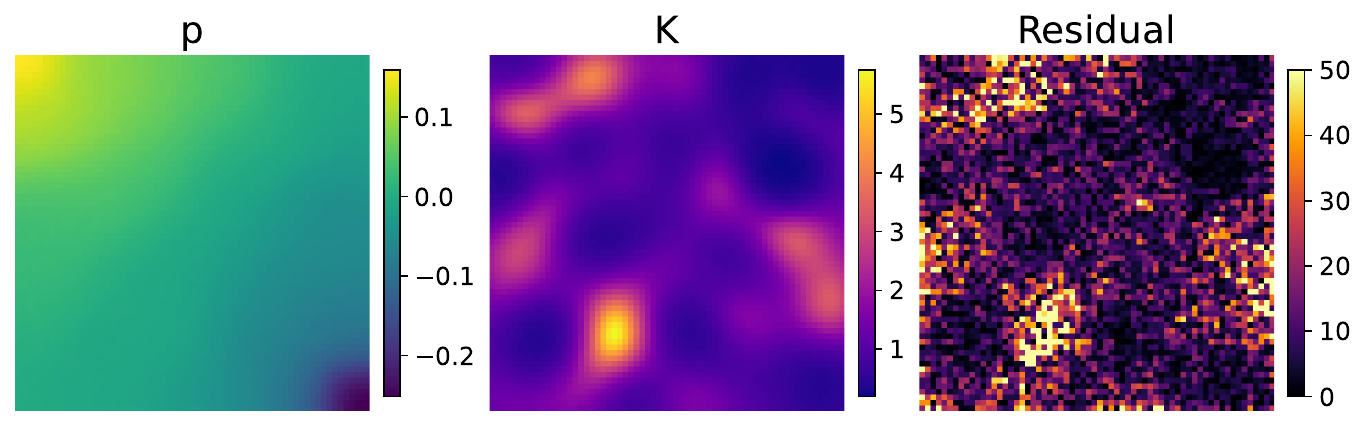}
        \hfill
        \includegraphics[width=0.48\linewidth]{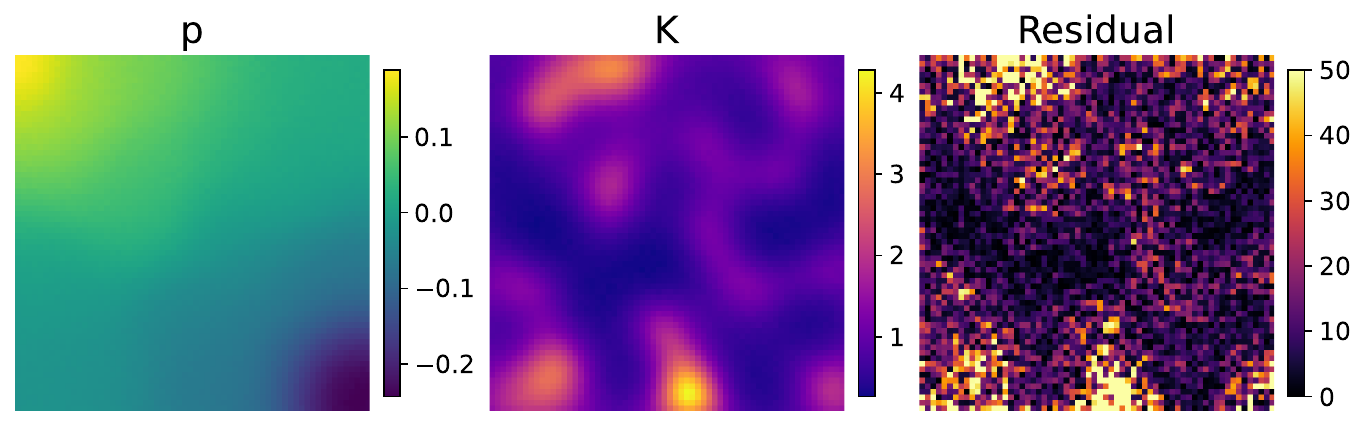}
        \caption{PG Diffusion.}
    \end{subfigure}

    \vspace{0.6em}

    \begin{subfigure}{\linewidth}
        \centering
        \includegraphics[width=0.48\linewidth]{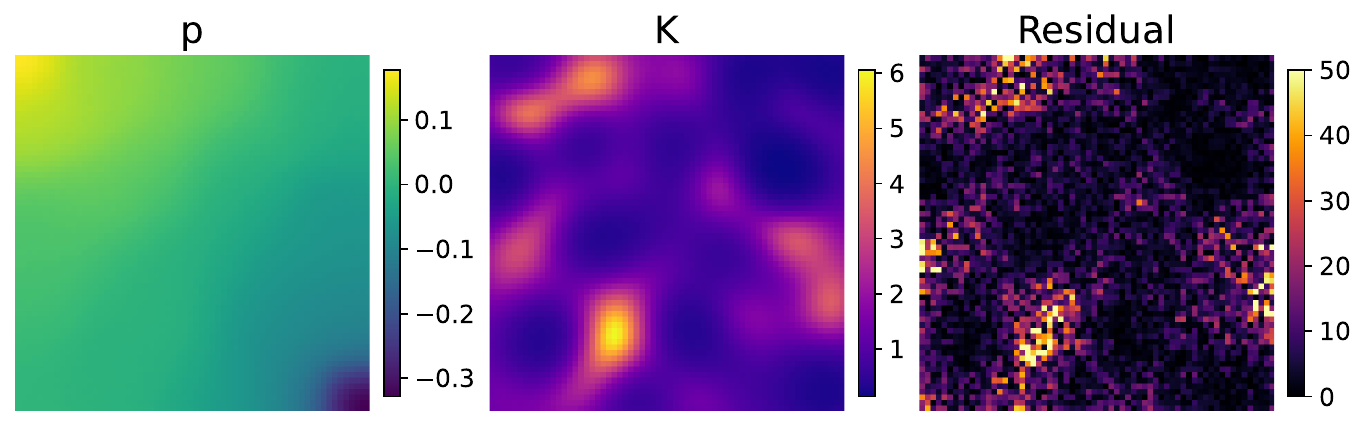}
        \hfill
        \includegraphics[width=0.48\linewidth]{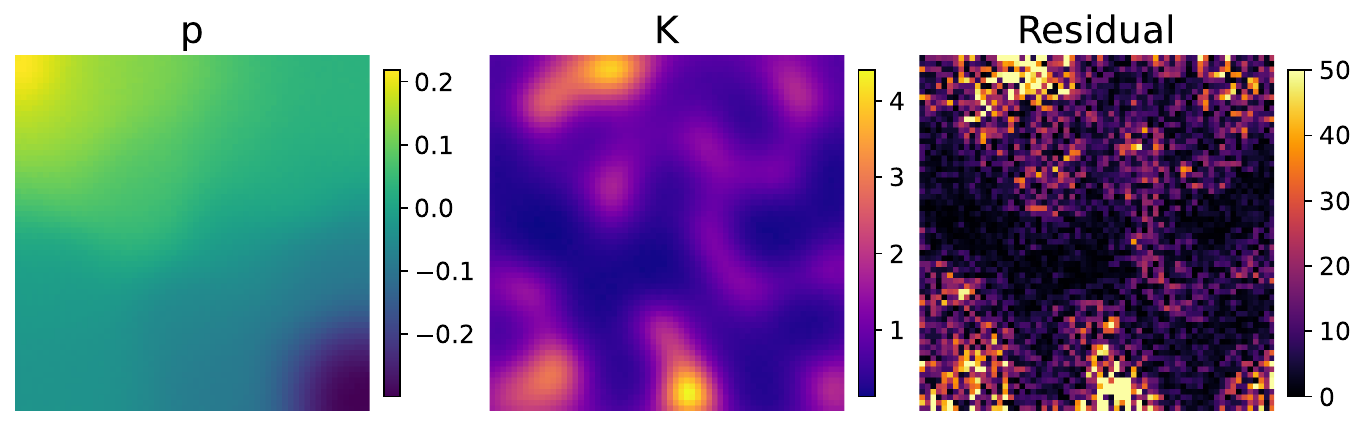}
        \caption{Proposed guidance model applied once.}
    \end{subfigure}

    \vspace{0.6em}

    \begin{subfigure}{\linewidth}
        \centering
        \includegraphics[width=0.48\linewidth]{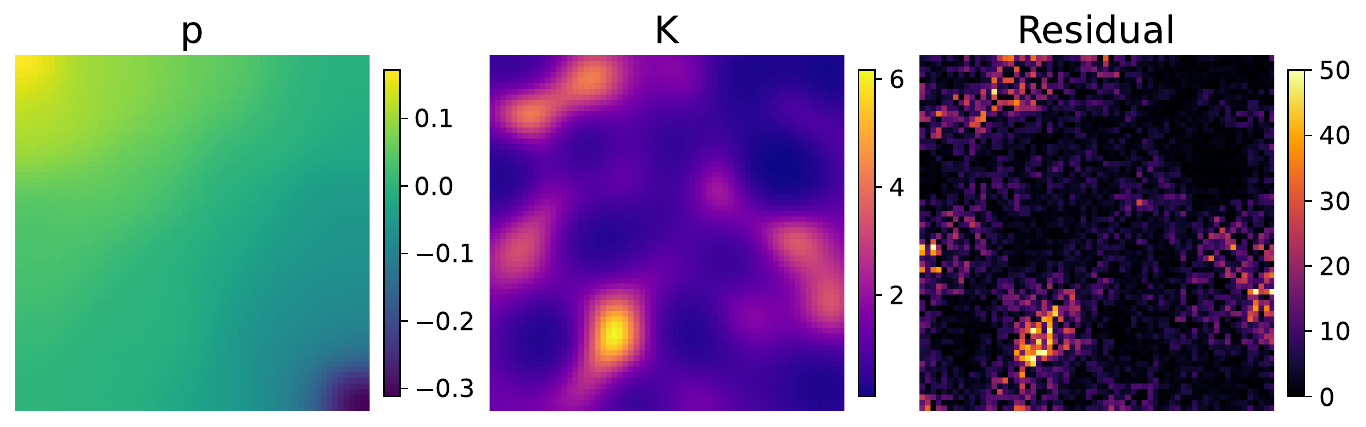}
        \hfill
        \includegraphics[width=0.48\linewidth]{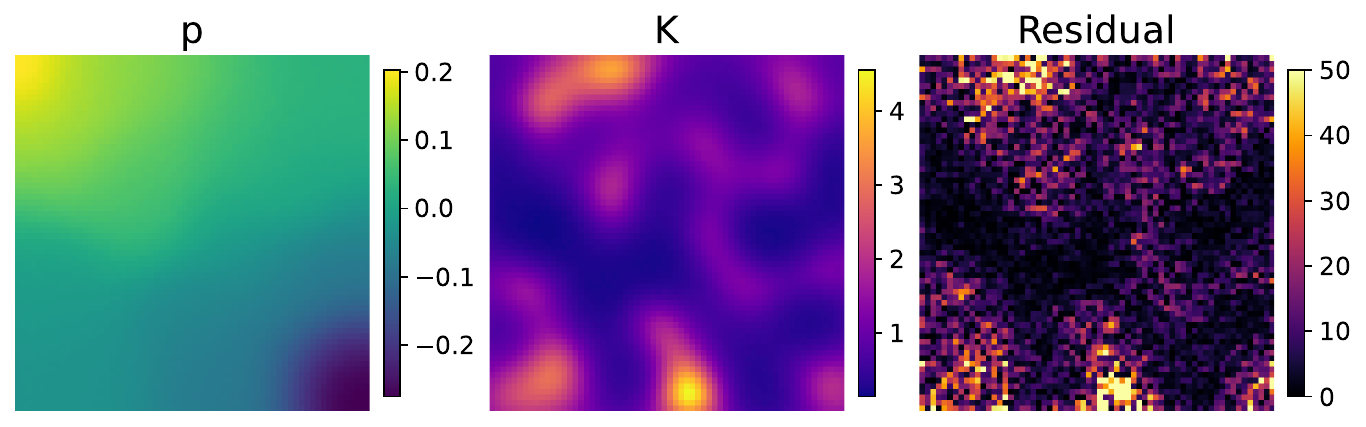}
        \caption{Proposed guidance model applied twice.}
    \end{subfigure}

    \vspace{0.6em}

    \begin{subfigure}{\linewidth}
        \centering
        \includegraphics[width=0.48\linewidth]{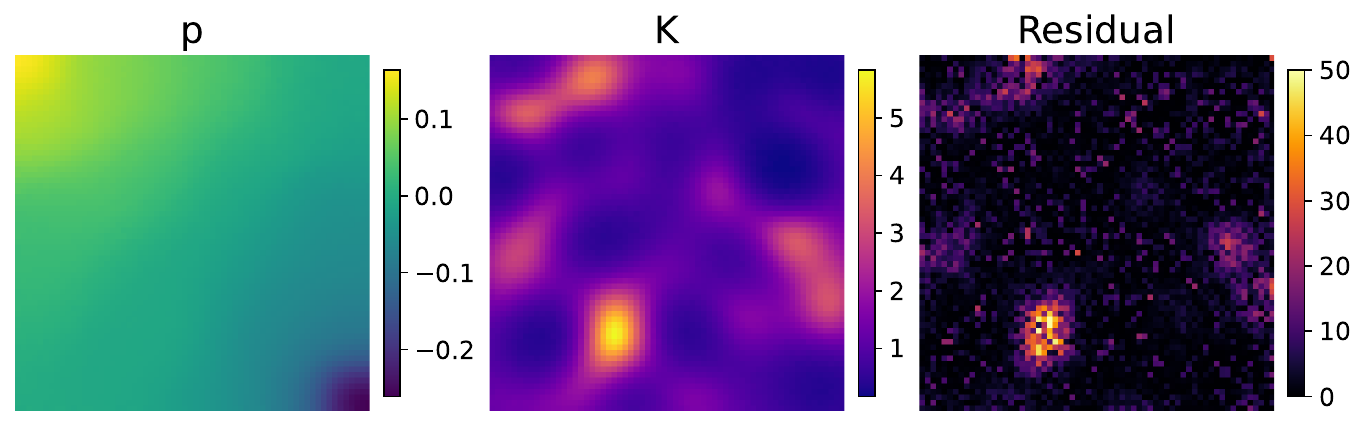}
        \hfill
        \includegraphics[width=0.48\linewidth]{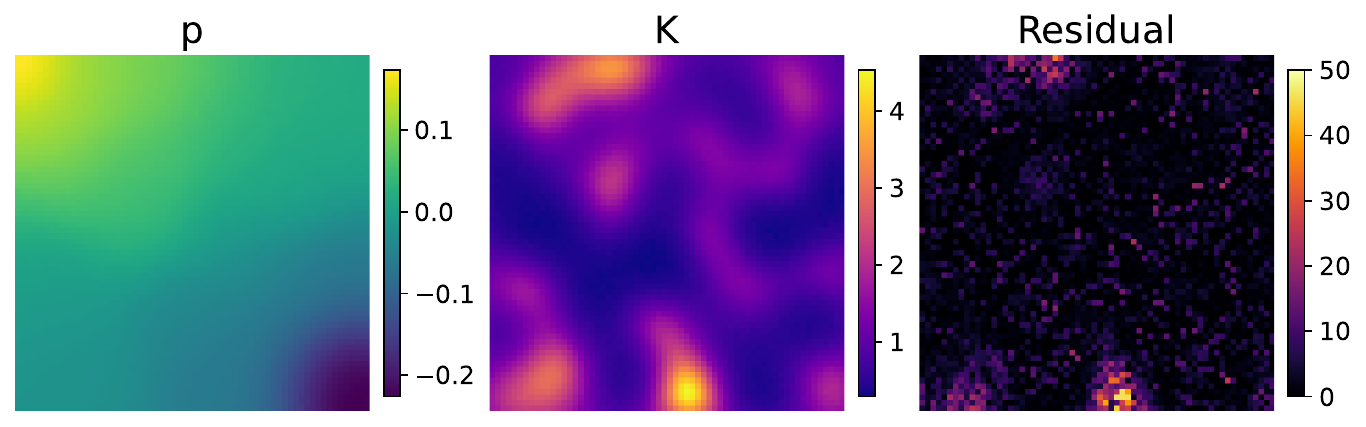}
        \caption{PIDM + CoCoGen.}
    \end{subfigure}

    \vspace{0.6em}

    \begin{subfigure}{\linewidth}
        \centering
        \includegraphics[width=0.48\linewidth]{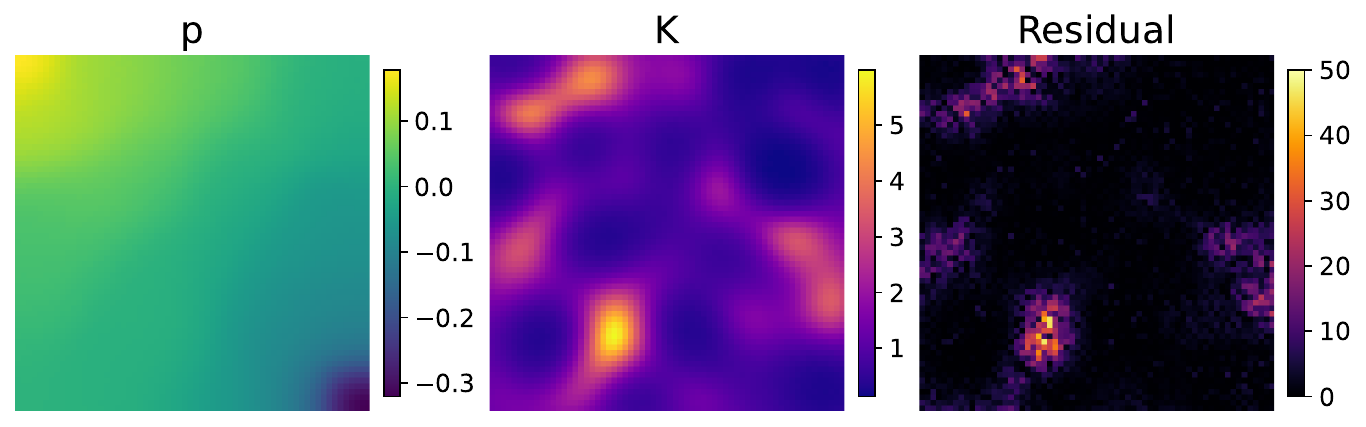}
        \hfill
        \includegraphics[width=0.48\linewidth]{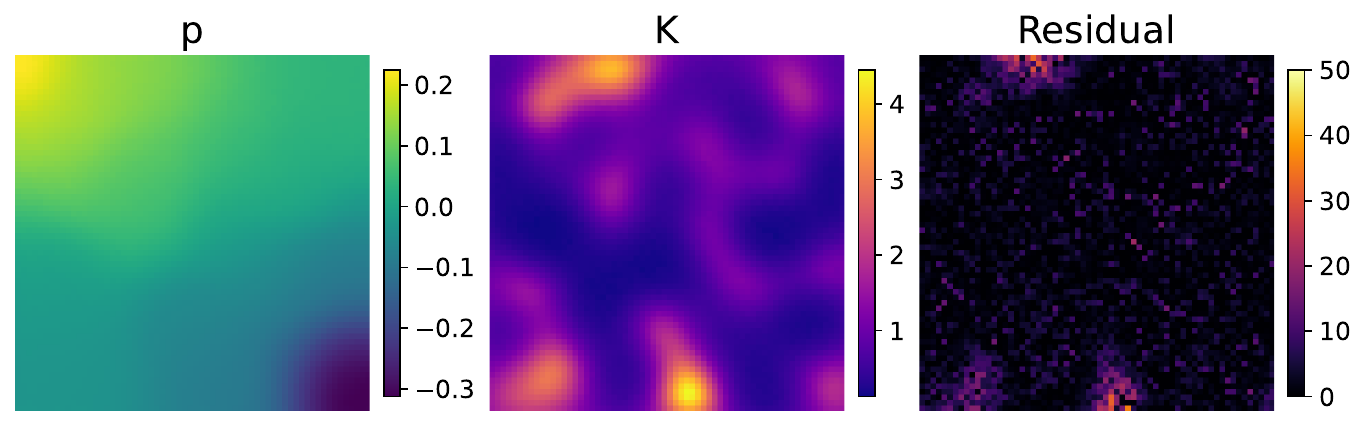}
        \caption{PIDM + CoCoGen with the proposed guidance applied once.}
    \end{subfigure}
    
    \caption{Comparison of samples and residuals generated under different conditions in the Darcy flow experiment.}
    \label{ap_comparison_darcy}
\end{figure}

Figure \ref{ap_comparison_darcy} shows samples generated under different experimental conditions. The residual represents the PDE residual of the Darcy flow equation. 

It illustrates how the proposed method and other physics-informed diffusion methods reduce the residuals. It also demonstrates the proposed method combined with PIDM and CoCoGen achieves the largest reduction.

\subsection{Novelty check}

We compare the generated samples with the training data to assess whether the reduction in residuals could be attributed to memorization of training samples that satisfy the governing equations. This analysis examines whether the model can generate novel, physically consistent samples rather than simply reproducing the training data.

\begin{figure}[t]
    \centering

    \begin{subfigure}{12cm}
        \centering
        \includegraphics[width=\linewidth]{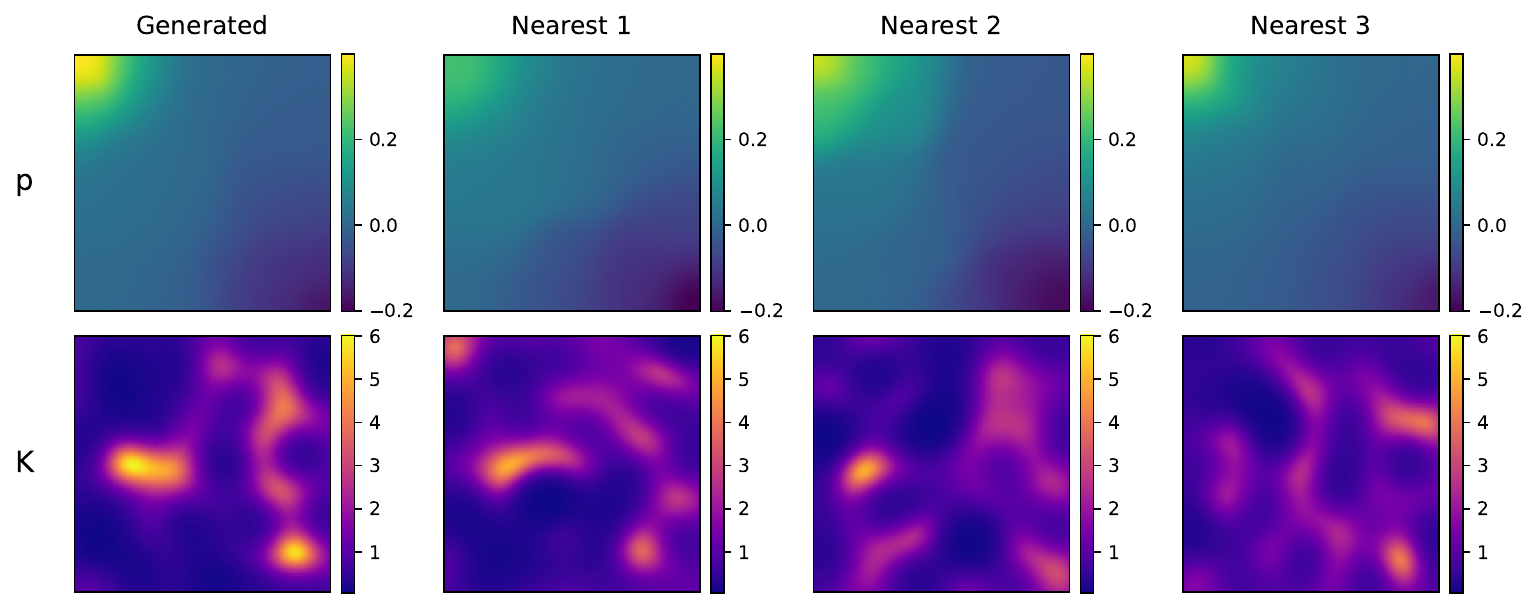}
    \end{subfigure}

    \vspace{0.1em}

    \begin{subfigure}{12cm}
        \centering
        \includegraphics[width=\linewidth]{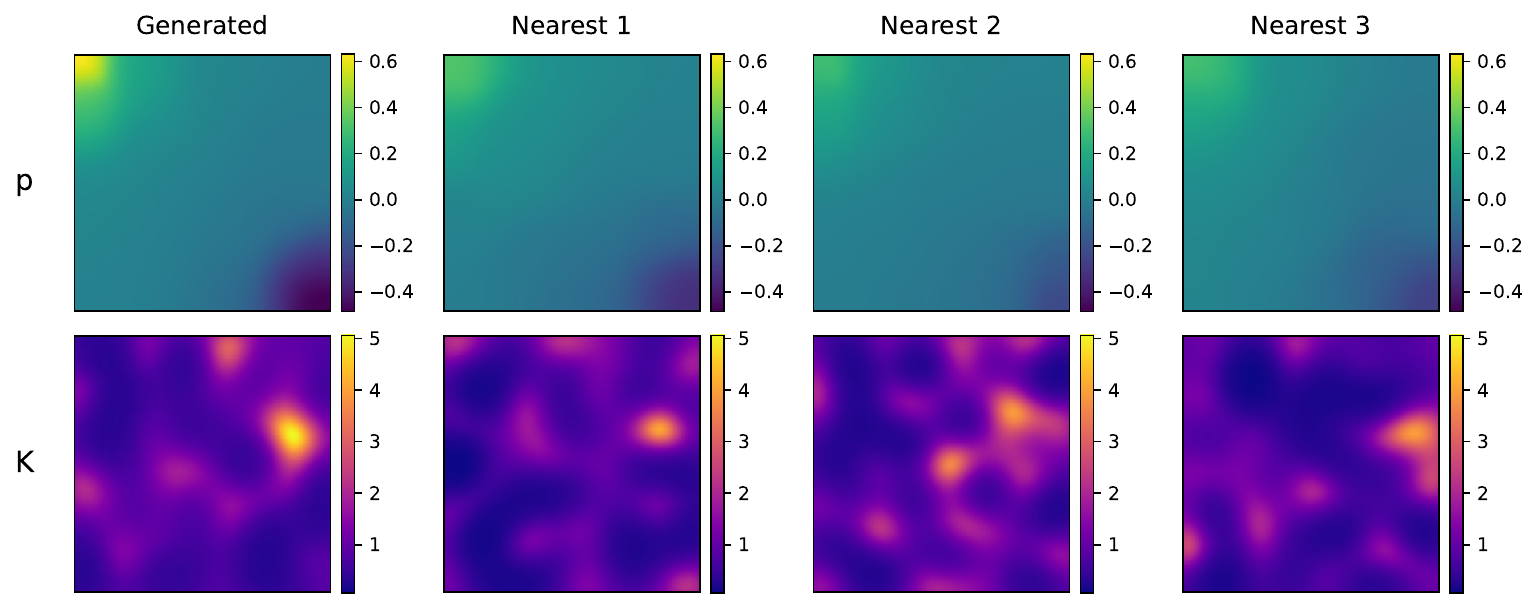}
    \end{subfigure}

    \vspace{0.1em}

    \begin{subfigure}{12cm}
        \centering
        \includegraphics[width=\linewidth]{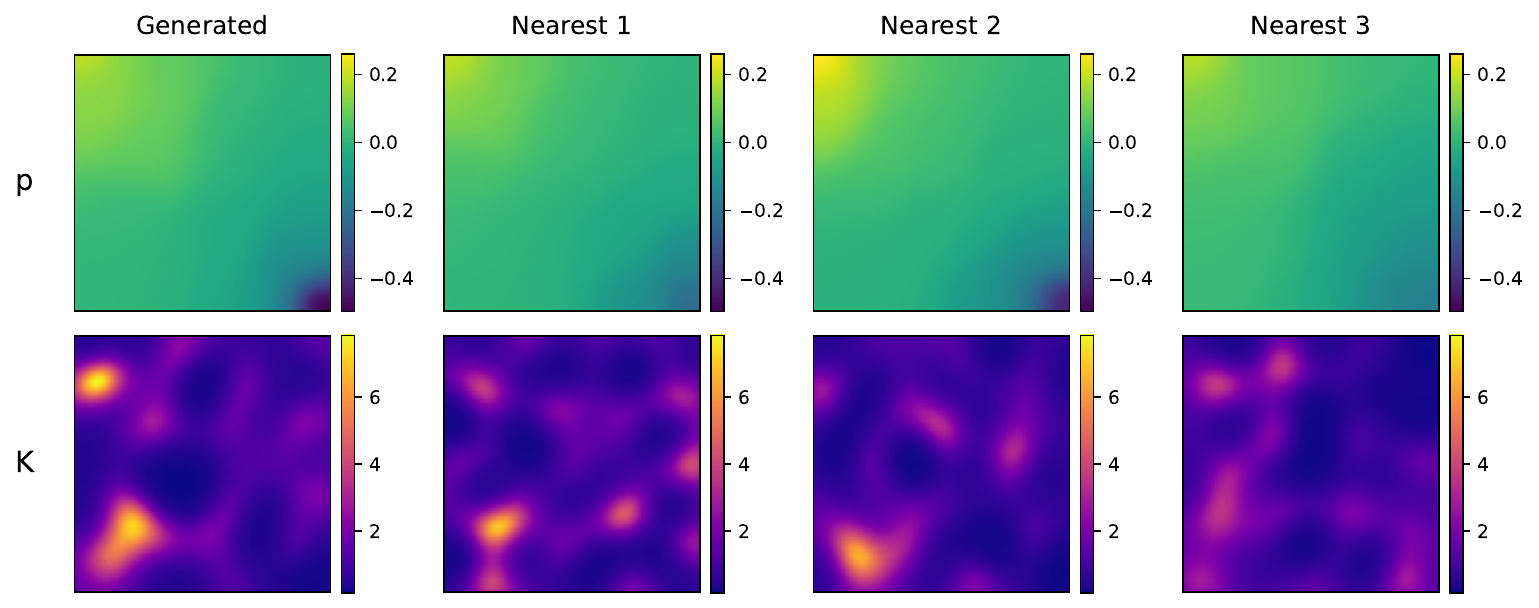}
    \end{subfigure}

    \vspace{0.1em}

    \begin{subfigure}{12cm}
        \centering
        \includegraphics[width=\linewidth]{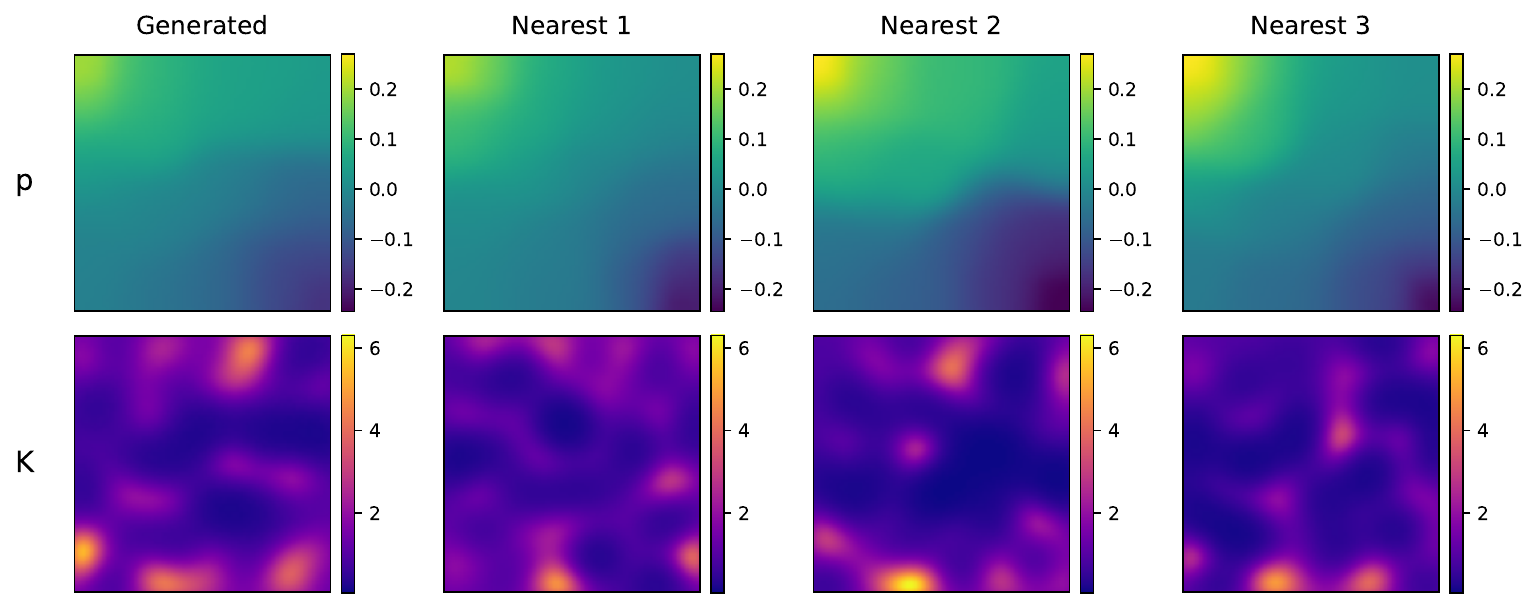}
    \end{subfigure}

    \caption{Samples generated using PIDM + CoCoGen with the proposed guidance and their three nearest neighbors in the training dataset for the Darcy flow experiment.}
    \label{novelty_check_darcy}
\end{figure}

In Figure \ref{novelty_check_darcy}, we randomly select four samples generated by PIDM + CoCoGen with the proposed guidance, which achieved the lowest residuals among the settings evaluated in Section \ref{sec_exp_darcy}. For each generated sample, we show the three closest training samples in terms of MSE. The generated samples differ from their nearest training samples, suggesting that the model does not simply reproduce the training data and can generate novel physics-aware samples.

\section{Details of the time-series 2D fluid experiment}\label{ap_fluid_set}
\subsection{Dataset}

\begin{figure}[t]
    \centering
    \begin{subfigure}{12cm}
        \centering
        \includegraphics[width=\linewidth]{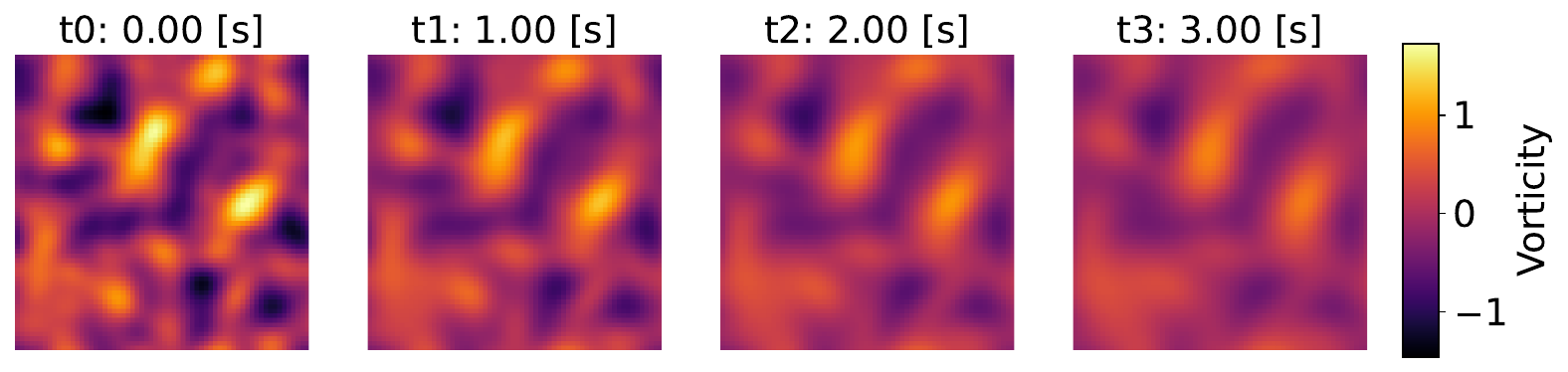}
    \end{subfigure}

    \vspace{0.5em}
    
    \begin{subfigure}{12cm}
        \centering
        \includegraphics[width=\linewidth]{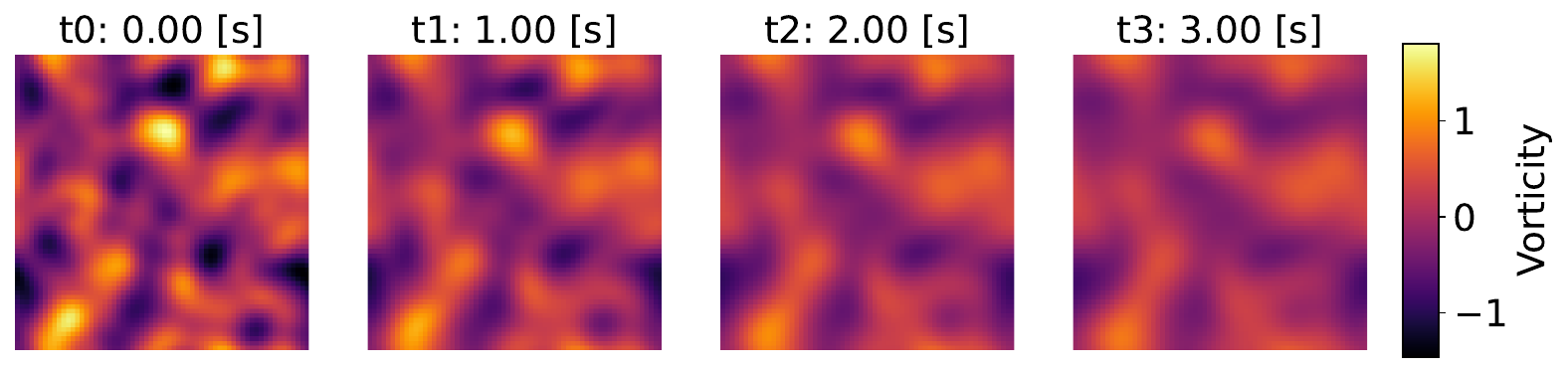}
    \end{subfigure}

    \vspace{0.5em}
    
    \begin{subfigure}{12cm}
        \centering
        \includegraphics[width=\linewidth]{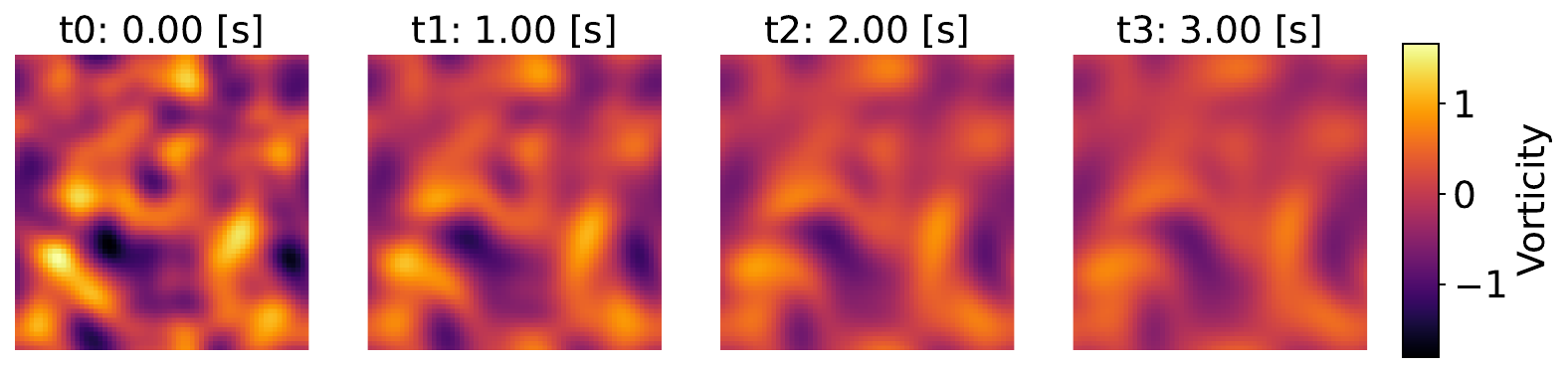}
    \end{subfigure}

    \vspace{0.5em}

    \begin{subfigure}{12cm}
        \centering
        \includegraphics[width=\linewidth]{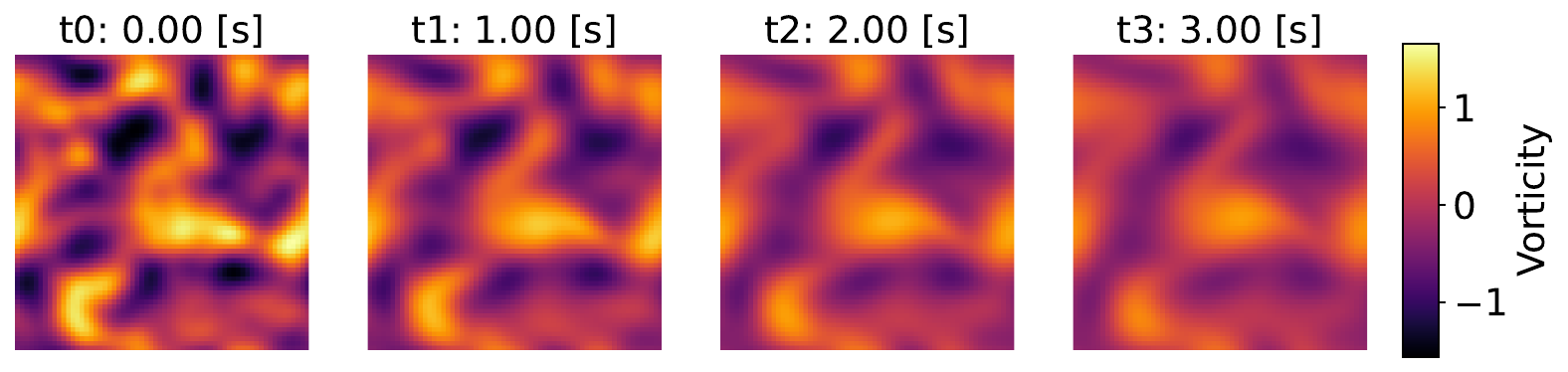}
    \end{subfigure}

    \vspace{0.5em}
    
    \begin{subfigure}{12cm}
        \centering
        \includegraphics[width=\linewidth]{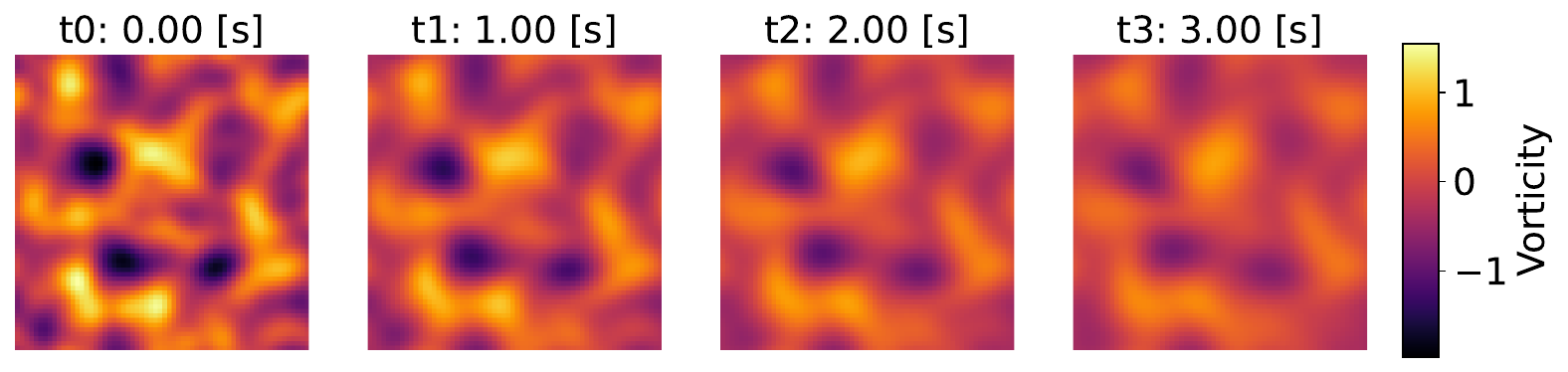}
    \end{subfigure}

    \caption{Examples of time-series 2D fluid data.}
    \label{fluid_data_samples}
\end{figure}

We construct a dataset of time-series vorticity snapshots $\omega$ using torch-cfd \citep{cao2025}, a PyTorch-based CFD library built on JAX-CFD \citep{kochkov2021, dresdner2023}. The viscosity is set to $1.5\times 10^{-2}$, the maximum initial velocity to 0.4, and the computational domain to $[0,5]\times[0,5]$. No external forcing is applied. The initial velocity fields are generated by applying a spectral filter to Gaussian noise. The governing equations are integrated with an internal time step of $0.05$. Each sample is represented as $x \in \mathbb{R}^{4\times64\times64}$, consisting of 4 snapshots at a resolution of $64\times64$. The snapshots span 3 seconds and are recorded at 1 second intervals. We generate 10000 training samples and 1000 validation samples. Examples are shown in Figure \ref{fluid_data_samples}.

\subsection{Implementation details}

The proposed method was implemented based on the code provided for DDIM \citep{song2021b}. All experiments were conducted on an NVIDIA GH200 GPU for both training and sampling.

The four vorticity channels were jointly normalized to $[-1,1]$ using the global minimum and maximum values computed from the corresponding training set. Before residual evaluation, generated samples were transformed back to their original physical scales.

Following \citet{song2021b}, we use a U-Net architecture \citep{ronneberger2015} to estimate the added noise. The model parameters are optimized using Adam \citep{kingma2017}. We use the cosine noise schedule proposed by \citet{nichol2021}. The fluid residual $r$ is rescaled to $r_\text{rescaled}\in[0, 1]$. The transformed residual is input to the model as a condition vector $\mathbf{c}=[r_\text{rescaled}, 1]\in\mathbb{R}^2$, where the second component distinguishes a valid condition from the null condition used for classifier-free guidance. The condition vector is embedded with an encoder neural network, and added to residual blocks of the U-Net. During classifier-free guidance training, the condition is replaced with the null condition $\mathbf{c}_{\varnothing}=[-1,0]$ with probability $0.2$. During sampling, the zero-residual target is represented by $\mathbf{c}=[0,1]$.

Table \ref{tb_unet_fluid} summarizes the model and experimental settings. The same model configuration is used for both the first-stage diffusion model and the subsequent guidance model.

\begin{table}[t]
    \centering
    \small
    \caption{Experimental settings for the time-series 2D fluid experiments.}
    \label{tb_unet_fluid}
    \begin{tabular}{ll}
        \toprule
        Parameter & Setting \\
        \midrule
        \multicolumn{2}{l}{\textbf{Model}} \\
        Architecture & U-Net \\
        Base channels & 64 \\
        Channel multipliers & $(1,2,2,4)$ \\
        Residual blocks & 2 per resolution \\
        Attention & $16\times16$ resolution \\
        Dropout & 0.2 \\
        EMA decay & 0.9999 \\

        \midrule
        \multicolumn{2}{l}{\textbf{Diffusion}} \\
        Number of diffusion steps & 1000 \\
        Variance schedule & Cosine (offset: 0.008) \\

        \midrule
        \multicolumn{2}{l}{\textbf{Training}} \\
        Optimizer & Adam \\
        Learning rate & $1.0\times10^{-4}$ \\
        Batch size & 32 \\
        Epochs & 200 \\
        Weight decay & $1.0\times10^{-4}$ \\
        Gradient clipping & 1.0 \\
        Classifier-free dropout & 0.2 \\

        \midrule
        \multicolumn{2}{l}{\textbf{Sampling}} \\
        Sampling steps & 100 \\
        DDIM stochasticity parameter $\eta$ & 1.0 \\
        Guidance scale $w$ & 2.0 \\
        \bottomrule
    \end{tabular}
\end{table}

\section{Additional results of the time-series 2D fluid experiment}\label{ap_fluid_ex}

\subsection{Residual distributions of individual trials}

\begin{figure}[t]
    \centering

    \begin{subfigure}{\linewidth}
        \centering
        \includegraphics[width=0.48\linewidth]{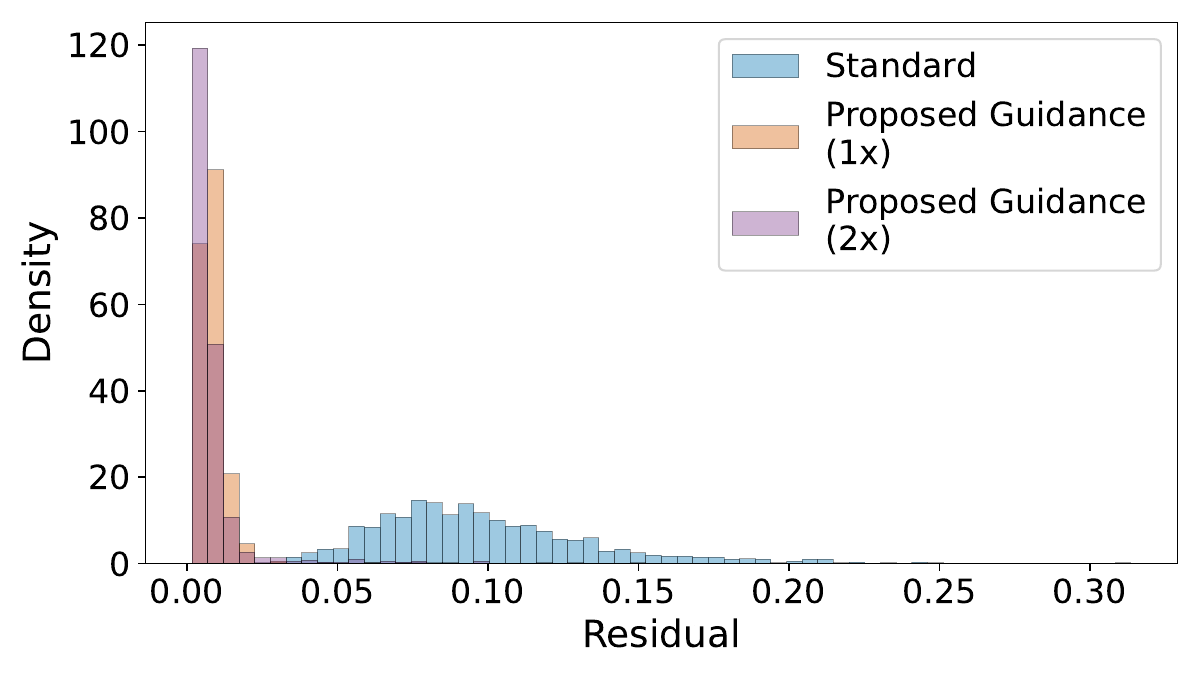}
        \hfill
        \includegraphics[width=0.48\linewidth]{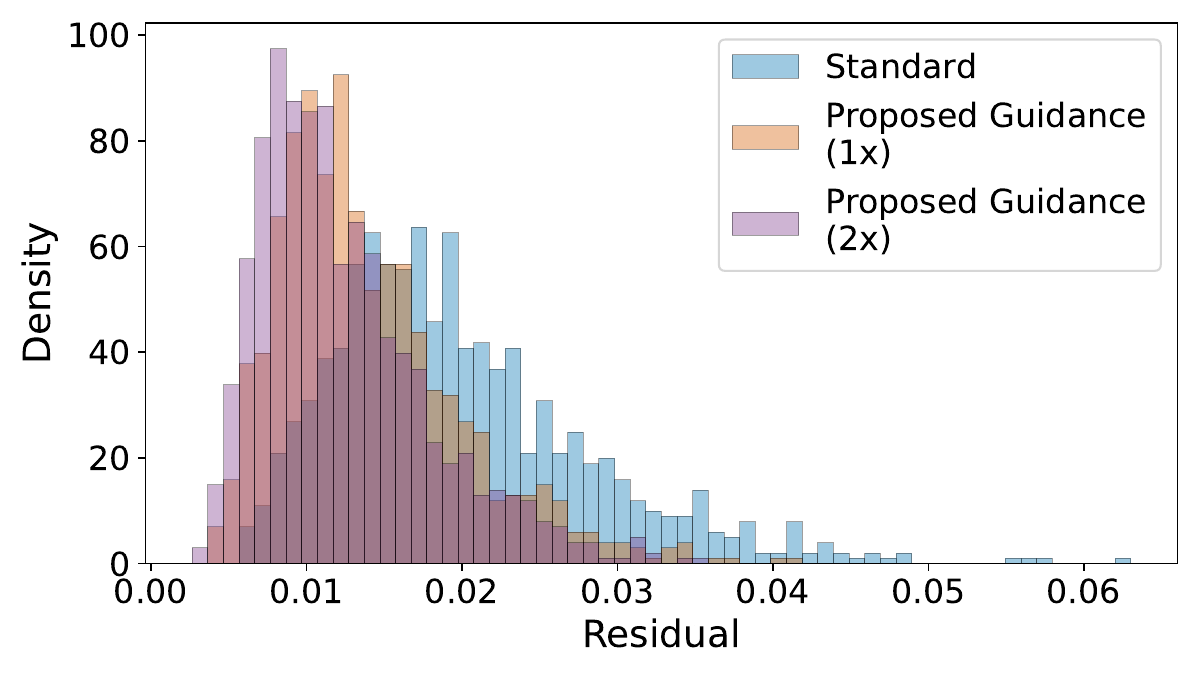}
    \end{subfigure}

    \vspace{0.5em}

    \begin{subfigure}{\linewidth}
        \centering
        \includegraphics[width=0.48\linewidth]{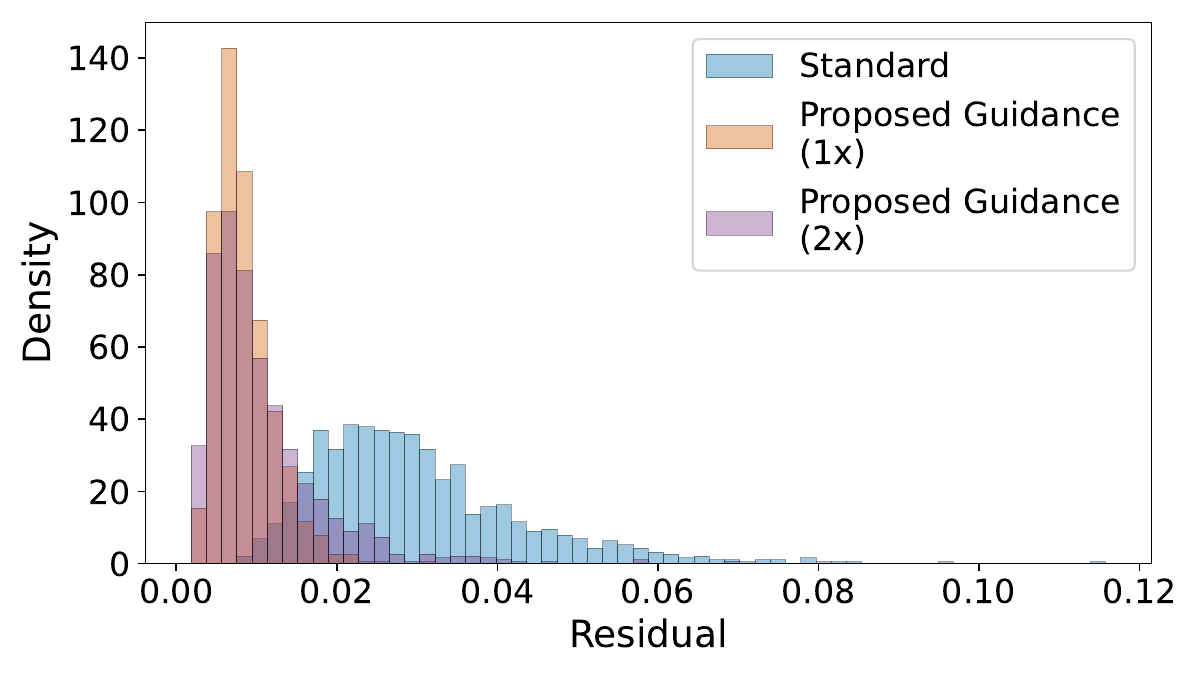}
        \hfill
        \includegraphics[width=0.48\linewidth]{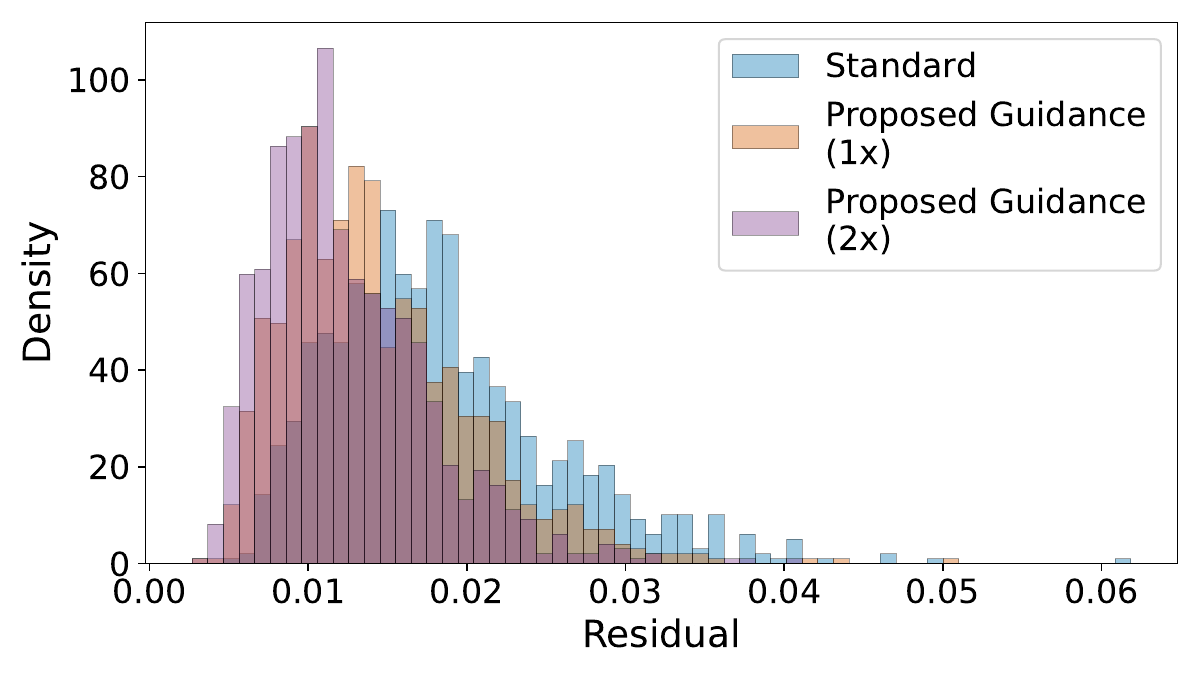}
    \end{subfigure}
    
    \vspace{0.5em}

    \begin{subfigure}{\linewidth}
        \centering
        \includegraphics[width=0.48\linewidth]{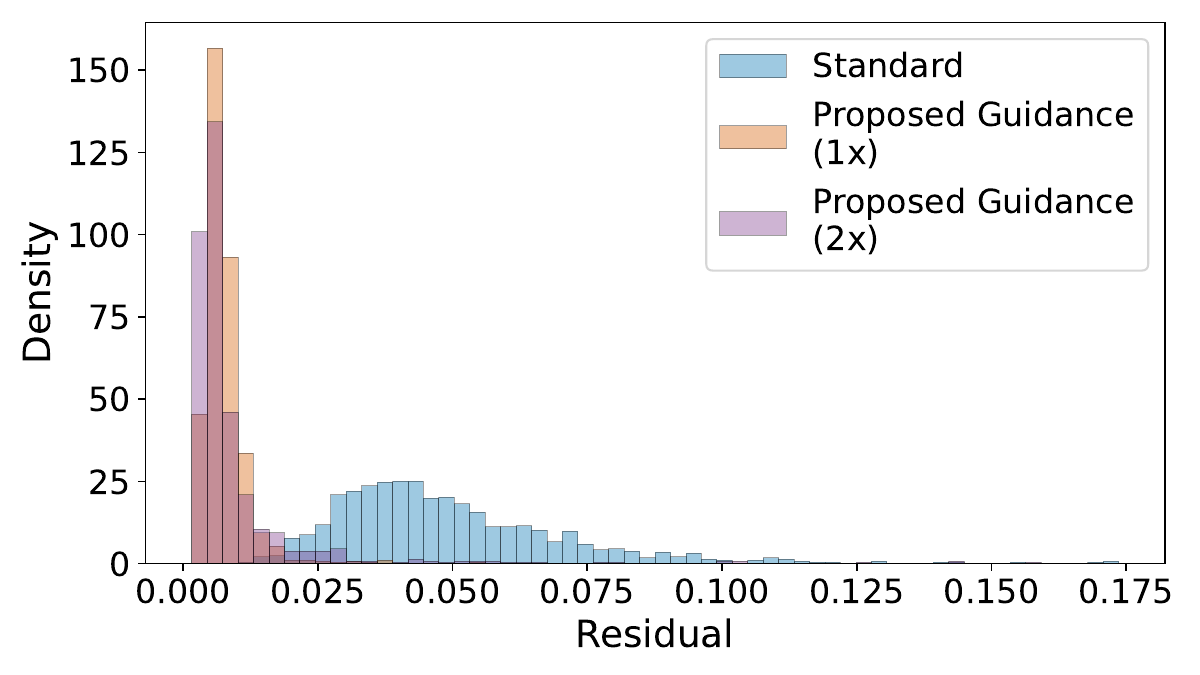}
        \hfill
        \includegraphics[width=0.48\linewidth]{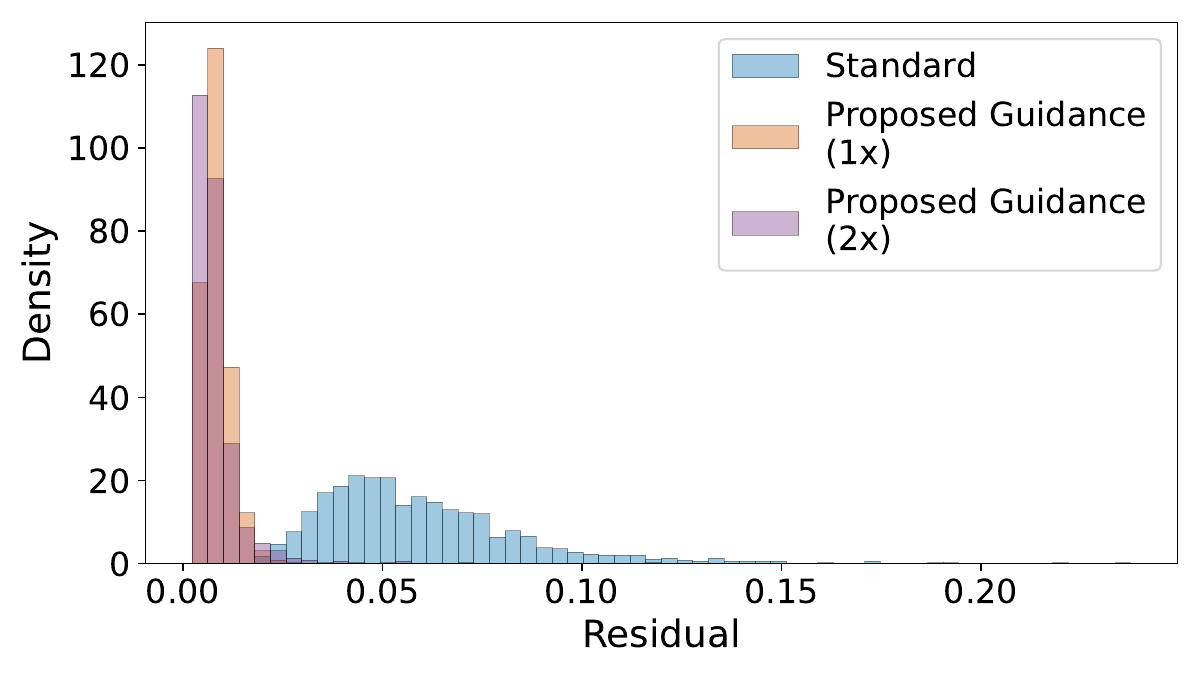}
    \end{subfigure}
    
    \caption{Residual distributions over 1000 generated samples for 6 independent trials in the time-series 2D fluid experiment. Each histogram corresponds to a different random seed.}
    \label{ap_dist_fluid_other}
\end{figure}

Figure \ref{2dist_fluid} in Section \ref{sec_exp_fluid} shows the residual distributions for 2 of the 8 trials, while Figure \ref{ap_dist_fluid_other} presents those for the remaining 6 trials. Each histogram is computed from 1000 samples generated in a single independent trial with a different random seed. In all trials, the distributions shift toward lower residual values when the proposed guidance is applied. The results also show a trend toward further reductions in the residuals when the proposed method is applied twice. These results are consistent with the findings reported in Section \ref{sec_exp_fluid}.

\subsection{Generated samples}

\begin{figure}[t]
    \centering

    \begin{subfigure}{\linewidth}
        \centering
        \includegraphics[width=0.48\linewidth]{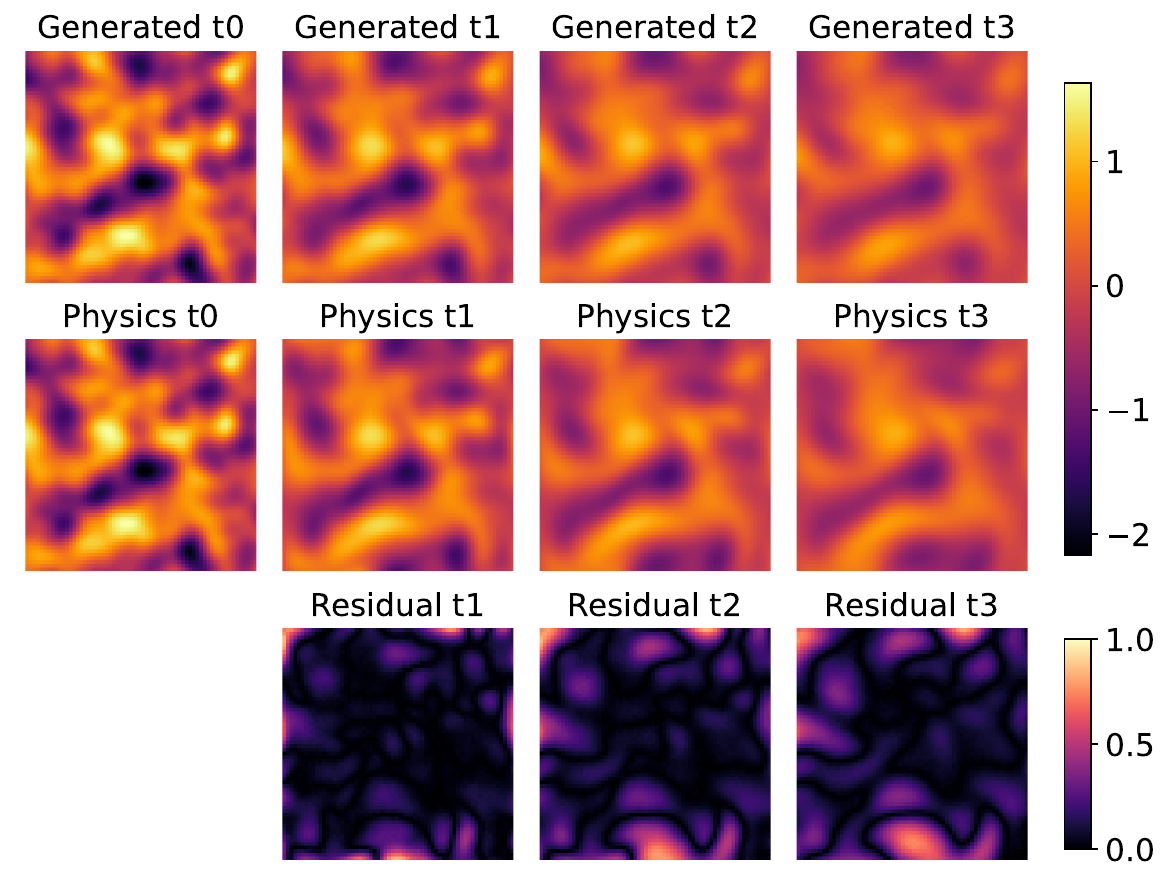}
        \hfill
        \includegraphics[width=0.48\linewidth]{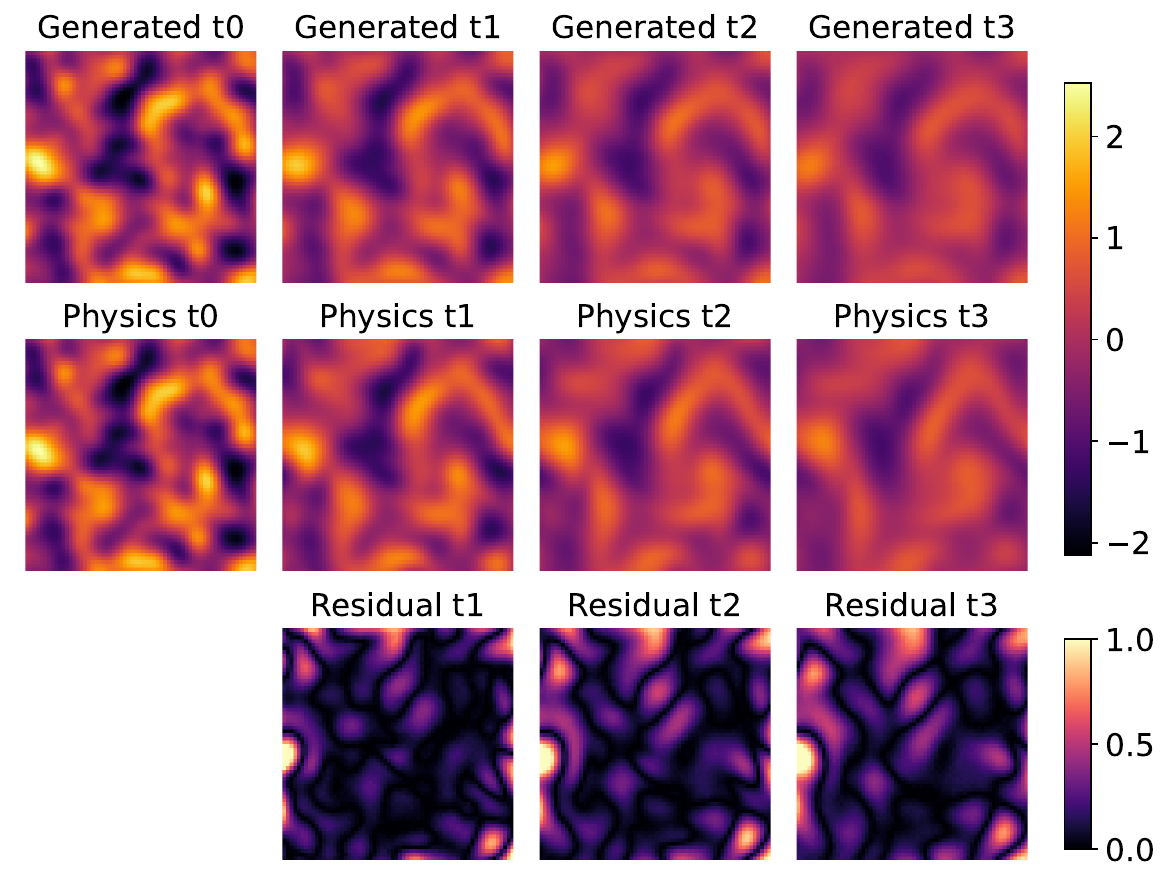}
        \caption{Standard diffusion model.}
    \end{subfigure}

    \vspace{0.8em}

    \begin{subfigure}{\linewidth}
        \centering
        \includegraphics[width=0.48\linewidth]{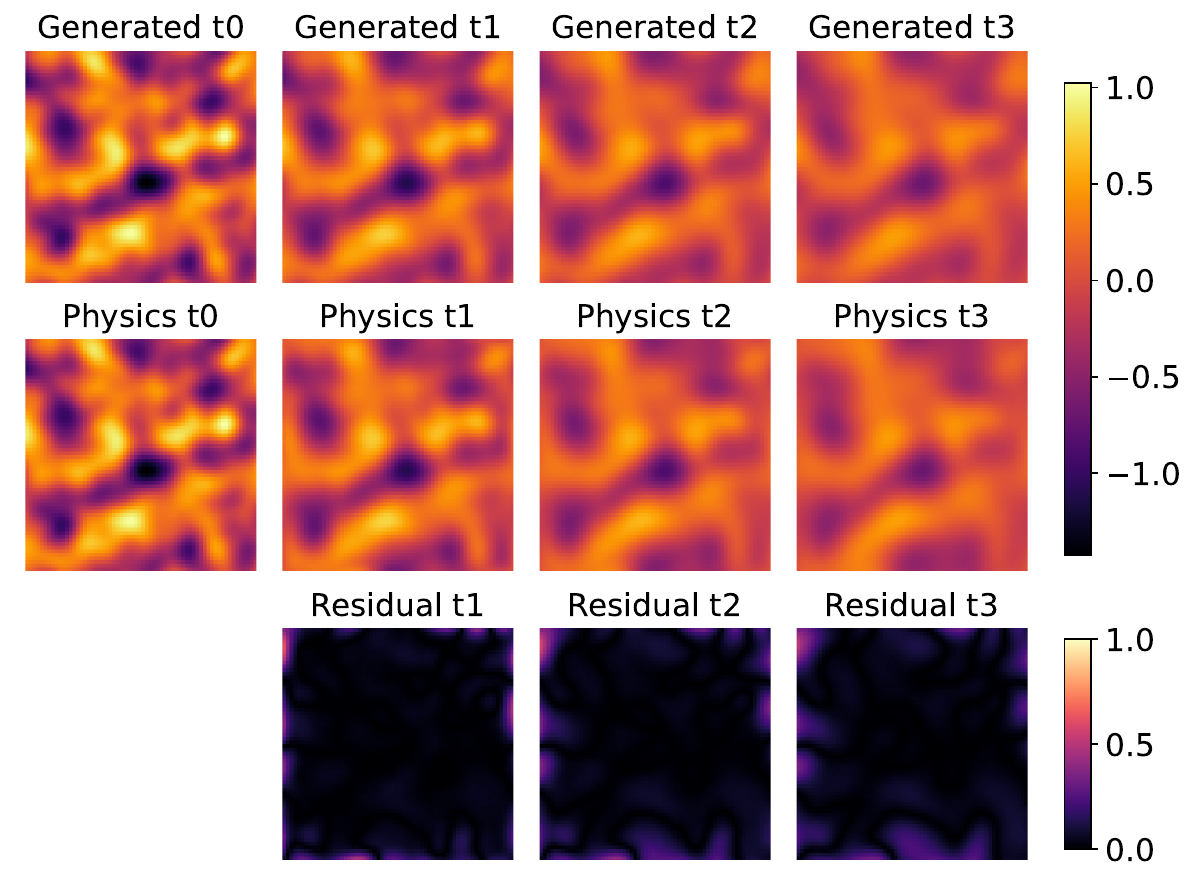}
        \hfill
        \includegraphics[width=0.48\linewidth]{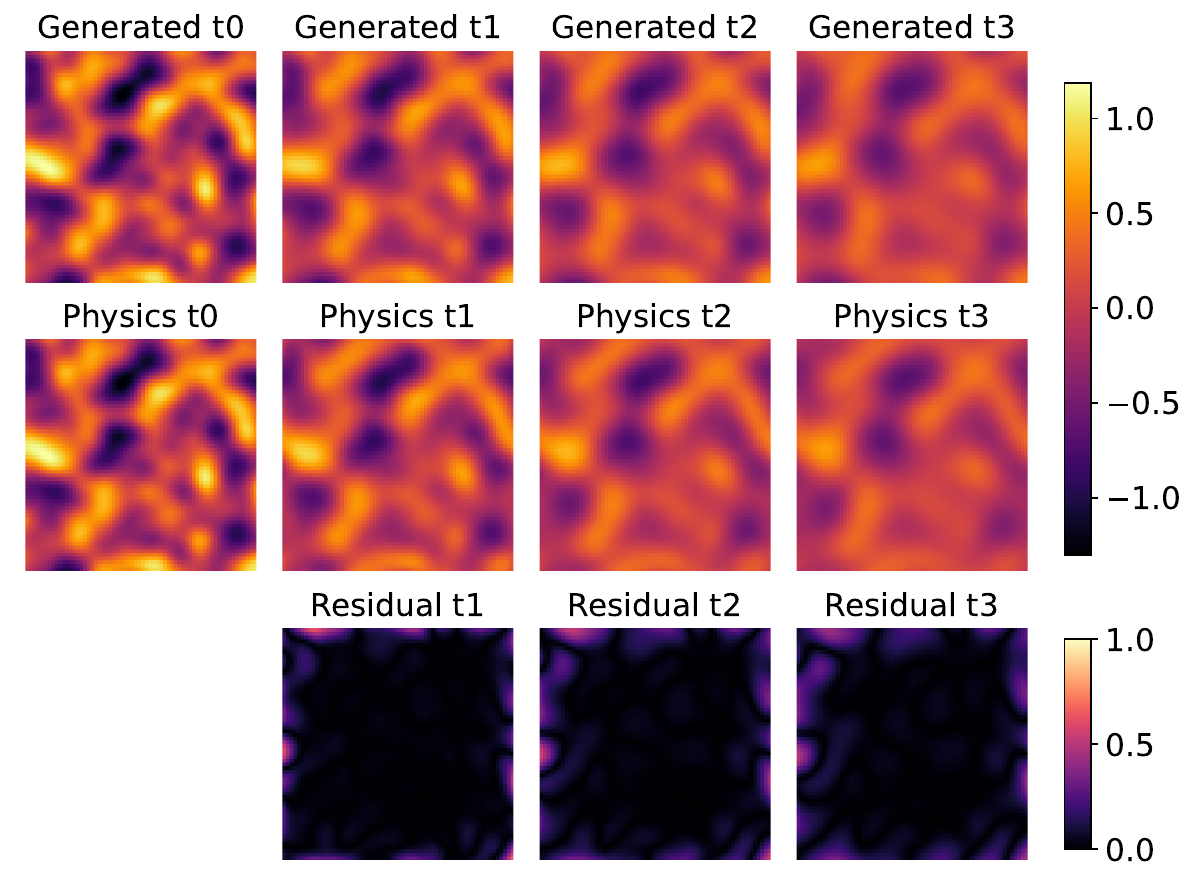}
        \caption{Proposed guidance model applied once.}
    \end{subfigure}

    \vspace{0.8em}

    \begin{subfigure}{\linewidth}
        \centering
        \includegraphics[width=0.48\linewidth]{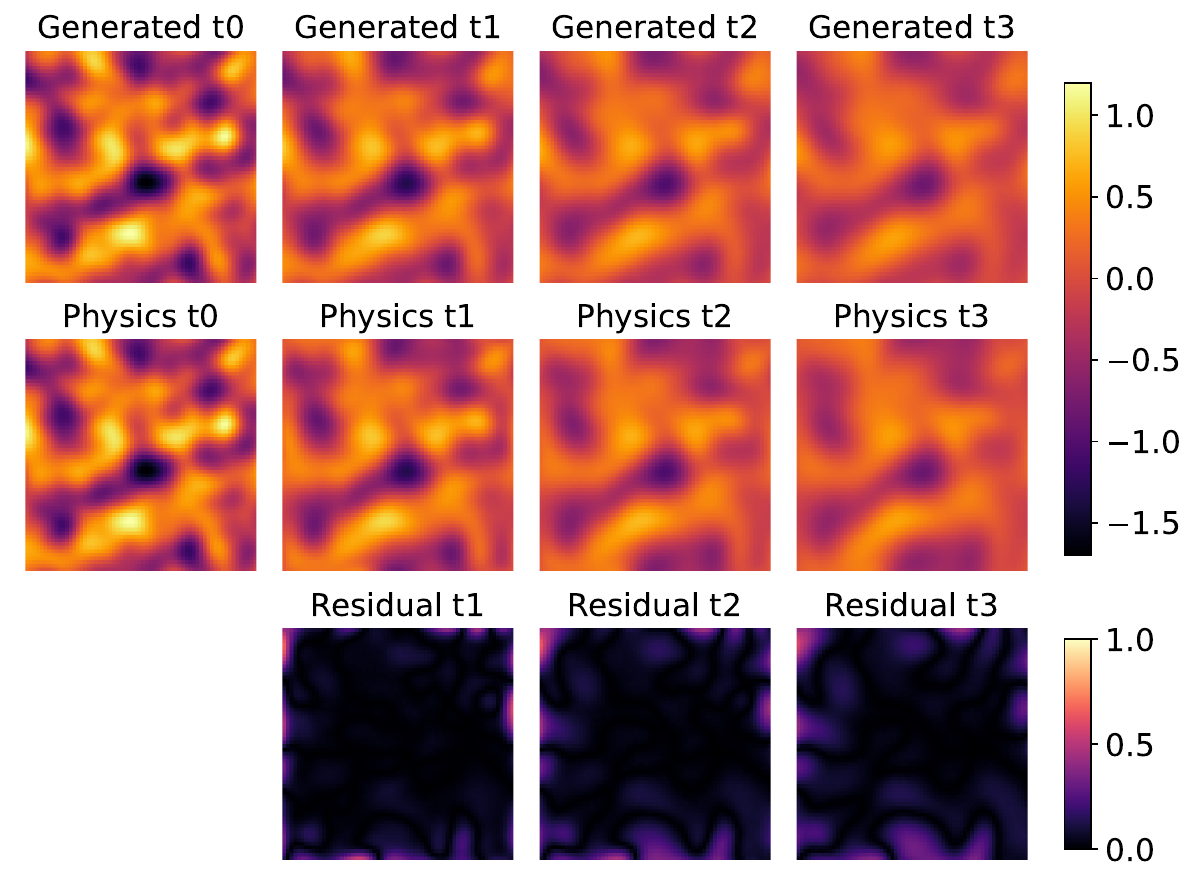}
        \hfill
        \includegraphics[width=0.48\linewidth]{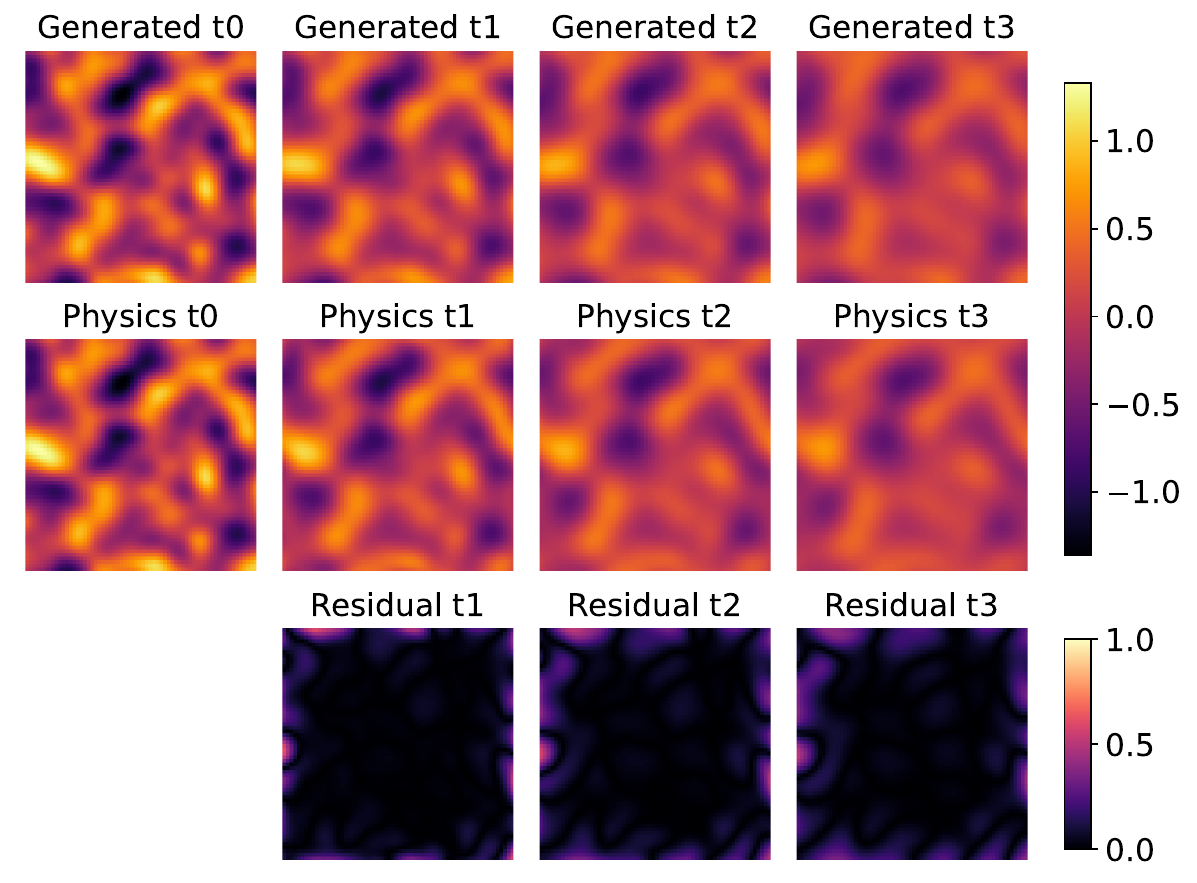}
        \caption{Proposed guidance model applied twice.}
    \end{subfigure}
    
    \caption{Comparison of samples and residuals generated by the standard diffusion baseline and the proposed guidance models in the time-series 2D fluid experiment.}
    \label{ap_comparison_fluid_1}
\end{figure}

\begin{figure}[t]
    \centering

    \begin{subfigure}{\linewidth}
        \centering
        \includegraphics[width=0.48\linewidth]{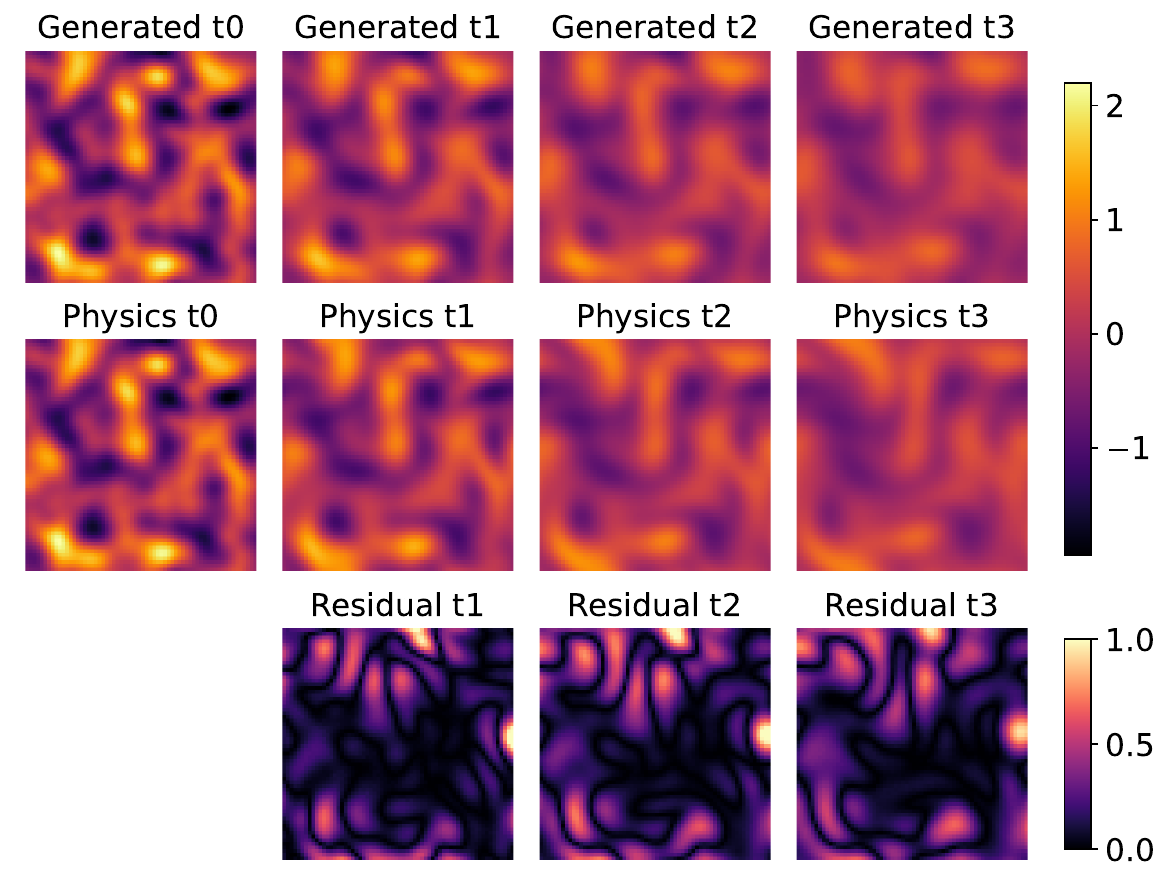}
        \hfill
        \includegraphics[width=0.48\linewidth]{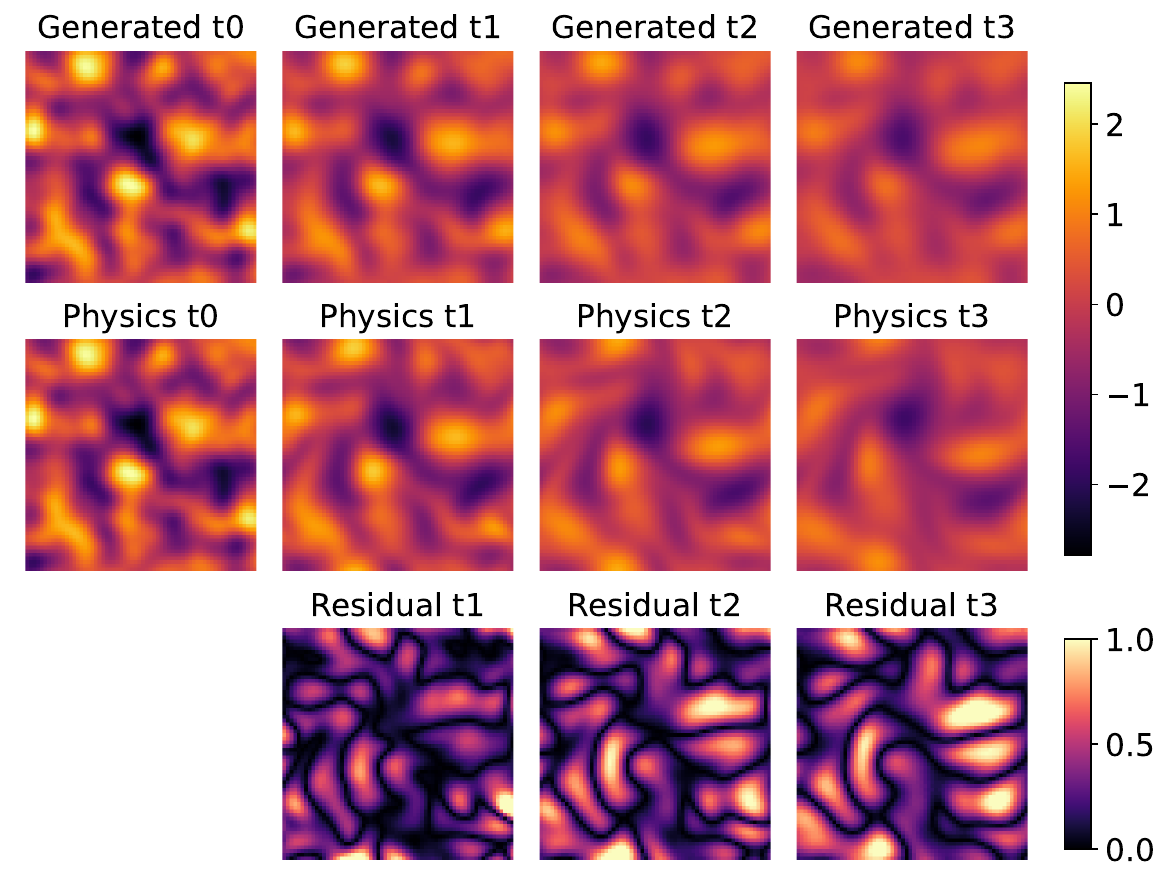}
        \caption{PG Diffusion.}
    \end{subfigure}

    \vspace{0.8em}

    \begin{subfigure}{\linewidth}
        \centering
        \includegraphics[width=0.48\linewidth]{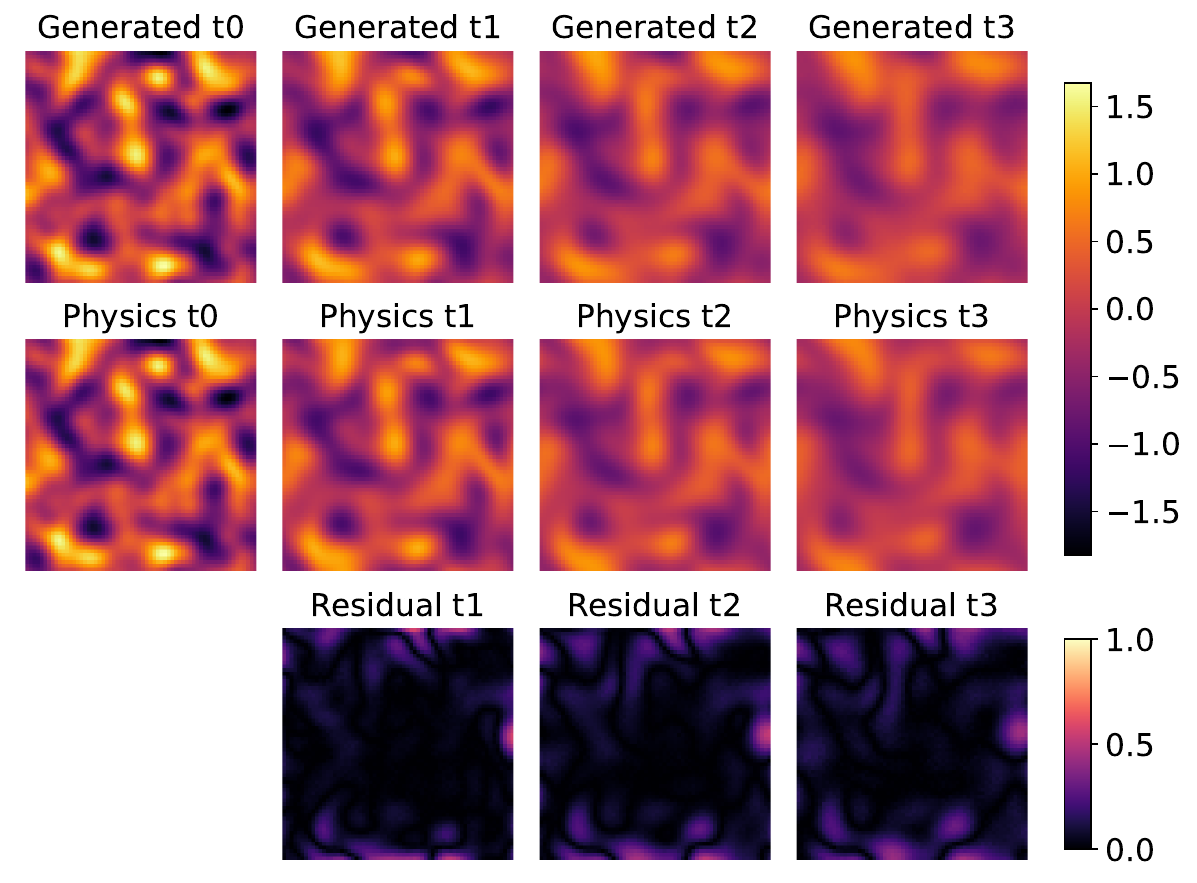}
        \hfill
        \includegraphics[width=0.48\linewidth]{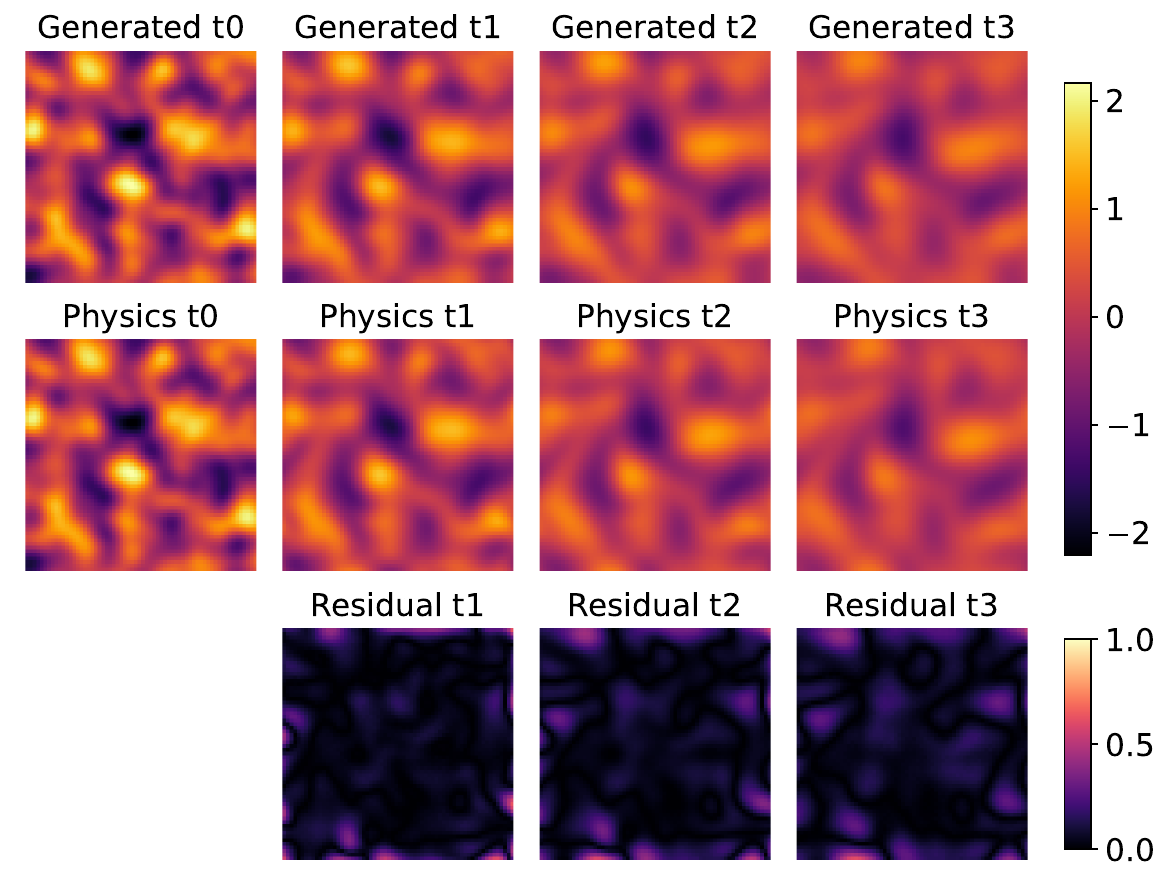}
        \caption{PIDM + CoCoGen.}
    \end{subfigure}

    \vspace{0.8em}

    \begin{subfigure}{\linewidth}
        \centering
        \includegraphics[width=0.48\linewidth]{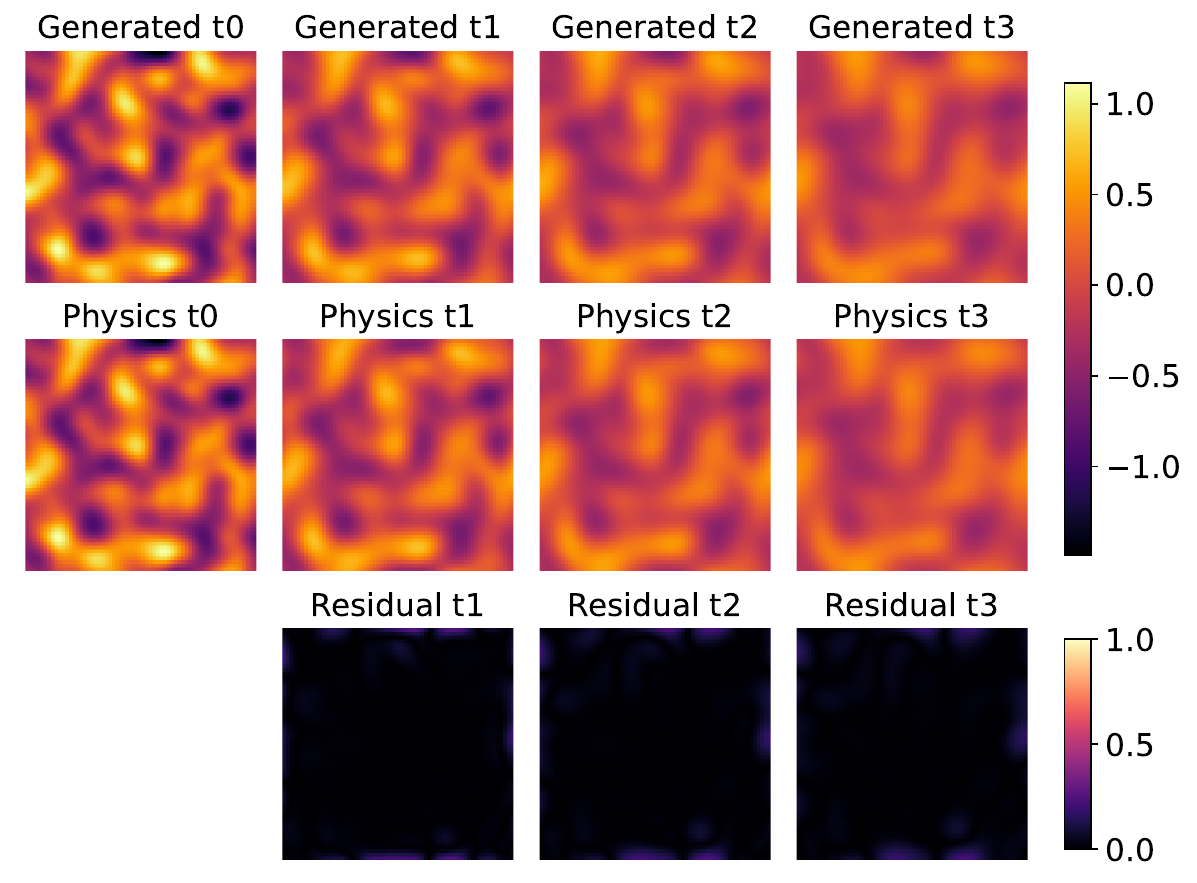}
        \hfill
        \includegraphics[width=0.48\linewidth]{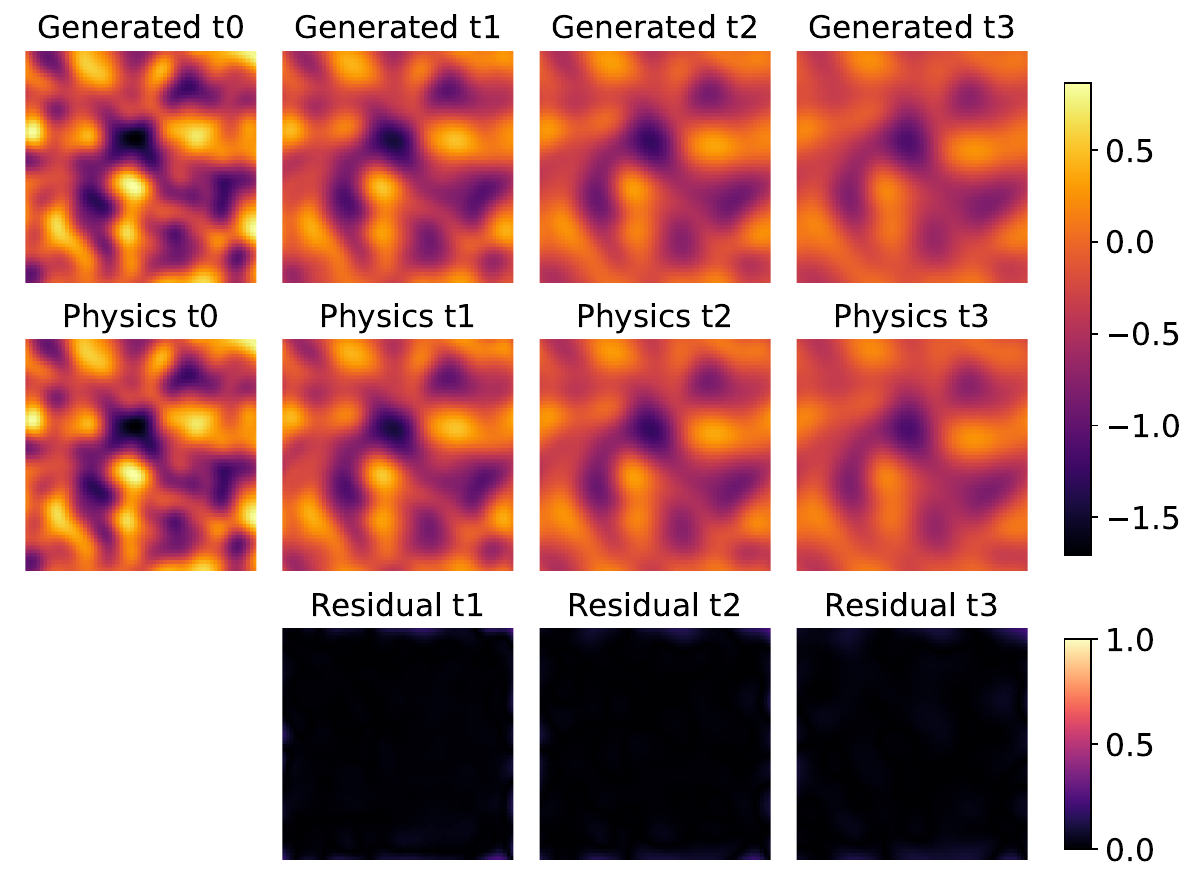}
        \caption{PIDM + CoCoGen with the proposed guidance applied once.}
    \end{subfigure}
    
    \caption{Comparison of samples and residuals generated by the physics-informed diffusion baselines and their combination with the proposed guidance in the time-series 2D fluid experiment.}
    \label{ap_comparison_fluid_2}
\end{figure}

Figures \ref{ap_comparison_fluid_1} and \ref{ap_comparison_fluid_2} show samples generated under different conditions. The residual represents the deviation between generated samples and the corresponding snapshots obtained by numerically integrating the Navier--Stokes equations from the same initial states. 

They illustrate that the proposed method reduces the residuals compared with the baselines and that its combination with PIDM and CoCoGen achieves the lowest residuals among the adopted settings.

\subsection{Novelty check}

As in the Darcy flow experiment, we compare the generated samples with the training dataset to assess whether the reduction in residuals could be attributed to memorization of training samples that satisfy the governing equations. This analysis examines whether the model can generate novel, physically consistent samples rather than simply reproducing the training data.

\begin{figure}[t]
    \centering
    \begin{subfigure}{0.48\linewidth}
        \centering
        \includegraphics[width=\linewidth]{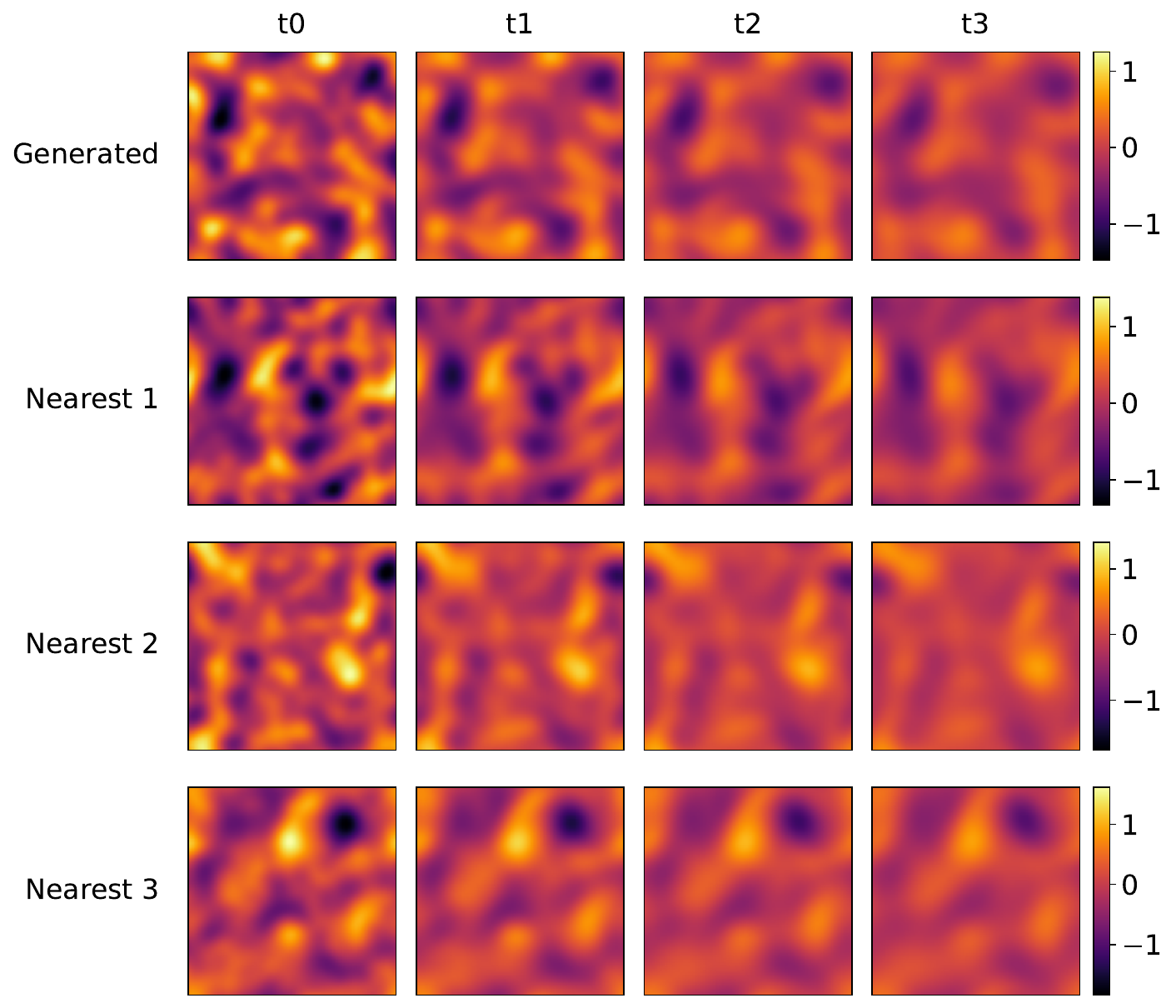}
    \end{subfigure}
    \hfill
    \begin{subfigure}{0.48\linewidth}
        \centering
        \includegraphics[width=\linewidth]{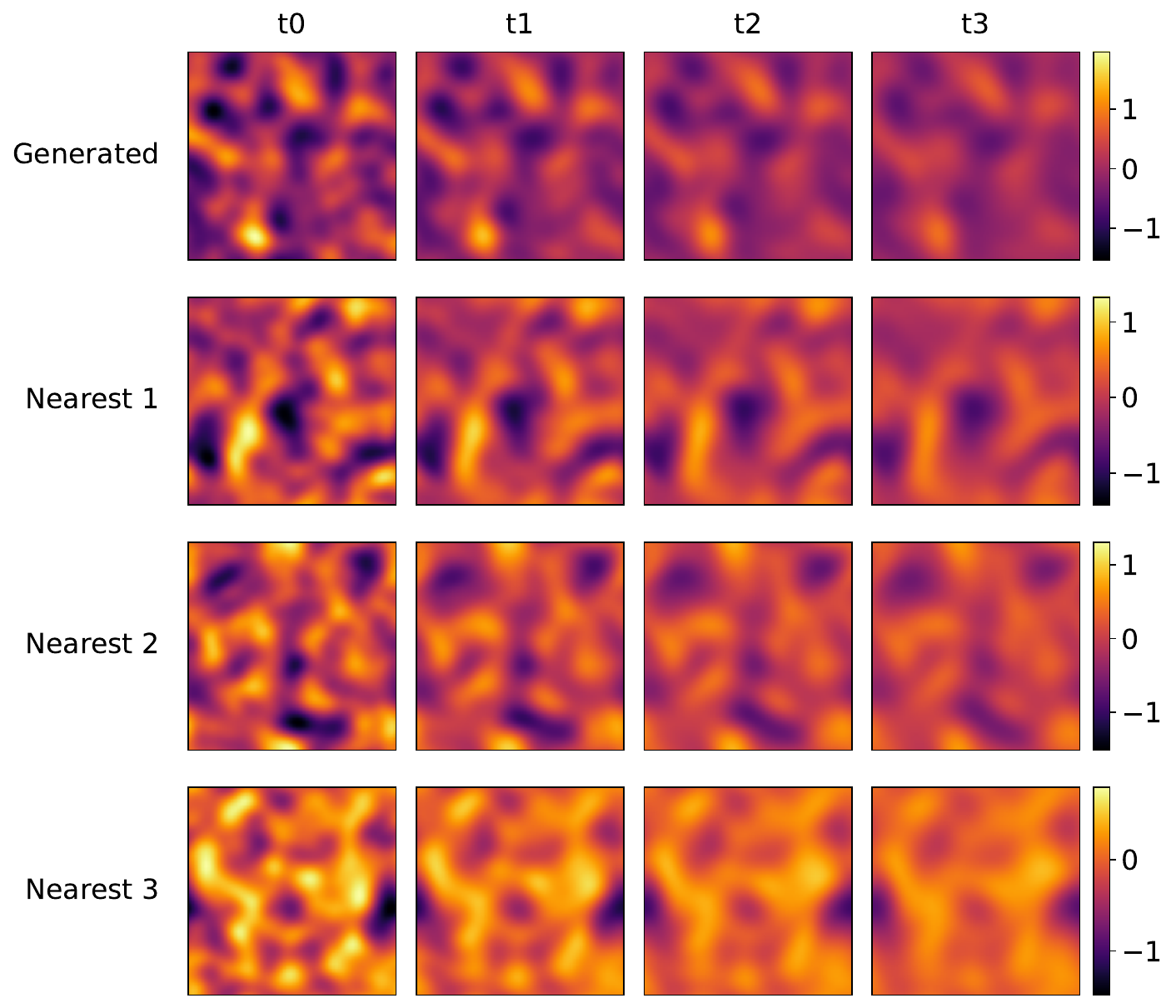}
    \end{subfigure}

    \vspace{0.5em}

    \begin{subfigure}{0.48\linewidth}
        \centering
        \includegraphics[width=\linewidth]{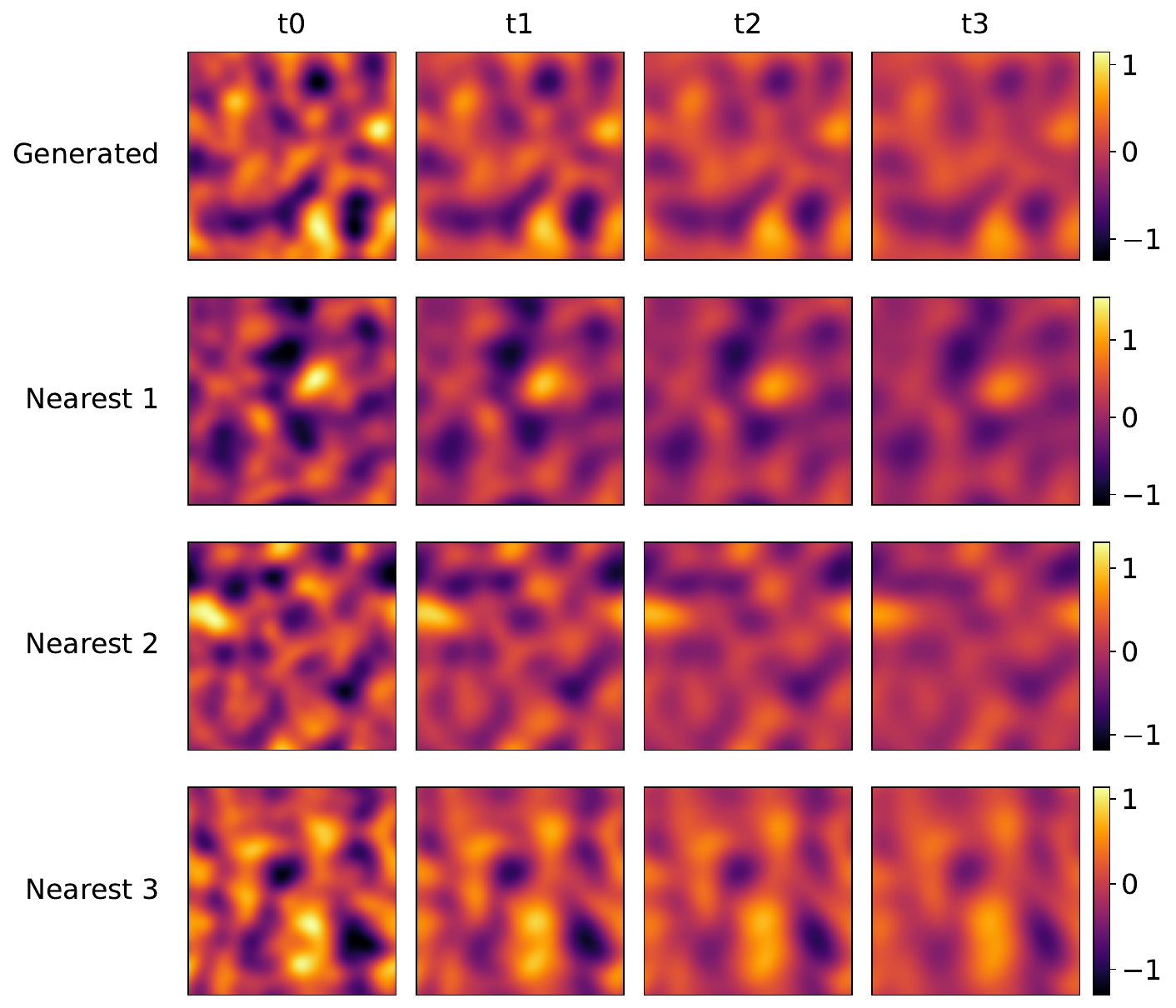}
    \end{subfigure}
    \hfill
    \begin{subfigure}{0.48\linewidth}
        \centering
        \includegraphics[width=\linewidth]{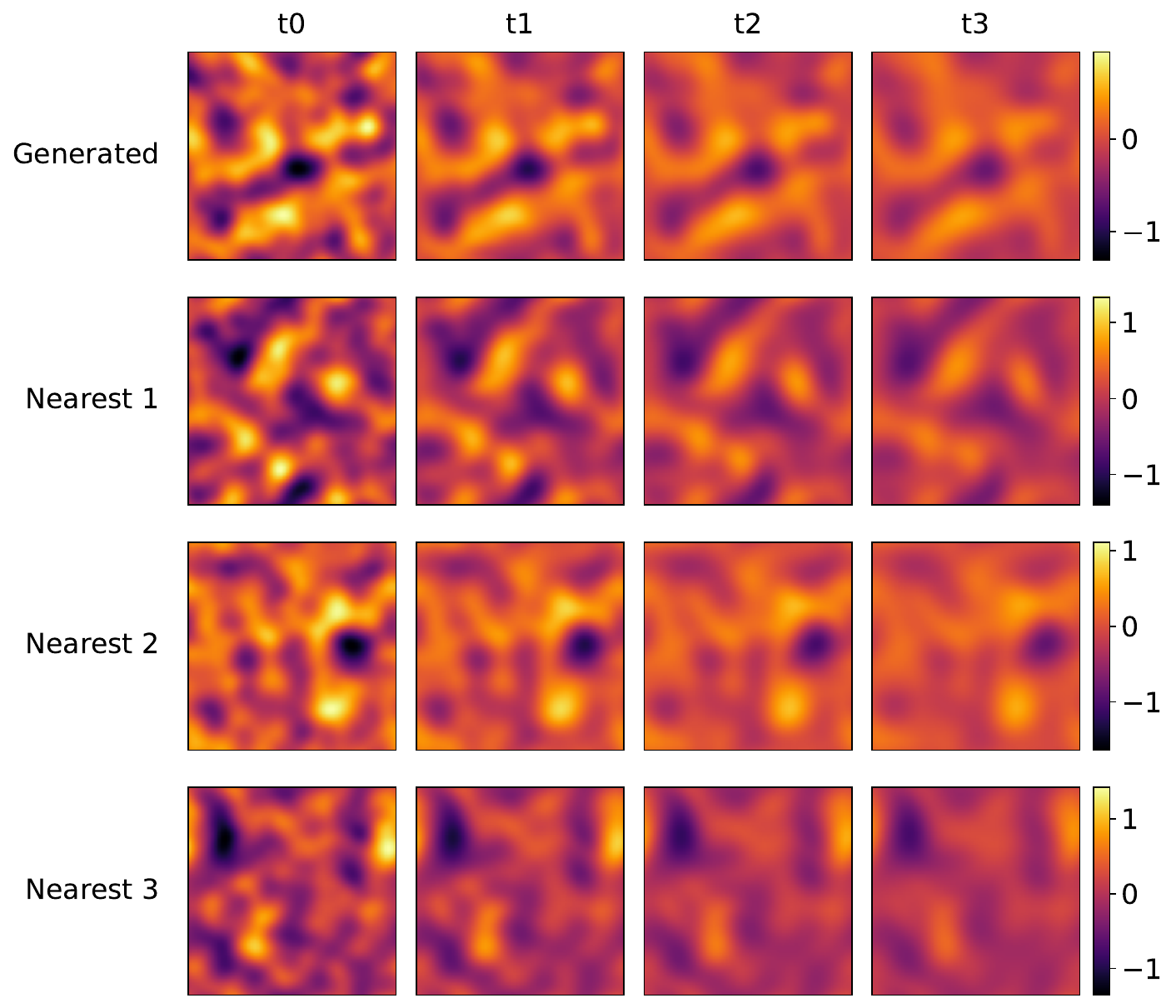}
    \end{subfigure}

    \caption{Samples generated using PIDM + CoCoGen with the proposed guidance and their three nearest neighbors in the training dataset for the time-series 2D fluid experiment.}
    \label{novelty_check_fluid}
\end{figure}

In Figure \ref{novelty_check_fluid}, we randomly select four samples generated by PIDM + CoCoGen with the proposed guidance, which achieved the lowest residuals among the settings evaluated in Section \ref{sec_exp_fluid}. Following the procedure used for the Darcy flow experiment in Figure \ref{novelty_check_darcy}, we show the three closest training samples for each generated sample in terms of MSE. The generated samples differ from their nearest training samples, suggesting that the model does not simply reproduce the training data and can generate novel physics-aware samples.

\end{document}

%% file: math_commands.tex
\usepackage{amsmath,amsfonts,bm}

\def\eqref#1{equation~\ref{#1}}

\def\1{\bm{1}}

\DeclareMathAlphabet{\mathsfit}{\encodingdefault}{\sfdefault}{m}{sl}
\SetMathAlphabet{\mathsfit}{bold}{\encodingdefault}{\sfdefault}{bx}{n}

